\documentclass[10pt,twocolumn,letterpaper]{article}

\usepackage{cvpr}              

\usepackage[dvipsnames]{xcolor}

\definecolor{cvprblue}{rgb}{0.21,0.49,0.74}
\usepackage[utf8]{inputenc}
\usepackage{times}
\usepackage{epsfig}
\usepackage{graphicx}
\usepackage{wrapfig}
\usepackage{gensymb}
\usepackage{amsmath,amsfonts,amssymb,amsthm}
\usepackage{color}
\usepackage{algorithm}
\usepackage{algorithmic}
\usepackage{caption}
\usepackage{subcaption}
\usepackage{xspace}
\usepackage{array}
\usepackage{cite}
\usepackage[pagebackref,breaklinks,colorlinks,citecolor=cvprblue]{hyperref}

\usepackage{multirow}

\newtoggle{arXiv}

\makeatletter
\DeclareRobustCommand\onedot{\futurelet\@let@token\@onedot}
\def\@onedot{\ifx\@let@token.\else.\null\fi\xspace}
\def\eg{e.g\onedot} 
\def\ie{i.e\onedot}

\def\wrt{wrt\onedot}

\makeatother

\newcommand{\boldparagraph}[1]{\vspace{0.2cm}\noindent{\bf #1:} }

\definecolor{darkgreen}{rgb}{0,0.7,0}

\def\confName{CVPR}
\def\confYear{2024}

\toggletrue{arXiv}

\title{IntrinsicAvatar: Physically Based Inverse Rendering of Dynamic Humans
from Monocular Videos via Explicit Ray Tracing}

\author{
  Shaofei Wang$^{1,2,3}$,\; Bo\v{z}idar Anti\'{c}$^{2,3}$,\; Andreas Geiger$^{2,3}$,\; Siyu Tang$^{1}$\\
  $^1$ETH Z\"{u}rich \quad
  $^2$University of T\"{u}bingen \quad
  $^3$T\"{u}bingen AI Center
}

\begin{document}
\twocolumn[{%
\renewcommand\twocolumn[1][]{#1}%
\maketitle
\centering
\begin{minipage}{1.0\linewidth}
\includegraphics[width=1.0\linewidth]{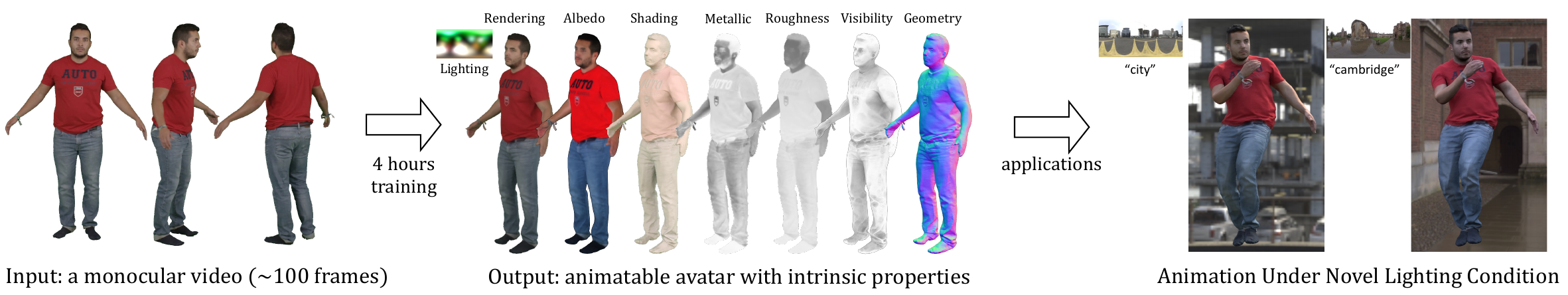}
\centering
\captionof{figure}{\textit{IntrinsicAvatar} aims to achieve physically based
inverse rendering of clothed humans from monocular videos.  \textbf{Left:} Our
model takes a monocular video as input and learns an avatar of the target
person. \textbf{Middle:} We show decomposed properties of the learned avatar.
Importantly, our model can produce such decomposition without any data-driven
prior on geometry, albedo, or material.  \textbf{Right:} With the learned avatar
and intrinsic properties, we can animate and relight the avatar using arbitrary
pose and arbitrary lighting condition.\label{fig:teaser}}
\vspace{1cm}
\end{minipage}
}]

\begin{abstract}
    We present IntrinsicAvatar, a novel approach to recovering the intrinsic properties of clothed human avatars including geometry, albedo, material, and environment lighting from only monocular videos. Recent advancements in human-based neural rendering have enabled high-quality geometry and appearance reconstruction of clothed humans from just monocular videos.  However, these methods bake intrinsic properties such as albedo, material, and environment lighting into a single entangled neural representation.  On the other hand, only a handful of works tackle the problem of estimating geometry and disentangled appearance properties of clothed humans from monocular videos. They usually achieve limited quality and disentanglement due to approximations of secondary shading effects via learned MLPs.  In this work, we propose to model secondary shading effects explicitly via Monte-Carlo ray tracing.  We model the rendering process of clothed humans as a volumetric scattering process, and combine ray tracing with body articulation.  Our approach can recover high-quality geometry, albedo, material, and lighting properties of clothed humans from a single monocular video, without requiring supervised pre-training using ground truth materials.  Furthermore, since we explicitly model the volumetric scattering process and ray tracing, our model naturally generalizes to novel poses, enabling animation of the reconstructed avatar in novel lighting conditions.
\end{abstract}
    
\section{Introduction}
\label{sec:introduction}
Photo-realistic reconstruction and animation of clothed human avatars is a
long-standing problem in augmented reality, virtual reality, and computer
vision. Existing solutions can achieve high-quality reconstruction for both
geometry and appearance of clothed humans given dense multi-view
cameras~\cite{Guo2019SIGGRAPH,Edoardo2022SIGGRAPH,Isik2023SIGGRAPH}. Recently,
reconstruction of clothed humans from monocular videos has also been
explored~\cite{Peng2024PAMI,Wang2022ECCV,Chen2023CVPR,Weng2022CVPR}. While these
approaches achieve satisfactory results, they model the appearance of clothed
humans as a single neural representation. This makes it difficult to edit the
physical properties of the reconstructed clothed human avatars, such as
reflectance and material, or to relight the reconstructed clothed human avatars
under novel lighting conditions. In this work, we aim to recover physically
based intrinsic properties for clothed human avatars including geometry, albedo,
material, and environment lighting from only monocular videos.

Physically based inverse rendering is a challenging problem in computer graphics
and computer vision. Traditional approaches tackle this problem as a pure
optimization problem with simplifying assumptions such as controlled, known
illumination. On the other hand, recent advances in neural fields have enabled
the high-quality reconstruction of geometry and surface normals from multi-view
RGB images. Given this progress, physically based inverse rendering of static
scenes under unknown natural illumination has been demonstrated
\cite{Jin2023CVPR,Zhu2023CVPR}.
Most recently, various works have combined human body priors with the
physically based inverse rendering pipeline to reconstruct clothed human avatars
with disentangled geometry, albedo, material, and lighting from monocular
videos~\cite{Chen2022ECCVa,Sun2023ICCV,Iqbal2023ICCV}. However, these methods either ignore
physical plausibility or model secondary shading effects via approximation,
resulting in limited quality of reconstructed human avatars.

Two major challenges are present for physically based inverse rendering of
clothed humans from monocular videos: {\it (1)} accurate geometry
reconstruction, especially normal estimates are essential for high-quality
inverse rendering.  {\it (2)} Modeling secondary shading effects such as shadows
and indirect illumination is expensive and requires a certain level of
efficiency to query the underlying neural fields. Existing monocular geometry
reconstruction methods of clothed humans all rely on large MLPs to achieve
high-quality geometry reconstruction.  However, using large MLPs negatively
impacts the efficiency of secondary shading computation. Therefore, most
existing methods are forced to rely on simple assumptions (no shadows, no
indirect illumination) or approximations (pre-trained MLPs) to model secondary
shading effects. More efficient neural field representations such as instant NGP
(iNGP~\cite{Mueller2022SIGGRAPH}) have proven to be effective for geometric
reconstruction given multiple input views of a static
scene~\cite{Alexandru2023CVPR,Li2023CVPR,Wang2023ICCV}, but it remains a
challenge to extend such representation to dynamic humans under monocular setup.

In this paper, we employ iNGP with hashing-based volumetric representation and
signed distance field (SDF) to achieve fast and high-quality reconstruction
of clothed humans from monocular videos.  The high-quality initial geometry
estimation and efficiency of iNGP facilitate the modeling of inverse rendering via
explicit Monte-Carlo ray tracing.  Furthermore, traditional surface-based
inverse rendering methods give ambiguous predictions at edges and boundaries.  We
propose to use volumetric scattering to model edges and boundaries in a more
physically plausible way.  Our experiments demonstrate that we can achieve
high-quality reconstruction of clothed human avatars with disentangled geometry,
albedo, material, and environment lighting from only monocular videos.  In
summary, we make the following contributions:
\begin{itemize}
    \item We propose a model for fast, high-quality geometry reconstruction of
        clothed humans from monocular videos.
    \item We propose to combine volumetric scattering with the human body
        articulation for physically based inversed rendering of dynamic clothed
        humans.  We use explicit Monte-Carlo ray tracing in canonical space
        to model the volumetric scattering process, enabling relighting for
        unseen poses.
    \item We demonstrate that our method can achieve high-quality reconstruction
        of clothed human avatars with disentangled geometry, albedo, material,
        and environment lighting from only monocular videos of clothed humans.
        We also show that our learned avatars can be rendered realistically
        under novel lighting conditions \textit{and} novel poses.
\end{itemize}
We have made our code and models publicly available\footnote{\iftoggle{arXiv}{\href{https://neuralbodies.github.io/IntrinsicAvatar/}{\color{black}{https://neuralbodies.github.io/IntrinsicAvatar/}}}{https://neuralbodies.github.io/IntrinsicAvatar/}}.
\section{Related Work}
\label{sec:related_work}

\boldparagraph{Traditional Inverse Rendering} Traditional approaches to inverse
rendering work on either single RGB images~\cite{Barron2015PAMI,
Li2018SIGGRAPH, Sengupta2019ICCV, Yu2019CVPR, Sang2020ECCV, Wei2020ECCV,
Li2020CVPRa, Lichy2021CVPR} or multi-view, multi-modality
inputs~\cite{Lensch2003SIGGRAPH,Schmitt2020CVPR,Park2020CVPR,Guo2019SIGGRAPH,Zhang2021SIGGRAPHa,Laffont2013TVCG,Nam2018SIGGRAPH,Philip2019SIGGRAPH,Goel20203DV}.
Recovering shape, reflectance, and illumination from a single RGB image is
heavily underconstrained and often works poorly on real-world setups such as
scene-level reconstruction and articulated object reconstruction. A more
practical approach is to reconstruct shapes from multi-view RGB(D) images and
make simplifying assumptions such as controlled lighting conditions
\cite{Schmitt2023PAMI, Luan2021EG, Nam2018SIGGRAPH}.  This kind of approach
often results in high-quality reconstruction of physical properties but lacks
flexibility.

\boldparagraph{Physically Based Inverse Rendering with Neural Fields} Since the
blossom of neural radiance fields (NeRF~\cite{Mildenhall2020ECCV}), a variety of
works have been proposed to tackle the inverse rendering problem using neural
field representations. However, many works make use of simplifying assumptions
such as known lighting conditions~\cite{Srinivasan2021CVPR}, ignoring shadowing
effects~\cite{Boss2021ICCV,Zhang2021CVPRb,Munkberg2022CVPR,Boss2021NeurIPS}, or
assuming constant material~\cite{Zhang2021CVPRb}.
NeRFactor~\cite{Zhang2021SIGGRAPHASIA} was the first work that enabled full
estimation of a scene's underlying physical properties (geometry, albedo, BRDF,
and lighting) under a single unknown natural illumination while also taking
shadowing effect into account.  InvRender~\cite{Zhang2022CVPRa} builds upon the
state-of-the-art shape and radiance field reconstruction
methods~\cite{Yariv2020NIPS,Wang2021NEURIPSa} and proposed to model indirect
illumination by distilling a pre-trained NeRF into auxiliary
MLPs.~\cite{Lyu2022ECCV} learns a neural radiance transfer field to enable
global illumination under novel lighting conditions, but relies on accurate
geometry initialization and does not optimize it jointly with material and
lighting. NVDiffRecMC~\cite{HasselgrenNEURIPS2022} tackles the inverse rendering
problem by exploring the combination of mesh-based Monte-Carlo ray tracing and
off-the-shelf denoisers. However, the mesh-based representation of NVDiffRecMC
gives less accurate reconstruction compared
to~\cite{Yariv2020NIPS,Wang2021NEURIPSa}.

Most recently, TensoIR~\cite{Jin2023CVPR} takes advantage of fast radiance field
data structures~\cite{Chen2022ECCV} and conducts explicit visibility and
indirect illumination estimation via ray marching. In comparison, we use an SDF
representation and combine iso-surface search technique with volumetric
scattering, resulting in better visibility modeling, especially for cloth
wrinkles. Most importantly, we target dynamic, animatable clothed avatar
reconstruction while TensoIR focuses on static scene reconstruction.

Microfacet fields~\cite{Mai2023ICCV} proposed to utilize volumetric scattering
with a surface BRDF and ad-hoc sampling strategies. Concurrent
to~\cite{Mai2023ICCV} , NeMF~\cite{Zhang2023ICCV} also proposed to use
volumetric scattering with microflake phase
functions~\cite{Jakob2010SIGGRAPH,Heitz2015SIGGRAPH} to replace surface-based
BRDF for volume scattering, resulting in the ability to reconstruct thin
structures and low-density volumes. Both methods focus on static scenes
reconstruction and relighting while using density fields to represent the
underlying geometry.

\boldparagraph{Neural Radiance Fields for Human Reconstruction} Neural radiance
fields have been used for human reconstruction from monocular videos. Most
works~\cite{Peng2021CVPR,Peng2021ICCV,Weng2022CVPR,Jiang2023CVPR,Li2022ECCV,Feng2022SIGGRAPHAsia}
focus on appearance reconstruction while using density fields as a noisy
geometry proxy. Some methods use SDFs to represent the geometry of humans and
achieve impressive results in both geometry reconstruction and photo-realistic
rendering~\cite{Xu2021NEURIPSb,Wang2022ECCV,Peng2024PAMI,Chen2023CVPR}.
However, these methods bake intrinsic properties such as albedo, material, and
lighting all into the learned neural representations, preventing the application
of these methods in relighting and material editing.

\boldparagraph{Physically Based Inverse Rendering of Humans} High-quality 3D
relightable human assets can be obtained via a multi-view, multi-modality
capture system with controlled
lighting~\cite{Guo2019SIGGRAPH,Bi2021SIGGRAPH,Zhang2021SIGGRAPHa,Iwase2023CVPR,Chen2024CVPR,Saito2024CVPR}
or by training regressors on high-quality digital 3D
assets~\cite{Alldieck2022CVPR,Corona2023CVPR,Zheng2023CVPR}.
RANA~\cite{Iqbal2023ICCV} pre-trains a mesh representation on multiple subjects
with ground truth 3D digital assets while using a simplified spherical
harmonics lighting model, thus cannot handle secondary shading effects such as
shadows and indirect illumination.~\cite{Sun2023ICCV} propose to model the
secondary shading effects via spherical Gaussian approximations, which do not
handle shadowing effects.  Relighting4D~\cite{Chen2022ECCVa} jointly estimates
the shape, lighting, and the albedo of dynamic humans from monocular videos
under unknown illumination by approximating visibility via learned MLPs. These
learned MLPs are over-smoothed approximations to real visibility values, while
also having the inherent problem of not being able to generalize to novel
poses. In contrast, we employ fast, exact visibility querying via explicit ray
tracing, and thus can generalize to any novel poses. 

\boldparagraph{Concurrent Works}
~\cite{Bharadwaj2023SIGGRAPHAsia} and~\cite{Kalshetti2024WACV} respectively
reconstruct relightable faces and hands from monocular videos. For full-body
relightable avatars,~\cite{Lin2024AAAI} proposes to construct part-wise light
visibility MLPs to achieve better novel pose generalization for relighting.
However, it needs to train light visibility MLPs on additional unseen poses. In
comparison, we use explicit ray tracing to compute secondary ray visibilities,
which generalizes to novel poses without additional training.~\cite{Xu2024CVPR}
designs a hierarchical distance query algorithm and extends
DFSS~\cite{Parker2018} to deformable neural SDF, achieving efficient light
visibility computation using sphere tracing.  However, the use of sphere tracing
and surface rendering results in visible artifacts around elbows and armpits, as
sphere tracing does not guarantee convergence, especially when combined with
human body articulation. In contrast, we use volumetric scattering to model the
human body, which results in less visual artifacts.

\section{Method}
\label{sec:method}
\begin{figure*}[t]
\centering
  \includegraphics[width=1.0\textwidth]{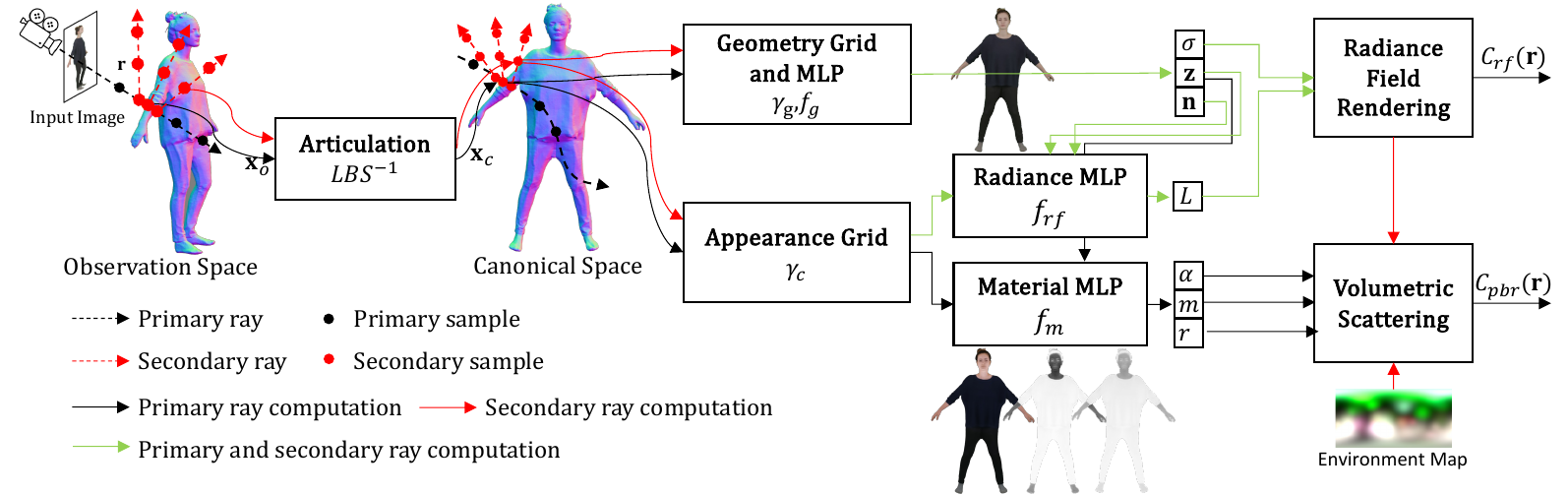}
  \caption{\textbf{Inverse Rendering of Clothed Avatars with Volumetric
    Scattering.} Given an input image and associated camera rays, we warp the
    rays to the canonical space and do both primary and secondary ray
    marching/tracing in canonical space. We model geometry with a geometry
    hash grid $\gamma_g$ and MLP $f_g$, while also modeling volumetric radiance
    and material with an appearance grid $\gamma_c$ and two additional MLPs
    $f_{rf}$, $f_m$. We supervise both $C_{rf}$ and $C_{pbr}$ using a L1 loss \wrt\
    the input image.} \label{fig:framework_intrinsic_avatar}
\end{figure*}
In this section, we first introduce basic concepts of neural radiance fields
(NeRF~\cite{Mildenhall2020ECCV}). Then we describe our framework of geometry
reconstruction of clothed avatars from monocular videos. The clothed avatars are
modeled as an articulated NeRF with SDF as its geometry representation. Next, we
introduce the volumetric scattering process from computer graphics and draw a
connection between it and NeRF. Finally, we describe our solution to secondary
ray tracing of volumetric scattering, which combines the explicit ray-marching
with iso-surface search and body articulation. The final outputs are
intrinsic properties of clothed avatars including geometry, material, albedo,
and lighting.

\subsection{Background: Neural Radiance Fields}
\label{sec:nerf}
Given a ray $\mathbf{r} = ( \mathbf{o}, \mathbf{d} )$ defined by its camera
center $\mathbf{o}$ and viewing direction $\mathbf{d}$, NeRF computes the output
radiance (\ie\ pixel color) of the ray via:
\begin{align}
\label{eqn:nerf_exact}
  C_{rf}(\mathbf{r}) =& \int_{t_n}^{t_f} T(t_n, t) \sigma_t (\mathbf{r}(t)) L (\mathbf{r}(t), -\mathbf{d}) dt \\
  \text{s.t} \quad & \mathbf{r} (t) = \mathbf{o} + t \mathbf{d} \nonumber \\
  & T(t_n, t) = \exp \left( - \int_{t_n}^t \sigma_t(\mathbf{r}(s)) ds \right ) \nonumber
\end{align}
where $(t_n, t_f)$ defines the near/far point for the ray integral. In practice, NeRF
uses a ray marching algorithm to approximate the exact value of the integral:
\begin{align}
\label{eqn:nerf_approx}
  C_{rf}(\mathbf{r}) \approx & \sum_{i=1}^{N} w^{(i)} L (\mathbf{r}(t^{(i)}), -\mathbf{d}) \\
  \text{s.t} \quad & \mathbf{r} (t) = \mathbf{o} + t \mathbf{d} \nonumber \\
  & w^{(i)} = T^{(i)} \left(1 - \exp(-\sigma_t(\mathbf{r}(t^{(i)})) \delta^{i} \right) \nonumber \\
  & T^{(i)} = \exp \left( -\sum_{j < i} \sigma_t(\mathbf{r}(t^{(j)})) \delta^{(j)} \right) \nonumber \\
  & \delta^{(i)} = t^{(i+1)} - t^{(i)} \nonumber
\end{align}
where $\{ t^{(1)}, \cdots, t^{(N)} \}$ are a set of sampled offsets on the ray.
$\sigma_t(\cdot)$ and $L(\cdot, \cdot)$ are represented as either neural
networks~\cite{Mildenhall2020ECCV,Oechsle2021ICCV,Wang2021NEURIPSa,Yariv2021NEURIPS}, explicit
grid data~\cite{Sun2022CVPR,AlexYuandSaraFridovich-Keil2022CVPR}, or a hybrid of
both~\cite{Garbin2021ICCV,Reiser2021ICCV,Mueller2022SIGGRAPH,Chen2022ECCV,Chen2023ARXIV,Chen2023TOG,Chen2023ICCV}.

\subsection{Clothed Humans Avatars as Articulated Neural Radiance Fields} We
follow the recent approaches of modeling humans as articulated
NeRF~\cite{Weng2022CVPR,Wang2022ECCV,Li2022ECCV,Jiang2023CVPR}. We assume body
articulations are based on the SMPL model~\cite{Loper2015SIGGRAPH}.  Following
previous works, we define the observation space as a space where the human is
observed, and the canonical space as a space where the human is in a canonical
pose. We apply inverse linear blend skinning (LBS) to transform 3D points in the
observation space $\mathbf{x}_o = \mathbf{r} (t)$ to the point in canonical
space $\mathbf{x}_c$. We model the radiance field, materials, and albedo all in the
canonical space.

\boldparagraph{Articulation via Inverse LBS} In the SMPL model, the linear
blender skinning (LBS) function is defined as:
\begin{align}
\label{eqn:LBS}
    \mathbf{x}_o = \text{LBS} (\mathbf{x}_c, \{ \mathbf{B}_b \}_{b=1}^{B}, w (\mathbf{x}_c)) = \left( \sum_{b=1}^{B} w (\mathbf{x}_c)_b \mathbf{B}_b \right) \mathbf{x}_c
\end{align}
where $\{ \mathbf{B}_b \}_{b=1}^{B}$ are the rigid bone-transformations defined
by estimated SMPL parameters. $w (\mathbf{x}_c)$ are skinning weights of point
$\mathbf{x}_c$. We use Fast-SNARF~\cite{Chen2023PAMI} to model the canonical
skinning weight function $w (\cdot)$ and the inverse skinning function:
\begin{align}
\label{eqn:inverse_LBS}
    \mathbf{x}_c = \text{LBS}^{-1} (\mathbf{x}_o, \{ \mathbf{B}_b \}_{b=1}^{B}, w (\mathbf{x}_c))
\end{align}
For simplicity, we drop the dependency on $\{ \mathbf{B}_b \}_{b=1}^{B}$ and $w
(\mathbf{x}_c)$  for the remainder of the paper.

\boldparagraph{Geometry} We use iNGP~\cite{Mueller2022SIGGRAPH} with SDF to
represent the underlying canonical shape of clothed humans. Specifically, given
a query point $\mathbf{x}_c$ in canonical space, we predict the SDF value of the
point and a latent feature $\mathbf{z}$:
\begin{align}
\label{eqn:iNGP_sdf}
    \left( \text{SDF}(\mathbf{x}_c), \mathbf{z} \right) = f_g (\gamma_g(\mathbf{x}_c))
\end{align}
where $\gamma_g(\cdot)$ is the iNGP hash grid feature of the input point, and
$f_g$ is a small MLP with a width of $64$ and one hidden layer. We use
VolSDF~\cite{Yariv2021NEURIPS} to convert from SDF to density $\sigma_t$.

\boldparagraph{Radiance and Material} 
Radiance and materials are predicted as follows:
\begin{align}
\label{eqn:radiance_and_materials}
  L (\mathbf{x}_c, \mathbf{d}) &= f_{rf} (\gamma_c (\mathbf{x}_c), \mathbf{z}, \text{ref}(\mathbf{d}, \mathbf{n}), \mathbf{n}) \\
  \alpha (\mathbf{x}_c), r (\mathbf{x}_c), m (\mathbf{x}_c) &= f_m (\gamma_c (\mathbf{x}_c), \mathbf{z})
\end{align}
where $\gamma_c(\cdot)$ is the feature from another iNGP hash grid designed
specifically for radiance and material prediction.  The same strategy was also
employed in ~\cite{Alexandru2023CVPR} for learning better geometric details.
$f_{rf}$ and $f_m$ are both MLPs with a width of $64$ and two hidden layers.
$\mathbf{n}$ is the analytical normal obtained from SDF fields.
$\text{ref}(\mathbf{d}, \mathbf{n})$ reflects the viewing direction $\mathbf{d}$
around the normal $\mathbf{n}$, similar to~\cite{Verbin2022CVPR}.  $L(\cdot,
\cdot)$ will be used for Eq.~\eqref{eqn:nerf_approx} whereas $\alpha$, $r$, and
$m$ are spatially varying \textit{albedo}, \textit{roughness}, and
\textit{metallic} parameters that will be used for physically based rendering.

For ray marching, we use 128 uniform samples and do two rounds of importance
sampling, each time with 16 samples, to obtain a final set of 160 samples per
ray.

With the aforementioned model, we can quickly reconstruct the detailed geometry of
clothed human avatars from a single monocular video in less than 30 minutes. 

\subsection{Physically Based Inverse Rendering via Volumetric Scattering}
\label{sec:volumetric_light_scattering}
With initial geometry and radiance estimation from previous sections,
we now account intrinsic properties of clothed human avatars, \ie\ material,
albedo, and lighting conditions for the rendering process.

With the standard equation of transfer of participating media in computer
graphics~\cite{Jakob2013Thesis,Pharr2023Booka}, we reach the NeRF formula Eq.~\eqref{eqn:nerf_exact} by assuming
 all the radiance that reaches the camera is modeled by neural networks.
On the other hand, if we think all the radiance that reaches the camera is
\textit{scattered} from some light sources (\eg\ environment maps) by a
volume of media, while the media itself does not emit any radiance, then
we are tackling the volume scattering problem.

Formally, we have the following integral to compute the radiance
scattered by the volume representing the human body along a certain camera ray
$(\mathbf{o}, \mathbf{d})$:
\begin{align}
\label{eqn:vls_exact}
  C_{pbr}(\mathbf{r}) =& \int_{t_n}^{t_f} T(t_n, t) \sigma_s (\mathbf{r}(t)) L_s (\mathbf{r}(t), -\mathbf{d}) dt \\
  \text{s.t} \quad & \mathbf{r} (t) = \mathbf{o} + t \mathbf{d} \nonumber \\
  & T(t_n, t) = \exp \left( - \int_{t_n}^t \sigma_t(\mathbf{r}(s)) ds \right ) \nonumber \\
    & L_s(\mathbf{x}, -\mathbf{d}) = \int_{S^2} f_p (\mathbf{x}, -\mathbf{d}, \bar{\mathbf{d}}) L_i (\mathbf{x}, -\bar{\mathbf{d}}) d \bar{\mathbf{d}} \nonumber \\
  & \sigma_t(\mathbf{r}(t)) = \sigma_a(\mathbf{r}(t)) + \sigma_s(\mathbf{r}(t)) \nonumber
\end{align}
$S^2$ is the domain of a unit sphere. $\sigma_s$ and $\sigma_a$ are the
\textit{scattering} coefficient and the \textit{absortion} coefficient,
respectively. Their sum is the \textit{attentuation} coefficient, which is also
known as the \textit{density} in NeRF literature. $f_p (\mathbf{x},
-\mathbf{d}, \bar{\mathbf{d}})$ is the \textit{phase function} that describes
the probability of light scattering from direction $\bar{\mathbf{d}}$ to
$-\mathbf{d}$ at point $\mathbf{x}$.  $L_i(\mathbf{x}, -\bar{\mathbf{d}})$ is
the incoming radiance towards point $\mathbf{x}$ along the direction
$-\bar{\mathbf{d}}$, it can be computed as a weighted sum of $C_{rf}
(\mathbf{x}, \bar{\mathbf{d}})$ (Eq.~\eqref{eqn:nerf_exact}) and radiance
from an environment map $\text{Env} (\bar{\mathbf{d}})$:
\begin{align}
\label{eqn:incoming_radiance}
  L_i(\mathbf{x}, -\bar{\mathbf{d}}) = & C_{rf} (\mathbf{x}, \bar{\mathbf{d}}) \nonumber \\
  & + \exp \left( - \int_{t_{n'}}^{t_{f'}} \sigma_t(\mathbf{x} + s \bar{\mathbf{d}}) ds \right) \text{Env} (\bar{\mathbf{d}})
\end{align}
where $t_{n'}$ and $t_{f'}$ are the near and far points of secondary rays.  In
traditional physically based rendering, the first term represents indirect
illumination while the second term represents direct illumination. Instead of
modeling indirect illumination with path tracing, we use the trained radiance
field to approximate it. This is also done in various recent
works~\cite{Jin2023CVPR,Zhang2022CVPRa} for modeling static scenes from
multi-view input images.  For Monte-Carlo estimation of $C_{pbr}(\mathbf{r})$,
we will have to sample the two integrals $\int_{t_n}^{t_f}$ and $\int_{S^2}$
separately. The first integral is estimated via quadrature as was done in
standard NeRF rendering.  We next describe how to sample the second integrals.

For approximating Eq.~\eqref{eqn:vls_exact}, we importance sample offsets $\{
    \bar{t}^{(1)}, \cdots, \bar{t}^{(M)} \}$ from the PDF estimated by radiance
field samples that have been used to estimate Eq.~\eqref{eqn:nerf_approx} . The
approximated Eq.~\eqref{eqn:vls_exact} becomes:
\begin{align}
\label{eqn:vls_approx}
  C_{pbr}(\mathbf{r}) \approx & \sum_{i=1}^M w^{(i)} \frac{\sigma_s (\mathbf{r}(\bar{t}^{(i)}))}{\sigma_t(\mathbf{r}(\bar{t}^{(i)}))} L_s (\mathbf{r}(\bar{t}^{(i)}), -\mathbf{d}) \\
  \text{s.t} \quad & \mathbf{r} (t) = \mathbf{o} + t \mathbf{d} \nonumber \\
  & w^{(i)} = T^{(i)} \left(1 - \exp(-\sigma_t(\mathbf{r}(\bar{t}^{(i)}) \delta^{(i)} ) \right) \nonumber \\
  & T^{(i)} = \exp \left( -\sum_{j < i} \sigma_t(\mathbf{r}(\bar{t}^{(j)})) \delta^{(j)} \right) \nonumber \\
    & L_s(\mathbf{r}(\bar{t}^{(i)}), -\mathbf{d}) = \frac{f_p (\mathbf{r}(\bar{t}^{(i)}), -\mathbf{d}, \bar{\mathbf{d}}^{(i)})}{\text{pdf} (\bar{\mathbf{d}}^{(i)})} \nonumber \\
    & \qquad \qquad \qquad \qquad \cdot L_i (\mathbf{r}(\bar{t}^{(i)}), -\bar{\mathbf{d}}^{(i)}) \nonumber \\
  & \sigma_t(\mathbf{r}(t)) = \sigma_a(\mathbf{r}(t)) + \sigma_s(\mathbf{r}(t)) \nonumber
\end{align}
%
\begin{figure}
    \includegraphics[width=0.45\textwidth]{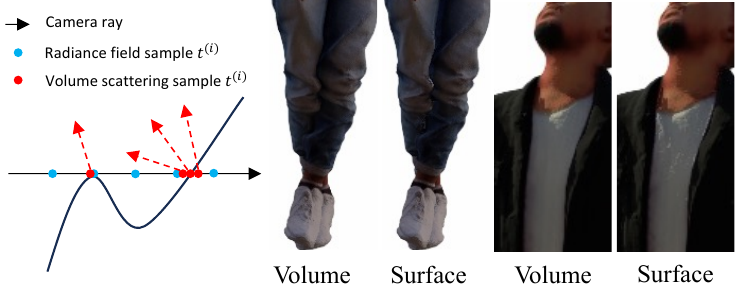}
    \caption{\textbf{Illustration of Volumetric Scattering.} Volumetric
    scattering can blend between multiple surfaces when a ray crosses edges
    (left).  This results in smooth transitions of appearance at boundaries,
    avoiding noisy shadow (middle) and lighting (right) at these locations.
    }
    \vspace{-0.3cm}
    \label{fig:importance_sampling}
\end{figure}
in which $\frac{\sigma_s
(\mathbf{r}(\bar{t}^{(i)}))}{\sigma_t(\mathbf{r}(\bar{t}^{(i)}))}$ corresponds
to the spatially varying \textit{albedo} and is analogous to that in
surface-based rendering. $\text{pdf} (\bar{\mathbf{d}}^{(i)})$ is the PDF from
which $\bar{\mathbf{d}}^{(i)}$ is sampled.

Essentially, we use quadrature to estimate the first integral
$\int_{t_n}^{t_f}$, and Monte-Carlo sampling to estimate the second integral
$\int_{S^2}$, all together with $M$ samples.  We refer readers to \iftoggle{arXiv}{Appendix~\ref{appx:volume_scattering_derivation}}{the Supp. Mat.} for a detailed derivation of Eq.~\eqref{eqn:vls_approx}. During training
$\bar{\mathbf{d}}^{(i)}$ is uniformly sampled from the unit sphere with $M=512$
and stratified jittering~\cite{Pharr2023Bookb}.  For relighting, we use light
importance sampling with $M=1024$ to sample from a known environment map.

We note that when using an SDF-density representation, most of the samples
$\bar{t}^{(i)}$ are concentrated around the surface of the human body. This
makes the volumetric scattering process similar to a surface-based rendering
process when there is a clear intersection between the ray and the surface.
On the other hand, rays at edges and boundaries may not have a well-defined
surface as the corresponding pixels may cover multiple surfaces. For these rays,
it would be difficult to employ surface-based rendering while volume scattering
suits naturally for this case (Fig.~\ref{fig:importance_sampling}).

We use a simplified version of Disney BRDF~\cite{Burley2012SIGGRAPH} to model the
combined effect of the volumetric albedo $\frac{\sigma_s
(\mathbf{r}(\bar{t}^{(i)}))}{\sigma_t(\mathbf{r}(\bar{t}^{(i)}))}$ and the phase function $f_p$. It
takes predicted albedo $\alpha$, roughness $r$ and metallic $m$ as inputs:
\begin{align}
    &\frac{\sigma_s}{\sigma_t} f_p (\mathbf{\omega}_o, \mathbf{\omega}_i) = \text{BRDF} (\mathbf{\omega}_o, \mathbf{\omega}_i, \alpha, r, m, \mathbf{n}) \max\left( \mathbf{n} \cdot \mathbf{\omega}_i, 0 \right) \nonumber
\end{align}
We drop dependency on spatial locations for brevity.  An extended implementation
detail of the above BRDF can be found in
\iftoggle{arXiv}{Appendix~\ref{appx:brdf_definition}.}{the Supp. Mat.} More
physically accurate phase functions for rendering surface-like volumes, such as
SGGX~\cite{Heitz2015SIGGRAPH} can also be plugged into our formulation, but we
empirically do not find them providing any advantage for our application.

\subsection{Articulated Secondary Ray Tracing}
\label{sec:secondary_ray_tracing}
Given the $M$ samples $\{ \bar{t}^{(i)} \}_{i=1}^{M}$ on a primary ray, we trace
one secondary ray for each of the samples, and compute opacity (or visibility in
a surface rendering setup) and radiance for each secondary ray.  Formally, we
trace a secondary ray $\bar{\mathbf{r}}^{(i)}$ from the corresponding sample,
where $\bar{\mathbf{r}}^{(i)} = (\bar{\mathbf{o}}^{(i)},
\bar{\mathbf{d}}^{(i)})$ with $\bar{\mathbf{o}}^{(i)} =
\mathbf{r}(\bar{t}^{(i)})$.

\boldparagraph{Secondary Ray Tracing} We note that traditional sphere tracing
could lead to non-convergence rays when the SDF is not smooth. This is
exacerbated when the SDF is approximated by neural networks and combined with
body articulation.  Furthermore, the sequential evaluation of SDF values on a
ray is not amenable to parallelization, especially when a large number of
secondary rays need to be evaluated and each evaluation involves neural
networks.

Given the underlying NeRF representation, precise surface location is often not
required to compute radiance, while the opacity is binary most of the time due
to the SDF-density representation.  This motivates us to use ray marching to
compute secondary shading effects.  However, we observe that the Laplace
density function of~\cite{Yariv2021NEURIPS} tends to assign non-negligible
density values to small positive SDF values. This will cause the secondary ray
marching to give non-zero weights to points that are very close to the surface,
\ie\ starting points $\bar{\mathbf{o}}$'s of secondary rays.  While
NeuS~\cite{Wang2021NEURIPSa} is more well-behaved as it only assigns high
weights for SDF zero-crossing intervals, estimating weights of ray segments
requires estimation of analytical surface normals, which usually doubles the
computation cost of ray marching.

Motivated by these facts, we propose a hybrid approach to secondary ray
marching by searching for the first SDF zero-crossing point of a set of uniform
samples on the secondary ray and only start accumulating importance weights
from that point. Given the weights of uniform samples, we sample 4 additional
samples on the secondary ray and compute the transmittance and radiance from
these 4 samples. The computed transmittance and radiance are inputs to incoming
radiance evaluation Eq.~\eqref{eqn:incoming_radiance}. 

Formally, given a secondary ray $\bar{\mathbf{r}}$, we first uniformly sample
64 offsets $\{ t'^{(1)}, \cdots, t'^{(64)} \}$ on the ray between the near and
far points, $t_{n'} = 0$ $t_{f'} = 1.5$. Each of the sampled offsets is
transformed to the canonical space to query its SDF value:
\begin{align}
\label{eqn:secondary_ray_sdf}
    \text{SDF}(\bar{\mathbf{r}}(t')) = f_g (\gamma_g(\text{LBS}^{-1} (\bar{\mathbf{r}}(t'))))
\end{align}
Alg.~\ref{alg:zero_crossing} describes the procedure of searching for the first
zero-crossing point and accumulating weights for each of the points.
\begin{algorithm}[t]
\caption{Zero-Crossing Search and Importance Weight Accumulation}
\label{alg:zero_crossing}
\begin{algorithmic}[1]
\REQUIRE $\{ \text{SDF}(\bar{\mathbf{r}}(t'^{(i)})) \}_{i=1}^{64}$, $\bar{\mathbf{r}} = (\bar{\mathbf{o}}, \bar{\mathbf{d}})$
\ENSURE Importance weights $\{ w^{(i)} \}_{i=1}^{63}$
\STATE $s \leftarrow 1$
\STATE $\{ w^{(i)} \}_{i=1}^{63} \leftarrow \mathbf{0}$
\WHILE{$s < 63$}
\IF{$\text{SDF}(\bar{\mathbf{r}}(t'^{(s)})) \cdot \text{SDF}(\bar{\mathbf{r}}(t'^{(s+1)})) < 0$}
\STATE \textbf{break}
\ENDIF
\STATE $s \leftarrow s + 1$
\ENDWHILE
\STATE $T (\bar{\mathbf{r}}) \leftarrow 1$
\FOR{$i = s$ \TO $63$}
\STATE $\delta^{(i)} \leftarrow t'^{(i+1)} - t'^{(i)}$
\STATE $w^{(i)} \leftarrow \left(1 - \exp(-\sigma_t(\bar{\mathbf{r}}(t'^{(i)})) \delta^{(i)}) \right) T (\bar{\mathbf{r}})$
\STATE $T (\bar{\mathbf{r}}) \leftarrow T (\bar{\mathbf{r}}) \exp \left( - \sigma_t(\bar{\mathbf{r}}(t'^{(i)})) \delta^{(i)} \right)$
\ENDFOR
\RETURN $\{ w^{(i)} \}_{i=1}^{63}$
\end{algorithmic}
\end{algorithm}
This is similar to the traditional sphere tracing algorithm, with the difference
that SDF values are evaluated uniformly in parallel instead of sequentially. We
parallelize Alg.~\ref{alg:zero_crossing} together with importance sampling over
rays with custom CUDA implementation. 

\subsection{Training Details}
We use standard L1 loss \wrt\ input images on radiance predicted by both
radiance field (RF loss) and volumetric scattering (PBR loss). We apply
eikonal loss~\cite{Gropp2020ICML} (throughout training) and curvature
loss~\cite{Alexandru2023CVPR} (only the first half of the training) to regularize
the SDF field. We also apply Lipschitz
regularization~\cite{Alexandru2023CVPR} and standard smoothness
regularization~\cite{Zhang2021SIGGRAPHASIA,Jin2023CVPR} to the material
predictions.  Details on losses and hyperparameters can be found in \iftoggle{arXiv}{Appendix~\ref{appx:loss_function}.}{the Supp. Mat.}

We train a total of 25k iterations with a learning rate of $0.001$ decayed by a
factor of $0.3$ at $12.5$k, $18.75$k, $22.5k$, and $23.75$k iterations,
respectively. The first 10k iterations are trained with the RF loss only,
while the rest of the iterations are trained on both the RF loss and
the PBR loss.  We use a batch size of $4096$ rays. Training is done
on a single NVIDIA RTX 3090 GPU in 4 hours.

\section{Experimental Evaluation}
\label{sec:experiments}
\subsection{Datasets}
We utilize 3 different datasets to conduct our experiments
\begin{itemize}
    \item \textbf{RANA~\cite{Iqbal2023ICCV}} To quantitatively evaluate our
    estimation of the physical properties of the reconstructed avatar, we use 8
    subjects from the RANA dataset.  The dataset is rendered using a standard path
    tracing algorithm, with ground truth albedo, normal, and relighted images
    available for evaluation. We follow protocol A in which the training set
    resembles a person holding an A-pose rotating in front of the camera under
    unknown illumination.  The test set consists of images of the same subject in
    random poses under novel illumination conditions.
    \item \textbf{PeopleSnapshot~\cite{Alldieck2018CVPR}} In PeopleSnapshot,
    subjects always hold a simple A-pose and rotate in front of the camera under
    natural illumination. We use 6 subjects from the dataset with refined pose
    estimation from~\cite{Chen2021ARXIVb,Jiang2023CVPR}.
    \item \textbf{SyntheticHuman-Relit} To additionally evaluate relighting on
        more complex training poses of continuous videos, we create a synthetic
        dataset by rendering two subjects from the SyntheticHuman
        dataset~\cite{Peng2024PAMI} with Blender under different illumination
        conditions. Due to space limits, we refer readers to the
        \iftoggle{arXiv}{Appendix~\ref{appx:dataset}
        and~\ref{appx:additional_quantitative_results}}{Supp. Mat.} for details
        and results on this dataset.
\end{itemize}

\subsection{Baselines}
To our knowledge, Relighting 4D (R4D~\cite{Chen2022ECCVa}) is the only baseline
with publicly available code for the physically based inversed rendering of
clothed human avatars under unknown illumination, without pretraining on any
ground truth geometry/albedo/materials. RANA~\cite{Iqbal2023ICCV} only provides
public access to their data at the time of our submission. Furthermore, RANA
pretrains on ground truth albedo, which is not available in our setting.

We note that the original R4D implementation does not employ any mask loss.  We
therefore also report a variant of R4D (denoted as R4D*) that employs a mask
loss. R4D* achieves overall better performance than R4D
(Tab.~\ref{tab:quantitative}) and thus we primarily compared our method to this
improved version of R4D.

\subsection{Evaluation Metrics}
On synthetic datasets, we evaluate the following metrics:
\begin{itemize}
    \item \textbf{Albedo PNSR/SSIM/LPIPS} we evaluate the standard image quality
    metrics on albedos rendered under training views. Due to ambiguity in
    estimating albedo and light intensity, we follow the practice
    of~\cite{Zhang2021SIGGRAPHASIA} to align the predicted albedo with the
    ground truth albedo. More details can be found in \iftoggle{arXiv}{Appendix~\ref{appx:albedo_evaluation}.}{the Supp. Mat.}
    \item \textbf{Normal Error} this metric evaluates normal estimation error
    (in degrees) between predicted normal images and the ground-truth normal
    images.
    \item \textbf{Relighting PSNR/SSIM/LPIPS} we also evaluate image quality
    metrics on images synthesized on novel poses with novel illumination.
    Relighting evaluation on training poses (\ie\ SyntheticHuman-Relit dataset)
    is reported in \iftoggle{arXiv}{Appendix~\ref{appx:additional_quantitative_results}.}{the Supp. Mat.}
\end{itemize}

On real-world datasets, \ie\ PeopleSnapshot, we primarily present qualitative
results including novel view/pose synthesis under novel illuminations.

\subsection{Comparison to Baselines}
\begin{figure*}
\includegraphics[width=\linewidth]{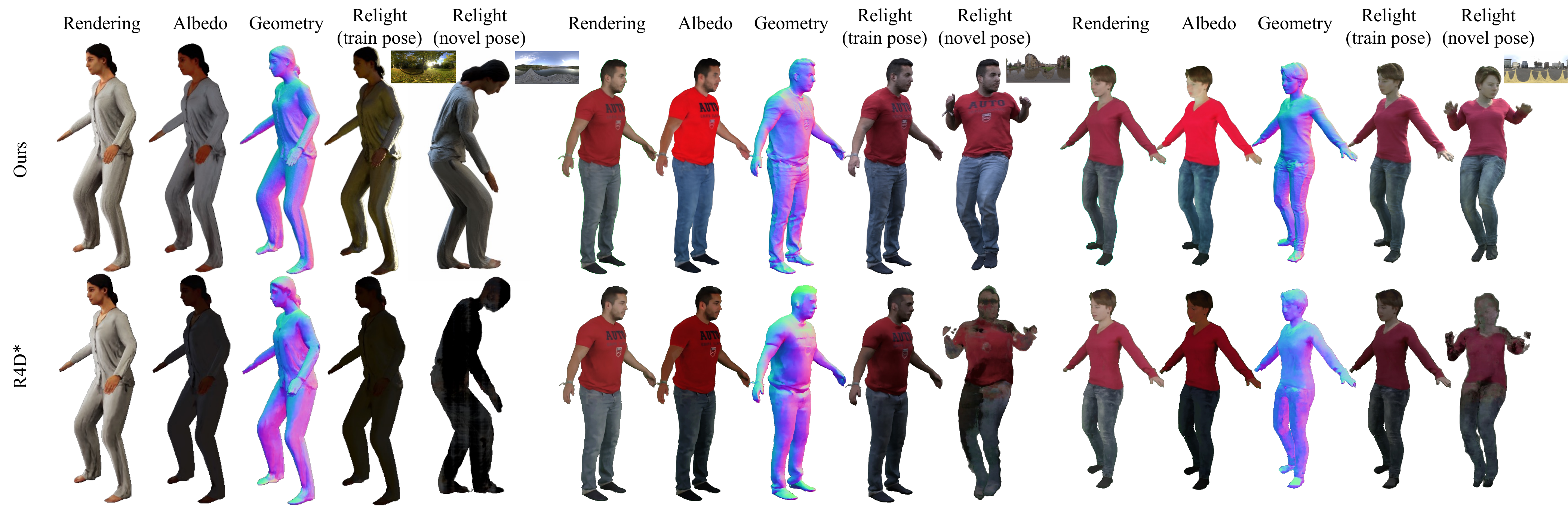}
\caption{\textbf{Qualitative comparison to the baseline.} We show the results
of our method and R4D* on both synthetic (left) and real
(middle, right) datasets. As indicated, R4D* struggles to recover intrinsic
properties of avatars and do not produce realistic relighting results.
Furthermore, it fails to generalize to novel poses. Our method produces
high-quality results on both synthetic and real datasets, while generalizing
well to novel poses and illuminations. More qualitative results can be found in
\iftoggle{arXiv}{Appendix~\ref{appx:additional_qualitative_results}}{the Supp. Mat.}}
\label{fig:qualitative}
\end{figure*}
\begin{table*}[t]
    \centering
    \begin{tabular}{l|c|c|c|c|c|c|c|c|}
        \toprule
        \multirow{2}{*}{Method} & \multicolumn{3}{c|}{Albedo} & \multirow{1}{*}{Normal} & \multicolumn{3}{c}{Relighting (Novel Pose)} \\
        \cmidrule{2-4} \cmidrule{6-8}
        & PSNR $\uparrow$ & SSIM $\uparrow$ & LPIPS $\downarrow$ & \multicolumn{1}{c|}{Error $\downarrow$} & PSNR $\uparrow$ & SSIM $\uparrow$ & LPIPS $\downarrow$ \\
        \midrule
        R4D  & 18.24 & 0.7780 & 0.2414 & 42.69 \degree & 14.37 & 0.8133 & 0.2017 \\
        R4D*  & 18.23 & 0.8254 & 0.2043 & 27.38 \degree & 16.62 & 0.8370 & 0.1726 \\
        Ours & \textbf{22.83} & \textbf{0.8816} & \textbf{0.1617} &
        \textbf{9.96} \degree & \textbf{18.18} & \textbf{0.8722} & \textbf{0.1279} \\
        \bottomrule
    \end{tabular}
    \caption{\textbf{Quantitative comparison to the baseline on the RANA dataset.}}
    \label{tab:quantitative}
\end{table*}
We present the average metrics on the RANA dataset in
Tab.~\ref{tab:quantitative}. Our method significantly outperforms R4D and R4D*
on all metrics, achieving 77\% and 64\% reduction in the normal estimation
error, respectively. This combined with our explicit ray tracing technique also
gives us a significant improvement in albedo-related metrics on training poses.

For relighting novel poses, we note that the SMPL model is not perfectly
aligned with images in the RANA dataset, which could make the PSNR metric less
meaningful.  Thus we argue that SSIM and LPIPS can better reflect the quality of
the relighting results.  Nevertheless, R4D* fails to produce reasonable results
due to its inability to generalize to novel poses.  On the other hand, our
method can produce high-quality re-posing and relighting results (Fig.~\ref{fig:qualitative}).

\subsection{Ablation Study}
\label{sec:ablation_study}
\begin{figure}[t]
    \centering
    \includegraphics[width=\linewidth]{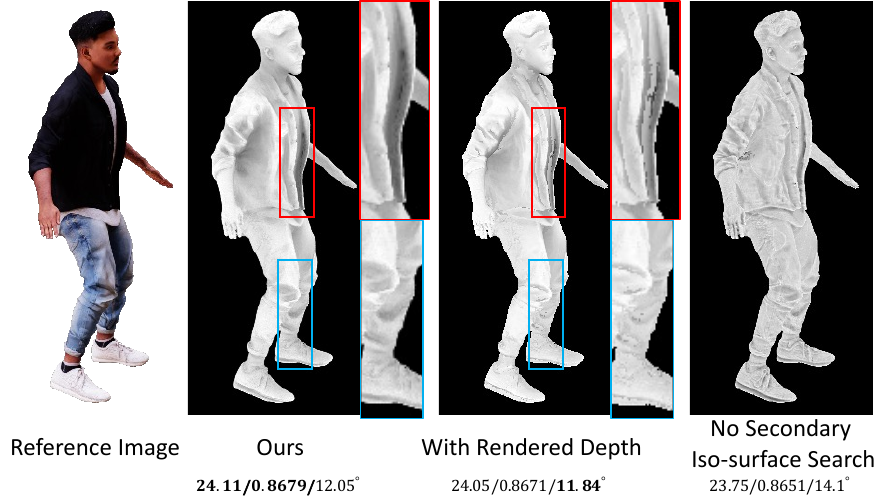}
    \caption{\textbf{Ablation study.} We visualize average visibility (AV) maps
    of each variant and report albedo PSNR ($\uparrow$)/albedo SSIM
    ($\uparrow$)/Normal Error ($\downarrow$). Surface scattering with rendered
    depth results in discontinuities at boundaries and edges. Without our
    proposed iso-surface search for secondary ray tracing, the visibility map
    is much darker and does not reflect true visibility. We also refer readers
    to Fig.~\ref{fig:importance_sampling} for qualitative relighting results}
    \label{fig:ablation}
\end{figure}
In this section, we ablate several of our design choices. We use subject 01 from
the RANA dataset for this ablation study. We visualize average visibility (AV)
maps which best reflect the quality of the reconstruction geometry and secondary
ray tracing.  The AV value of a primary ray $\mathbf{r}$ is defined as:
\begin{equation}
    \text{AV} (\mathbf{r}) = 2 * \frac{1}{M} \sum_{i=1}^M V(\bar{\mathbf{r}}^{(i)})
\end{equation}
where $V(\bar{\mathbf{r}}_i)$ is the visibility of the $i$-th secondary ray (1
for not occluded, 0 for occluded), and $M$ is the number of secondary rays
sampled for each primary ray. We multiply visibility by 2 as we sample
secondary rays on a unit sphere instead of a hemisphere. The results are
summarized in Fig.~\ref{fig:ablation}. We describe different variants
in the following:
\begin{itemize}
    \item \textbf{Ours:} Our full method with all the components described in
    Section~\ref{sec:method}.
    \item \textbf{Rendered Depth with Surface Scattering:} This variant corresponds
    to~\cite{Jin2023CVPR} that uses rendered depth and surface scattering.
    \item \textbf{No Iso-surface Search for Secondary Ray Tracing:} In this
    variant we do not perform the iso-surface search for secondary ray tracing
    (Sec.~\ref{sec:secondary_ray_tracing}) and start accumulating weights from the
    first sample of the 64 samples on the secondary ray.
\end{itemize}

\section{Conclusion}
\label{sec:conclusion}
We have presented a novel approach to the inverse rendering of dynamic humans
from only monocular videos.  Our method can achieve high-quality
reconstruction of clothed human avatars with disentangled geometry, albedo,
material, and environment lighting from only monocular videos.  We have also
shown that our learned avatars can be rendered realistically under novel
lighting conditions \textit{and} novel poses. Experiment results show that our
method significantly outperforms the state-of-the-art method both qualitatively
and quantitatively.  We discuss limitations and future work in \iftoggle{arXiv}{Appendix~\ref{appx:limitations}.}{the Supp. Mat.}

\section{Acknowledgment}
\label{sec:acknowledgment}
We thank Umar Iqbal and Zhen Xu for the helpful discussion on setting up
synthetic datasets. We thank Zijian Dong and Naama Pearl for constructive
feedback on our early draft. SW, BA and AG were supported by the ERC Starting
Grant LEGO-3D (850533) and the DFG EXC number 2064/1 - project number 390727645.
SW and ST acknowledge the SNSF grant 200021 204840.
{
    \small
    \bibliographystyle{ieeenat_fullname}
    \bibliography{main,bibliography,bibliography_custom}
}
\appendix
\onecolumn
\numberwithin{equation}{section}
\setcounter{equation}{0}
\numberwithin{figure}{section}
\setcounter{figure}{0}
\numberwithin{table}{section}
\setcounter{table}{0}

\section{Volume Scattering Derivation}
\label{appx:volume_scattering_derivation}
In this section, we derive the volume scattering approximation equation
(Eq.~\iftoggle{arXiv}{\eqref{eqn:vls_approx}}{(\textcolor{red}{10})} in the main paper) from the equation of
transfer~\cite{Pharr2023Booka}.  The general equation of transfer accounting for
both volume emission and volume scattering is as follows:
\begin{align}
  \label{eq:equation_of_transfer}
  C_{pbr} (\mathbf{r}) &= \int_{t_n}^{t_f} T(t_n, t) \sigma_t (\mathbf{r}(t)) L (\mathbf{r}(t), -\mathbf{d}) dt \\
  \text{s.t} \quad & \mathbf{r} (t) = \mathbf{o} + t \mathbf{d} \nonumber \\
  & T(t_n, t) = \exp \left( - \int_{t_n}^t \sigma_t(\mathbf{r}(s)) ds \right) \nonumber \\
  & \sigma_t(\mathbf{r}(t)) = \sigma_a(\mathbf{r}(t)) + \sigma_s(\mathbf{r}(t)) \nonumber
\end{align}
As defined in the main paper, $\sigma_s$ and $\sigma_a$ are the
\textit{scattering} coefficient and the \textit{absorption} coefficient,
respectively. They define the probability of light being scattered/absorbed by
the participating media per unit length.  $\sigma_t$ is the \textit{attenuation}
coefficient, which is the sum of $\sigma_s$ and $\sigma_a$. Physically, it
describes the probability of light being either out-scattered or absorbed per
unit length, both of which will reduce the amount of radiance that reaches the
camera.  We refer readers to~\cite{Pharr2023Bookc} for detailed explaination on
physical meanings of these parameters. With some abuse of notation, we define
$L$ as the radiance accounting for both volume emission and volume scattering:
\begin{align}
    L(\mathbf{r}(t), -\mathbf{d}) &= \frac{\sigma_a (\mathbf{r}(t))}{\sigma_t (\mathbf{r}(t))} L_e(\mathbf{r}(t), -\mathbf{d}) + \frac{\sigma_s (\mathbf{r}(t))}{\sigma_t (\mathbf{r}(t))} L_s(\mathbf{r}(t), -\mathbf{d}) \\
    \text{s.t.} \quad & L_s(\mathbf{x}, -\mathbf{d}) = \int_{S^2} f_p
    (\mathbf{x}, -\mathbf{d}, \bar{\mathbf{d}}) L_i (\mathbf{x}, -\bar{\mathbf{d}}) d \bar{\mathbf{d}} \nonumber
\end{align}
where $L_e$ is the volume emission radiance, $L_s$ is the volume scattering
radiance. Since we assume the scene (human body) does not emit energy itself,
$L_e$ should always be 0.  $L_i$ is the incoming radiance via either direct
illumination or indirect illumination, as described in Eq.~\iftoggle{arXiv}{\eqref{eqn:incoming_radiance}}{(\textcolor{red}{9})}
in the main paper.  $f_p (\mathbf{x}, -\mathbf{d}, \bar{\mathbf{d}})$ is the
\textit{phase function} that describes the probability of light scattering from
direction $\bar{\mathbf{d}}$ to $-\mathbf{d}$ at point $\mathbf{x}$. Given these
facts, Eq.~\eqref{eq:equation_of_transfer} can be re-written as:
\begin{align}
\label{eqn:vls_exact_app}
  C_{pbr}(\mathbf{r}) =& \int_{t_n}^{t_f} T(t_n, t) \sigma_s (\mathbf{r}(t)) L_s (\mathbf{r}(t), -\mathbf{d}) dt \\
  \text{s.t} \quad & \mathbf{r} (t) = \mathbf{o} + t \mathbf{d} \nonumber \\
  & T(t_n, t) = \exp \left( - \int_{t_n}^t \sigma_t(\mathbf{r}(s)) ds \right ) \nonumber \\
    & L_s(\mathbf{x}, -\mathbf{d}) = \int_{S^2} f_p (\mathbf{x}, -\mathbf{d}, \bar{\mathbf{d}}) L_i (\mathbf{x}, -\bar{\mathbf{d}}) d \bar{\mathbf{d}} \nonumber \\
  & \sigma_t(\mathbf{r}(t)) = \sigma_a(\mathbf{r}(t)) + \sigma_s(\mathbf{r}(t)) \nonumber
\end{align}
which corresponds to Eq.~\iftoggle{arXiv}{\eqref{eqn:vls_exact}}{(\textcolor{red}{8})} in the main paper.  We next
describe how to further approximate Eq.~\eqref{eqn:vls_exact_app} with discrete
samples.

The general idea is to sample offsets from the probability density function
(PDF) of $T(t_n, t)$ and approximate the integral with Monte-Carlo integration.
Define $\text{pdf}(t)$ as the PDF of $t$ from which we sample $M$ offsets
$\{ \bar{t}^{(i)} \}_{i=1}^M$, we have:
\begin{align}
\label{eqn:vls_approx_app}
    C_{pbr}(\mathbf{r}) \approx & \frac{1}{M} \sum_{i=1}^M \frac{ T(t_n, \bar{t}^{(i)}) \sigma_s (\mathbf{r}(\bar{t}^{(i)}))}{\text{pdf}( \bar{t}^{(i)} )} L_s (\mathbf{r}(\bar{t}^{(i)}), -\mathbf{d})
\end{align}
in the next two subsections, we describe how to sample from $\text{pdf}(t)$.

\subsection{Homogeneous Volume}
If we assume homogeneous volume, \ie\ $\sigma_t (\mathbf{r}(t)) = \sigma_t$,
then we can simplify $T(t_n, t)$ according to Beer's law:
\begin{align}
    \label{app:eqn:beer_law}
    T(t_n, t) = \exp \left( - \sigma_t | t - t_n | \right )
\end{align}
Sampling from $T(t_n, t)$ is equivalent to sampling from an exponential
distribution, where the PDF is given by:
\begin{align}
    \label{app:eqn:exp_pdf}
    \text{pdf}(t) = c \exp \left( - \sigma_t | t - t_n | \right )
\end{align}
where $c$ is a normalization constant.  The cumulative distribution function
(CDF) of $t$ should satisfy:
\begin{align}
    \label{app:eqn:exp_cdf_constant}
    \int_{t_n}^{\infty} c \exp \left( - \sigma_t | t - t_n | \right ) dt = -\frac{c}{\sigma_t} \exp \left( - \sigma_t | t - t_n | \right ) \Big|_{t_n}^{\infty} = \frac{c}{\sigma_t} = 1
\end{align}
thus $c = \sigma_t$ and we have the following PDF and CDF of $t$ accordingly:
\begin{align}
  \text{pdf}(t) =& \sigma_t \exp \left( - \sigma_t | t - t_n | \right ) \\
  P(t) =& 1 - \exp \left( - \sigma_t | t - t_n | \right ) \label{eqn:cdf_homo}
\end{align}
\subsection{Heterogeneous Volume}
If the homogeneous assumption is lifted, we can still approximate the integral
by dividing the ray segment $(t_n, t_f)$ into intervals and assuming $\sigma_t$
to be constant within each interval.

Formally, let us assume the ray segment is divided into $N - 1$ intervals, each
defined by $[t^{(1)}, t^{(2)}), \cdots [t^{(i)},
t^{(i+1)}), \cdots [t^{(N-1)}, t^{(N)} )$ with $t^{(1)}
= t_n, t^{(N)} = t_f$.  With our assumption on constant $\sigma_t$
inside each interval, \ie\ $\sigma_t (\mathbf{r}(t)) = \sigma_t
(\mathbf{r}(t^{(i)})), \forall t \in [t^{(i)}, t^{(i+1)})$,
define $\delta^{(i)} = | t^{(i+1)} - t^{(i)} |$, we define the
following:
\begin{align}
    T(t^{(i)}, t^{(i+1)}) =& \exp \left( - \sigma(\mathbf{r}(t^{(i)})) \delta^{(i)} \right), \forall i \in \{1, \cdots, N - 1 \} \nonumber \\
    T(t^{(1)}, t^{(i)}) =& \prod_{j < i} T(t^{(j)}, t^{(j+1)}) \nonumber \\
    T(t^{(1)}, t^{(1)}) =& 1 \nonumber
\end{align}
To obtain the exact PDF from which we sample $t$, we extend
Eq.~\eqref{app:eqn:beer_law} such that we sample from $T(t^{(1)}, t)$ that
contains a homogeneous part and a heterogeneous part:
\begin{align}
    T (t^{(1)}, t) =& T (t^{(i)}, t) T(t^{(1)}, t^{(i)}) \\
    \text{s.t.} \quad &t^{(i)} \leq t < t^{(i+1)} \nonumber
\end{align}
where $T(t^{(1)}, t^{(i)})$ is the accumulated transmittance from
the heterogeneous volume before $t^{(i)}$.  Similar to
Eq.~\eqref{app:eqn:exp_pdf} and Eq.~\eqref{app:eqn:exp_cdf_constant} we can
derive the normalization constant as $\sigma_t(\mathbf{r}(t)) = \sigma_t(\mathbf{r}(t^{(i)}))
$, thus the PDF of $t$ is:
\begin{align}
    \label{eqn:pdf_heter}
    \text{pdf}(t) =& \sigma_t(\mathbf{r}(t^{(i)})) T(t^{(i)}, t) T(t^{(1)}, t^{(i)}) \nonumber \\
    =& \sigma_t(\mathbf{r}(t)) T(t^{(1)}, t) \\
    \text{s.t.} \quad &t^{(i)} \leq t < t^{(i+1)} \nonumber
\end{align}

Plug Eq.~\eqref{eqn:pdf_heter} into Eq.~\eqref{eqn:vls_approx_app}, one will
note that the $T(t_n, t)$ term is in both the numerator and the denominator.
Thus Eq.~\eqref{eqn:vls_approx_app} simplifies to:
\begin{align}
\label{eqn:vls_approx_app_simplified}
    C_{pbr}(\mathbf{r}) \approx & \frac{1}{M} \sum_{i=1}^M \frac{\sigma_s (\mathbf{r}(\bar{t}^{(i)}))}{\sigma_t(\mathbf{r}(\bar{t}^{(i)}))} L_s (\mathbf{r}(\bar{t}^{(i)}), -\mathbf{d})
\end{align}
Since we define the combined effect of $\frac{\sigma_s (\mathbf{r}(\bar{t}^{(i)}))}{\sigma_t(\mathbf{r}(\bar{t}^{(i)}))}$ and the phase function as a BRDF function, which becomes unrelated to $\sigma_t$, while we also need to be able to differentiate \wrt the geometry represented by $\sigma_t$, we use quadrature weights $\{ w^{(i)} \}$ from NeRF~\cite{Mildenhall2020ECCV}, resulting in Eq.~\iftoggle{arXiv}{\eqref{eqn:vls_approx}}{(\textcolor{red}{10})} in the main paper:
\begin{align}
\label{eqn:vls_approx_final_app}
  C_{pbr}(\mathbf{r}) \approx & \sum_{i=1}^M w^{(i)} \frac{\sigma_s (\mathbf{r}(\bar{t}^{(i)}))}{\sigma_t(\mathbf{r}(\bar{t}^{(i)}))} L_s (\mathbf{r}(\bar{t}^{(i)}), -\mathbf{d}) \\
  \text{s.t} \quad & \mathbf{r} (t) = \mathbf{o} + t \mathbf{d} \nonumber \\
  & w^{(i)} = T^{(i)} \left(1 - \exp(-\sigma_t(\mathbf{r}(\bar{t}^{(i)}) \delta^{(i)} ) \right) \nonumber \\
  & T^{(i)} = \exp \left( -\sum_{j < i} \sigma_t(\mathbf{r}(\bar{t}^{(j)})) \delta^{(j)} \right) \nonumber \\
    & L_s(\mathbf{r}(\bar{t}^{(i)}), -\mathbf{d}) = \frac{f_p (\mathbf{r}(\bar{t}^{(i)}), -\mathbf{d}, \bar{\mathbf{d}}^{(i)})}{\text{pdf} (\bar{\mathbf{d}}^{(i)})} L_i (\mathbf{r}(\bar{t}^{(i)}, -\bar{\mathbf{d}}^{(i)}) \nonumber \\
  & \sigma_t(\mathbf{r}(t)) = \sigma_a(\mathbf{r}(t)) + \sigma_s(\mathbf{r}(t)) \nonumber
\end{align}

\section{BRDF Definition}
\label{appx:brdf_definition}
As mentioned in the main paper, we use a simplified version of Disney
BRDF~\cite{Burley2012SIGGRAPH} to model the combined effect of the volumetric
albedo $\frac{\sigma_s
(\mathbf{r}(\bar{t}^{(i)}))}{\sigma_t(\mathbf{r}(\bar{t}^{(i)}))}$ and the phase
function $f_p$. It takes predicted albedo $\alpha$, roughness $r$ and metallic
$m$ as inputs:
\begin{align}
\label{eqn:disney_brdf}
    \frac{\sigma_s}{\sigma_t} f_p (\mathbf{\omega}_o, \mathbf{\omega}_i) &= \text{BRDF} (\mathbf{\omega}_o, \mathbf{\omega}_i, \alpha, r, m, \mathbf{n}) \max\left( \mathbf{n} \cdot \mathbf{\omega}_i, 0 \right)
\end{align}
where $\mathbf{\omega}_o$ and $\mathbf{\omega}_i$ are the outgoing and incoming
directions (\ie\ surface to camera direction and surface to light direction,
respectively).  $\mathbf{\omega}_h$ is the half vector between
$\mathbf{\omega}_o$ and $\mathbf{\omega}_i$, \ie\ $\mathbf{\omega}_h =
\frac{\mathbf{\omega}_o + \mathbf{\omega}_i}{\|\mathbf{\omega}_o +
\mathbf{\omega}_i\|_2}$. $\mathbf{n}$ is the surface normal.  The BRDF is defined
as follows:
\begin{equation}
    \text{BRDF} (\mathbf{\omega}_o, \mathbf{\omega}_i, \alpha, r, m, \mathbf{n}) = (1 - m) \frac{\alpha}{\pi} + \frac{F(\mathbf{\omega}_o, \alpha, m) D(\mathbf{\omega}_h, \mathbf{n}, r) G(\mathbf{\omega}_o, \mathbf{\omega}_i, \mathbf{n})}{4 (\mathbf{n} \cdot \mathbf{\omega}_o) (\mathbf{n} \cdot \mathbf{\omega}_i)} 
\end{equation}
in which the term $(1 - m) \frac{\alpha}{\pi}$ is the diffuse component while
the remaining are specular components. For the specular component, $F$ is the
Fresnel-Schlick approximation to the exact Fresnel term, $D$ is the isotropic
GGX microfacet distribution~\cite{Heitz2018JCGT} and $G$ is Smith's shadowing
term.  They are defined as follows: 
\begin{align}
    F(\mathbf{\omega}_o, \alpha, m) &= F_0 (\alpha, m) + (1 - F_0 (\alpha, m)) 2^{ (-5.55473 \mathbf{\omega}_o \cdot \mathbf{\omega}_h - 6.98316) \mathbf{\omega}_o \cdot \mathbf{\omega}_h } \\
    D(\mathbf{\omega}_h, \mathbf{n}, r) &= \frac{r^2}{\pi ((\mathbf{n} \cdot \mathbf{\omega}_h)^2 (r^2 - 1) + 1)^2} \\
    G(\mathbf{\omega}_o, \mathbf{\omega}_i, \mathbf{n}) &= G_1 (\mathbf{\omega}_o, \mathbf{n}) G_1 (\mathbf{\omega}_i, \mathbf{n}) \\
    \text{s.t.} \quad & F_0(\alpha, m) = 0.04 (1 - m) + \alpha m \nonumber \\
    & G_1 (\mathbf{\omega}, \mathbf{n}) = \frac{(\mathbf{n} \cdot \mathbf{\omega})}{(\mathbf{n} \cdot \mathbf{\omega}) (1 - k) + k} \quad \text{and} \quad k = \frac{(r + 1)^2}{8} \nonumber
\end{align}
note that for interpolating $F$, instead of using the typical Schlick
approximation, we use the spherical Gaussian
approximation~\cite{Sebastien2012,Karis2013SIGGRAPH} which is slightly more
efficient.  $G_1$ is the \textit{Schlick-GGX} approximation to the exact Smith's
shadowing term.

\section{Implementation Details}
\label{appx:implementation_details}
In this section, we provide more details about the implementation of our
method.

\subsection{Loss Function}
\label{appx:loss_function}
In this subsection, we define $I^{(p)} \in [0, 1]^3$ as the $p$-th pixel's color
of the input image, $C_{rf}(\mathbf{r}^{(p)}) \in [0, 1]^3$ as the predicted
pixel color of the radiance field, $C_{pbr}(\mathbf{r}^{(p)}) \in [0, 1]^3$ as
the predicted pixel color of the physically based rendering, $\mathbf{M}^{(p)}
\in \{0, 1\}$ as the $p$-th pixel's ground truth binary mask value, $O^{(p)} \in
[0, 1]$ as the predicted ray opacity of the $p$-th pixel from the SDF-density
field.  Further, let $P$ denote the set of all pixels in a training batch. We
define the following loss functions:

\boldparagraph{Radiance Field (RF) Loss} We use L1 loss to measure the difference
between the predicted pixel color from the radiance field and the input image:
\begin{equation}
    \mathcal{L}_{\text{RF}} = \frac{1}{|P|} \sum_{p \in P} \left| C_{rf}(\mathbf{r}^{(p)}) - I^{(p)} \right|
\end{equation}
\boldparagraph{Physically Based Rendering (PBR) Loss} We use L1 loss to measure
the difference between the predicted pixel color from the physically based
rendering and the input image:
\begin{equation}
    \mathcal{L}_{\text{PBR}} = \frac{1}{|P|} \sum_{p \in P} \left| C_{pbr}(\mathbf{r}^{(p)}) - I^{(p)} \right|
\end{equation}
\boldparagraph{Mask Loss} We use binary cross entropy loss to measure the
difference between the predicted ray opacity and the ground truth binary mask
$\mathbf{M}$:
\begin{equation}
    \mathcal{L}_{\text{mask}} = \frac{1}{|P|} \sum_{p \in P} \left[ \mathbf{M}^{(p)} \log O^{(p)} + (1 - \mathbf{M}^{(p)}) \log (1 - O^{(p)}) \right]
\end{equation}

\boldparagraph{Eikonal Loss} We also apply Eikonel regularization to analytical
gradient of the predicted SDF value at the canonical locations $\{
    \mathbf{x}_c^{(s)} = \text{LBS}^{-1}(\mathbf{x}_o^{(s)}) \}_{s \in
\mathcal{S}}$, where $\mathbf{x}_o^{(s)}$ is a sampled point on a ray in the
observation space.  $\mathcal{S}$ is the set of all sampled points in a training
batch during ray marching of the radiance field, excluding those removed by
occupancy grids.
\begin{equation}
    \mathcal{L}_{\text{eikonal}} = \frac{1}{|\mathcal{S}|} \sum_{s \in \mathcal{S}} \left( \left\| \nabla \text{SDF} (\mathbf{x}_c^{(s)}) \right\|_2 - 1 \right)^2
\end{equation}

\boldparagraph{Curvature Loss} We apply curvature
regularization~\cite{Alexandru2023CVPR} to the same set of points on which we
compute the Eikonal loss. The curvature loss is defined as follows:
\begin{equation}
    \mathcal{L}_{\text{curv}} = \frac{1}{|\mathcal{S}|} \sum_{s \in \mathcal{S}} \left( \mathbf{n}^{(s)} \cdot \mathbf{n}^{(s)}_{\epsilon} - 1 \right)^2
\end{equation}
where $\mathbf{n}^{(s)}$ is the analytical normal at the canonical location
$\mathbf{x}_c^{(s)}$, \ie\ \textit{normalized} analytical gradient of the SDF
$\mathbf{n}^{(s)} = \frac{\nabla \text{SDF} (\mathbf{x}_c^{(s)})}{\|\nabla
\text{SDF} (\mathbf{x}_c^{(s)})\|_2}$. $\mathbf{n}^{(s)}_{\epsilon}$ is the
analytical normal of a nearby point $\mathbf{x}_c^{(s)} + \epsilon
\mathbf{t}^{(s)}$, here $\epsilon = 0.0001$ and $\mathbf{t}^{(s)}$ is a random
direction that is tangential to normal direction $\mathbf{n}^{(s)}$.

\boldparagraph{Local Smoothness Loss} We apply local smoothness regularization
on predicted albedo $\alpha$, roughness $r$ and metallic $m$ values, in a
similar way to~\cite{Zhang2021SIGGRAPHASIA,Jin2023CVPR}:
\begin{align}
    \mathcal{L}_{\text{smoothness}} &= \frac{1}{|\mathcal{P}|} \sum_{p \in \mathcal{P}} \left[ \sum_{i=1}^{N^{(p)}} w^{(p,i)} \Delta \alpha^{(p,i)} + \sum_{i=1}^{N^{(p)}} w^{(p,i)} \Delta r^{(p,i)} + \sum_{i=1}^{N^{(p)}} w^{(p,i)} \Delta m^{(p,i)} \right] \\
    \text{s.t.} \quad & \Delta \alpha^{(p,i)} = \frac{\alpha^{(p,i)} - \alpha^{(p,i)}_{\epsilon}}{\max( \max(\alpha^{(p,i)}, \alpha^{(p,i)}_{\epsilon}), 1e-6 )} \nonumber \\ 
    & \Delta r^{(p,i)} = \frac{r^{(p,i)} - r^{(p,i)}_{\epsilon}}{\max( \max(r^{(p,i)}, r^{(p,i)}_{\epsilon}), 1e-6 )} \nonumber \\
    & \Delta m^{(p,i)} = \frac{m^{(p,i)} - m^{(p,i)}_{\epsilon}}{\max( \max(m^{(p,i)}, m^{(p,i)}_{\epsilon}), 1e-6 )} \nonumber
\end{align}
where $N^{(p)}$ is the number of samples on ray $p$. $w^{(p,i)}$ is the
quadrature weight of the $i$-th sample on ray $p$.  $\alpha^{(p,i)}_{\epsilon}$,
$r^{(p,i)}_{\epsilon}$, $m^{(p,i)}_{\epsilon}$ are albedo, roughness and
metallic queried at a perturbated location near the $i$-th sample of ray $p$.

\boldparagraph{Lipschitz Bound Loss}
Lastly, we apply the Lipschitz bound loss~\cite{Liu2022SIGGRAPH} to enforce
Lipschitz smoothness of the material MLP.  ~\cite{Alexandru2023CVPR} uses the
same technique to regularize the radiance MLP. Formally, given an MLP's $i$-th
layer $y = \text{actv}(W_i x + b_i)$ along with a trainable Lipschitz bound
$k_i$, the layer is reparameterized as
\begin{equation}
    y = \text{actv} \left( \widehat{W}_i x + b_i \right), \quad \widehat{W}_i = \text{norm} (W_i, \text{softplus}(k_i))
\end{equation}
where $\text{norm}(\cdot, \cdot)$ normalizes the weight matrix $W_i$ by rescaling
each row such that the row sum's absolute value is less than or equal to the
$\text{softplus}(k_i)$. The Lipschitz bound loss is defined as follows:
\begin{equation}
    \mathcal{L}_{\text{Lip}} = \prod_{i=1}^L \text{softplus}(k_i)
\end{equation}
where $L$ is the number of layers in the MLP.

\boldparagraph{Combined Loss} The final loss function is defined as follows:
\begin{equation}
    \mathcal{L} = \mathcal{L}_{\text{RF}} + \lambda_{\text{PBR}} \mathcal{L}_{\text{PBR}} + \lambda_{\text{mask}} \mathcal{L}_{\text{mask}} + \lambda_{\text{eikonal}} \mathcal{L}_{\text{eikonal}} + \lambda_{\text{curv}} \mathcal{L}_{\text{curv}} + \lambda_{\text{smoothness}} \mathcal{L}_{\text{smoothness}} + \lambda_{\text{Lip}} \mathcal{L}_{\text{Lip}}
\end{equation}
where we set $\lambda_{\text{PBR}} = 0.2$, $\lambda_{\text{mask}} = 0.1$, $\lambda_{\text{eikonal}} = 0.1$, $\lambda_{\text{smoothness}} = 0.01$. We set $\lambda_{\text{curv}} = 1.5$ for the first $12.5$k iterations and $0$ after that. We set $\lambda_{\text{Lip}} = 1e-5$ after $12.5$k iterations and $0$ before that. 

\subsection{Albedo Evaluation}
\label{appx:albedo_evaluation}
For evaluating the predicted albedo image, we first align the predicted albedo
with the ground truth albedo.  Given $N$ samples on a ray, the predicted albedo
of a ray $\mathbf{r}$ is defined as follows:
\begin{equation}
    \hat{A}(\mathbf{r}) = \sum_{i=1}^{N} w^{(i)} \alpha^{(i)}
\end{equation}
we compute per-channel scaling factors $\mathbf{s} = (s_r, s_g, s_b)$ to align
the predicted albedo with the ground truth albedo. Given $A_r^{(p)}$ as the
ground truth red albedo of the $p$-th pixel, the following equation is computed for
$s_r$:
\begin{equation}
    \label{eqn:albedo_scaling}
    s_r = \frac{\sum_{p \in P} \hat{A}_r (\mathbf{r}^{(p)}) \cdot A_r^{(p)}}{\sum_{p \in P} \hat{A}_r (\mathbf{r}^{(p)}) \cdot \hat{A}_r (\mathbf{r}^{(p)})}
\end{equation}
while $s_g$ and $s_b$ are computed similarly.  We evaluate image quality metrics
(\ie\ PSNR, SSIM, LPIPS) on the aligned predicted albedo.  We visualize the
aligned predicted albedo on synthetic data and the non-aligned predicted albedo
on real data.

\subsection{Additional Implementation Details}
We use a mixture of 64 spherical Gaussians to represent environment lighting
during training.  During relighting, we do not use indirect illumination as the
learned radiance field on training data is not applicable to the new lighting
condition. We clip the pixel prediction from both the radiance field and the PBR
to $[0, 1]$ and apply standard gamma correction to obtain the final image in
sRGB space. For a fair comparison, we also integrate light importance sampling
into R4D for relighting, which directly samples 1024 directions on the
high-resolution environment map.

We also implement the pose optimization module following~\cite{Wang2022ECCV}.
This module is enabled for the SyntheticHuman-Relit dataset as the motion is
more complex compared to other datasets, while the original ground-truth SMPL
estimations from~\cite{Peng2024PAMI} are also slightly misaligned with the
image.

To stay consistent with R4D and~\cite{Zhang2021SIGGRAPHASIA,Ward1998Book}, we
also calibrate our albedo prediction to the range $[0.03, 0.8]$. We note this
technique is especially useful when the subject wears near-black clothes (\ie\
albedo $< 0.1$ for all channels).


\boldparagraph{Temporal Occupancy Grids} A common technique to reduce computation
is to maintain an occupancy grid to mark occupied voxels during training and
skip unoccupied voxels during ray
marching/tracing~\cite{Mueller2022SIGGRAPH,AlexYuandSaraFridovich-Keil2022CVPR,Chen2022ECCV,Li2023ICCV}
.  This also applies to temporal reconstruction as one can define the occupancy
grid as the union of all shapes from different time steps~\cite{Jiang2023CVPR}.
To further reduce the computational cost, we employ a 4D occupancy grid
structure~\cite{Peng2023CVPR} in which we maintain a $64 ^ 3$ occupancy grid for
each training frame.  At the beginning of training, we first use a single
occupancy grid for all frames.  After we have attained a reasonable SDF we
re-initialize the occupancy grid for each frame using the learned canonical SDF.

\section{SyntheticHuman-Relit Dataset} 
\label{appx:dataset}
To properly compare with R4D on relighting of training poses, we created a new
dataset, SyntheticHuman-Relit, which is a subset of the SyntheticHuman dataset
\cite{Peng2024PAMI} relit using new material and lighting conditions.  The
dataset consists of 2 synthetic humans (\textit{Jody} and \textit{Leonard}),
each rotating in front of a fixed camera.

\begin{figure*}[h]
    \centering
    \begin{subfigure}[b]{0.24\linewidth}
        \centering
        \includegraphics[width=\linewidth, trim=0cm 0cm 0cm 0cm, clip]{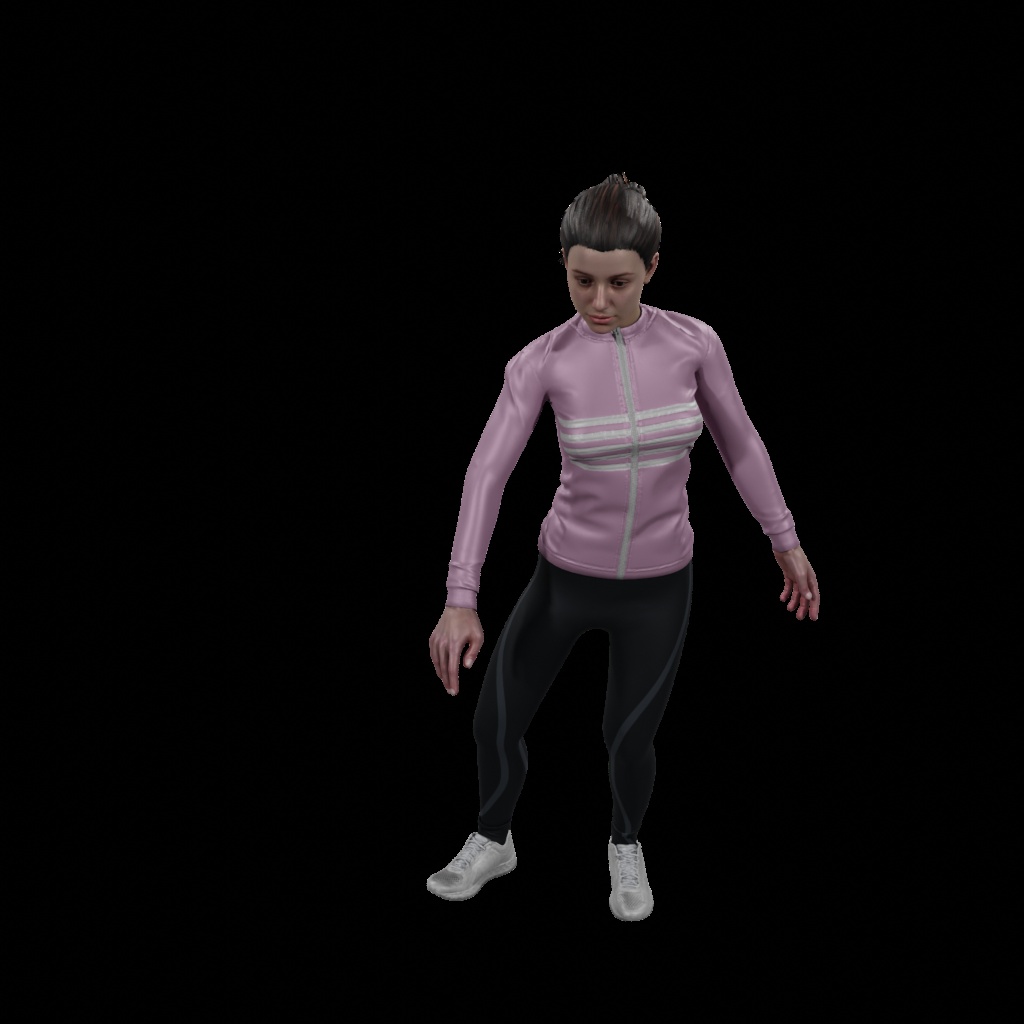}
        \caption{SyntheticHuman - Jody}
    \end{subfigure}
    \begin{subfigure}[b]{0.24\linewidth}
        \centering
        \includegraphics[width=\linewidth, trim=0cm 0cm 0cm 0cm, clip]{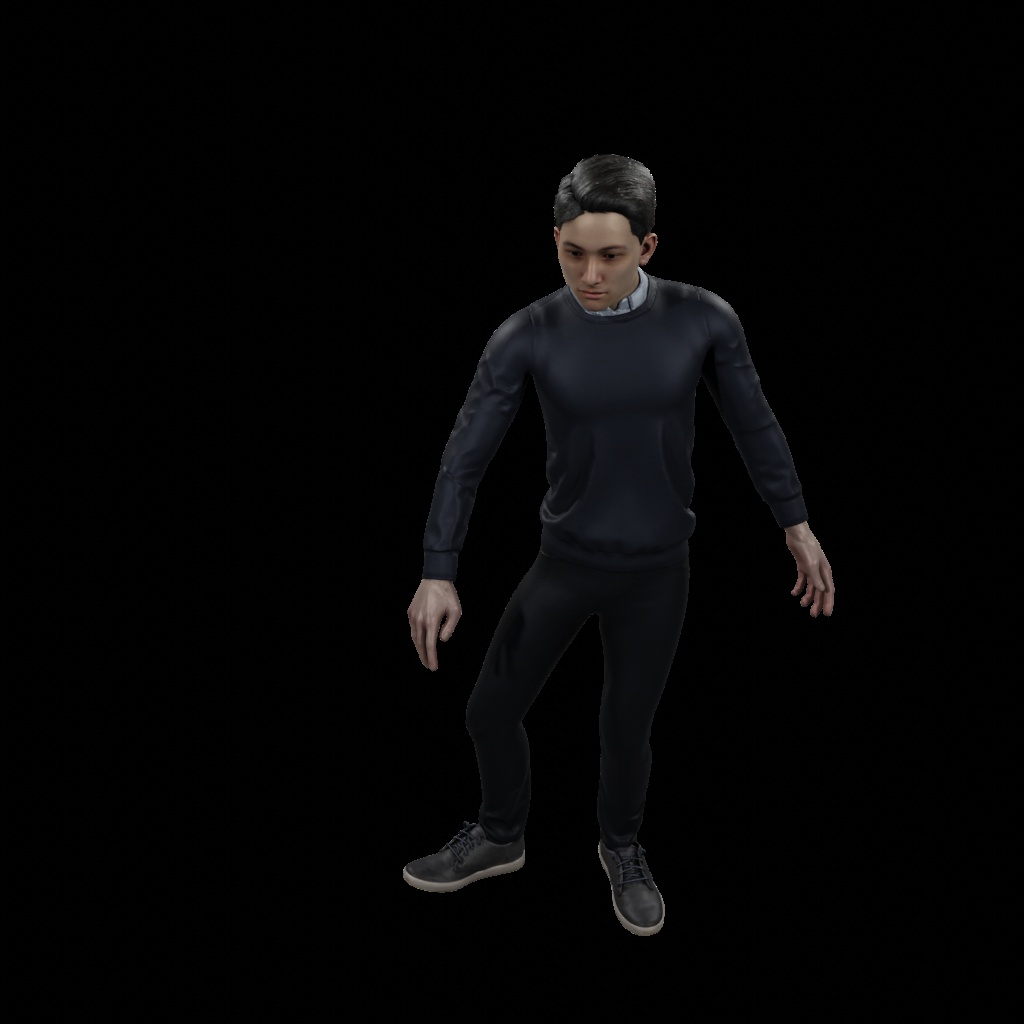}
        \caption{SyntheticHuman - Leonard}
    \end{subfigure}
    \begin{subfigure}[b]{0.24\linewidth}
        \centering
        \includegraphics[width=\linewidth, trim=0cm 0cm 0cm 0cm, clip]{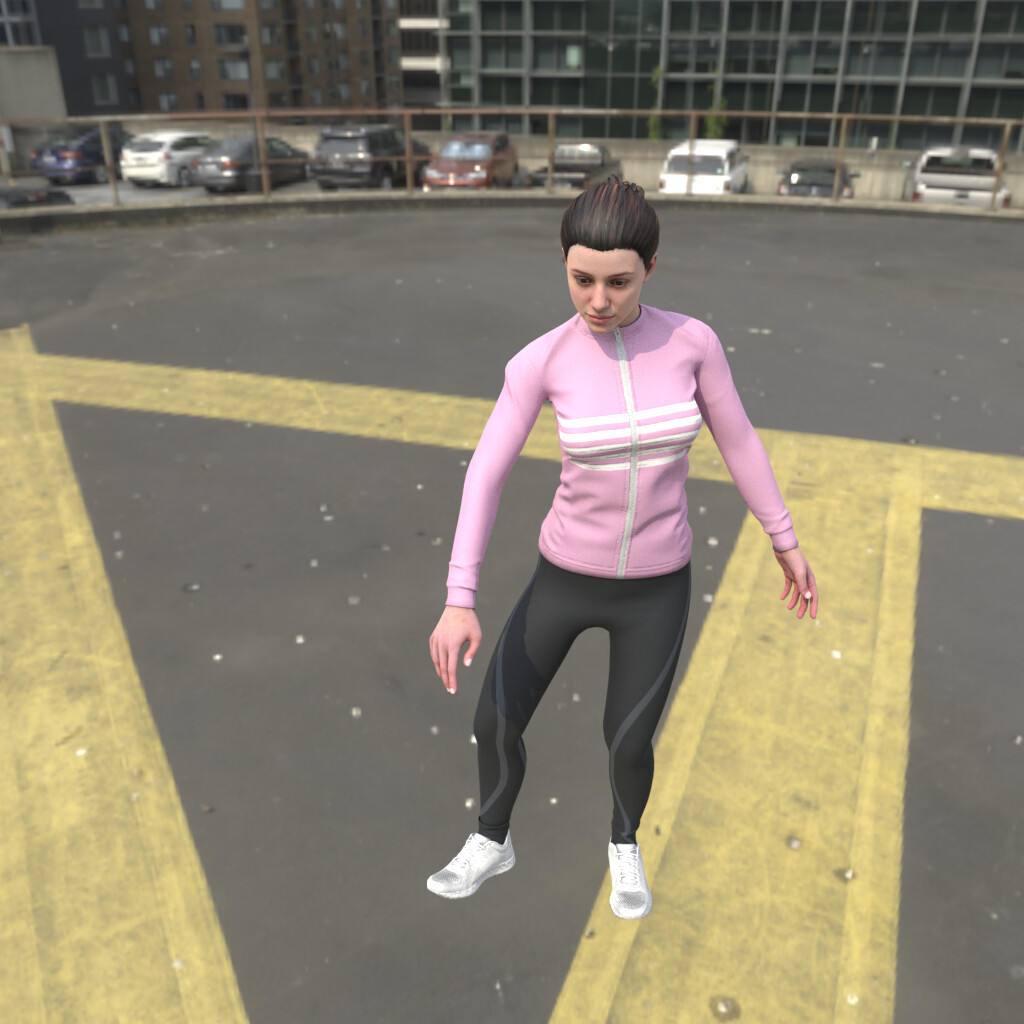}
        \caption{SyntheticHuman-Relit - Jody}
    \end{subfigure}
    \begin{subfigure}[b]{0.24\linewidth}
        \centering
        \includegraphics[width=\linewidth, trim=0cm 0cm 0cm 0cm, clip]{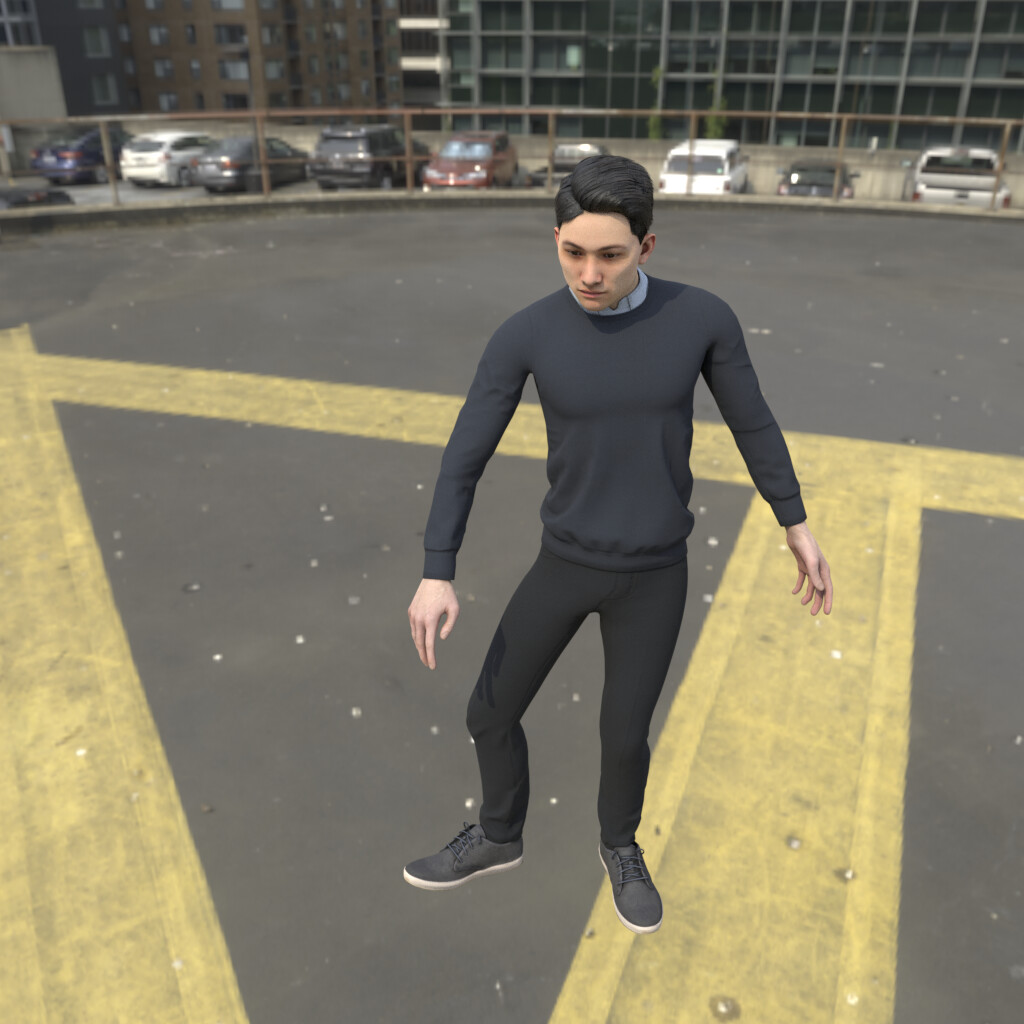}
        \caption{SyntheticHuman-Relit - Leonard}
    \end{subfigure}
    \caption{\textbf{Comparison between the SyntheticHuman dataset and the SyntheticHuman-Relit dataset.} Note that the SyntheticHuman dataset is overly specular compared to real humans, while the light sources are also studio-like. In contrast, SyntheticHuman-relit adopts a more diffuse appearance which is closer to real humans, while the subjects are lit under natural, outdoor illumination.}
    \label{app:fig:synhuman-relit_dataset}
\end{figure*}

We note that the original SyntheticHuman dataset was rendered under studio
lighting and the materials were overly specular compared to real humans.  We
thus adjusted the materials and re-rendered the dataset under more natural
lighting conditions.  See a comparison of the original SyntheticHuman dataset and
the new SyntheticHuman-Relit dataset in Fig.~\ref{app:fig:synhuman-relit_dataset}.

To test relighting on training poses, we further re-rendered each training pose
of the SyntheticHuman-Relit dataset under a random environment map that was
not used in the training set.

\captionsetup[subfigure]{labelformat=empty}

\section{Additional Quantitative Results}
\label{appx:additional_quantitative_results}
The per-subject and average metrics of R4D, R4D*, and Ours are reported in
Tab.~\ref{tab:metric_all_rana}. We also tested a variant of our approach that
does not calibrate the albedo into the range $[0.03, 0.8]$, denoted as
Ours$^\dagger$.  Since R4D* and Ours achieve overall better performance than
their variants (R4D and Ours$^\dagger$) on the RANA dataset, we only report R4D*
and Ours on the SyntheticHuman-Relit dataset in
Tab.~\ref{tab:metric_all_synhuman}.  We also additionally report
ARAH~\cite{Wang2022ECCV}'s results on geometry reconstruction, evaluated by the
normal error metric. Albedo estimation and relighting are not evaluated as ARAH
does not predict the intrinsic properties of avatars.

\begin{table*}[t]
    \centering
    \begin{tabular}{l|c|c|c|c|c|c|c|c|}
        \toprule
        \multirow{2}{*}{Subject} & \multirow{2}{*}{Method} & \multicolumn{3}{c|}{Albedo} & \multirow{1}{*}{Normal} & \multicolumn{3}{c}{Relighting (Novel Pose)} \\
        \cmidrule{3-5} \cmidrule{7-9}
        & & PSNR $\uparrow$ & SSIM $\uparrow$ & LPIPS $\downarrow$ & \multicolumn{1}{c|}{Error $\downarrow$} & PSNR $\uparrow$ & SSIM $\uparrow$ & LPIPS $\downarrow$ \\
        \cmidrule{1-9}
        \multirow{5}{*}{Subject 01}
        & ARAH  & - & - & - & 12.89 \degree & - & - & - \\
        & R4D  & 20.64 & 0.7673 & 0.2199 & 64.07 \degree & 11.73 & 0.7865 & 0.2028 \\
        & R4D* & 20.04 & 0.8525 & 0.2079 & 33.61 \degree & 18.22 & 0.8425 & 0.1612 \\
        & Ours$^\dagger$ & 23.69 & 0.7998 & 0.1916 & \textbf{11.35 \degree} & 18.35 & 0.8727 & \textbf{0.1200} \\
        & Ours & \textbf{24.11} & \textbf{0.8679} & \textbf{0.1827} & 12.05 \degree & \textbf{18.48} & \textbf{0.8859} & 0.1219 \\
        \cmidrule{1-9}
        \multirow{5}{*}{Subject 02}
        & ARAH  & - & - & - & 11.92 \degree & - & - & - \\
        & R4D  & 15.14 & 0.8089 & 0.2926 & 30.20 \degree & 15.08 & 0.8361 & 0.1954 \\
        & R4D* & 12.13 & 0.7690 & 0.2599 & 28.34 \degree & 14.38 & 0.8128 & 0.1787 \\
        & Ours$^\dagger$ & 20.25 & 0.8733 & 0.1898 & \textbf{9.27 \degree} & 18.86 & 0.8781 & 0.1336 \\
        & Ours & \textbf{20.94} & \textbf{0.8892} & \textbf{0.1854} & 9.29 \degree & \textbf{19.08} & \textbf{0.8812} & \textbf{0.1323} \\
        \cmidrule{1-9}
        \multirow{5}{*}{Subject 05}
        & ARAH  & - & - & - & 9.78 \degree & - & - & - \\
        & R4D  & 19.66 & 0.8223 & 0.2484 & 31.18 \degree & 16.59 & 0.8354 & 0.1916 \\
        & R4D* & 19.74 & 0.8151 & 0.2488 & 26.14 \degree & \textbf{17.72} & 0.8469 & 0.1780 \\
        & Ours$^\dagger$ & 21.06 & 0.8159 & 0.2262 & \textbf{9.51 \degree} & 17.40 & 0.8750 & 0.1466 \\
        & Ours & \textbf{22.24} & \textbf{0.8591} & \textbf{0.2071} & 9.52 \degree & 17.47 & \textbf{0.8769} & \textbf{0.1453} \\
        \cmidrule{1-9}
        \multirow{5}{*}{Subject 06}
        & ARAH  & - & - & - & 12.06 \degree & - & - & - \\
        & R4D  & 17.26 & 0.5954 & 0.3466 & 81.79 \degree & 7.31 & 0.7567 & 0.2821 \\
        & R4D* & 21.57 & 0.7992 & 0.2177 & 25.83 \degree & 17.54 & 0.8866 & 0.1636 \\
        & Ours$^\dagger$ & 21.07 & 0.7093 & 0.2241 & 8.91 \degree & 17.89 & 0.8647 & 0.1294 \\
        & Ours & \textbf{22.94} & \textbf{0.8233} & \textbf{0.1928} & \textbf{8.89} \degree & \textbf{18.14} & \textbf{0.8932} & \textbf{0.1271} \\
        \cmidrule{1-9}
        \multirow{5}{*}{Subject 33}
        & ARAH  & - & - & - & 10.28 \degree & - & - & - \\
        & R4D  & 17.95 & 0.8335 & 0.1900 & 27.53 \degree & 16.08 & 0.8202 & 0.1960 \\
        & R4D* & 18.35 & 0.8426 & 0.1887 & 25.24 \degree & 16.78 & 0.8173 & 0.1859 \\
        & Ours$^\dagger$ & \textbf{21.78} & 0.8395 & \textbf{0.1259} & \textbf{9.07 \degree} & 17.62 & 0.8352 & \textbf{0.1332} \\
        & Ours & 21.67 & \textbf{0.8703} & 0.1351 & 9.52 \degree & \textbf{18.03} & \textbf{0.8426} & 0.1366 \\
        \cmidrule{1-9}
        \multirow{5}{*}{Subject 36}
        & ARAH  & - & - & - & 11.62 \degree & - & - & - \\
        & R4D  & 20.38 & 0.9091 & 0.1844 & 43.44 \degree & 15.99 & 0.8200 & 0.1899 \\
        & R4D* & 23.80 & \textbf{0.9100} & 0.1611 & 24.76 \degree & 17.05 & 0.8574 & 0.1707 \\
        & Ours$^\dagger$ & 24.30 & 0.7946 & 0.1739 & \textbf{9.09 \degree} & 17.25 & 0.8520 & 0.1308 \\
        & Ours & \textbf{24.88} & 0.8900 & \textbf{0.1324} & 9.22 \degree & \textbf{17.46} & \textbf{0.8726} & \textbf{0.1284} \\
        \cmidrule{1-9}
        \multirow{5}{*}{Subject 46}
        & ARAH  & - & - & - & \textbf{10.38 \degree} & - & - & - \\
        & R4D  & 16.40 & 0.8381 & 0.1455 & 32.64 \degree & 16.05 & 0.8289 & 0.1720 \\
        & R4D* & 18.13 & 0.8777 & 0.1238 & 33.27 \degree & 16.30 & 0.8338 & 0.1649 \\
        & Ours$^\dagger$ & 22.17 & 0.9314 & 0.0744 & 10.41 \degree & 16.89 & 0.8377 & \textbf{0.0965} \\
        & Ours & \textbf{22.47} & \textbf{0.9391} & \textbf{0.0725} & 10.69 \degree & \textbf{17.08} & \textbf{0.8406} & 0.1000 \\
        \cmidrule{1-9}
        \multirow{5}{*}{Subject 48}
        & ARAH  & - & - & - & \textbf{10.13 \degree} & - & - & - \\
        & R4D  & 18.50 & 0.8502 & 0.3037 & 30.67 \degree & 16.10 & 0.8224 & 0.1840 \\
        & R4D* & 12.10 & 0.7370 & 0.2264 & 21.84 \degree & 14.98 & 0.7985 & 0.1776 \\
        & Ours$^\dagger$ & 23.28 & 0.9075 & \textbf{0.1838} & 10.32 \degree & 19.50 & 0.8823 & \textbf{0.1307} \\
        & Ours & \textbf{23.36} & \textbf{0.9137} & 0.1857 & 10.49 \degree & \textbf{19.70} & \textbf{0.8849} & 0.1313 \\
        \cmidrule{1-9}
        \multirow{5}{*}{Average}
        & ARAH  & - & - & - & 11.13 \degree & - & - & - \\
        & R4D  & 18.24 & 0.7780 & 0.2414 & 42.69 \degree & 14.37 & 0.8133 & 0.2017 \\
        & R4D* & 18.23 & 0.8254 & 0.2043 & 27.38 \degree & 16.62 & 0.8370 & 0.1726 \\
        & Ours$^\dagger$ & 22.20 & 0.8339 & 0.1737 & \textbf{9.74 \degree} & 17.97 & 0.8622 & \textbf{0.1276} \\
        & Ours & \textbf{22.83} & \textbf{0.8816} & \textbf{0.1617} & 9.96 \degree & \textbf{18.18} & \textbf{0.8722} & 0.1279 \\
        \bottomrule
    \end{tabular}
\caption{\textbf{Metrics on the RANA dataset.}}
    \label{tab:metric_all_rana}
\end{table*}

\begin{table*}[t]
    \centering
    \begin{tabular}{l|c|c|c|c|c|c|c|c|}
        \toprule
        \multirow{2}{*}{Subject} & \multirow{2}{*}{Method} & \multicolumn{3}{c|}{Albedo} & \multirow{1}{*}{Normal} & \multicolumn{3}{c}{Relighting (Training Pose)} \\
        \cmidrule{3-5} \cmidrule{7-9}
        & & PSNR $\uparrow$ & SSIM $\uparrow$ & LPIPS $\downarrow$ & \multicolumn{1}{c|}{Error $\downarrow$} & PSNR $\uparrow$ & SSIM $\uparrow$ & LPIPS $\downarrow$ \\
        \cmidrule{1-9}
        \multirow{3}{*}{Jody} 
        & ARAH  & - & - & - & 15.79 \degree & - & - & - \\
        & R4D* & 17.95 & 0.7275 & 0.2319 & 33.51 \degree & 21.85 & 0.9012 & 0.1277 \\
        & Ours & \textbf{23.10} & \textbf{0.8353} & \textbf{0.1584} & \textbf{13.90} \degree & \textbf{22.24} & \textbf{0.9336} & \textbf{0.1055} \\
        \cmidrule{1-9}
        \multirow{3}{*}{Leonard}
        & ARAH  & - & - & - & 15.96 \degree & - & - & - \\
        & R4D* & 25.67 & \textbf{0.8838} & 0.1841 & 25.93 \degree & 23.23 & 0.9216 & 0.1296 \\
        & Ours & \textbf{26.98} & 0.7872 & \textbf{0.1568} & \textbf{14.45} \degree & \textbf{24.23} & \textbf{0.9490} & \textbf{0.0954} \\
        \cmidrule{1-9}
        \multirow{3}{*}{Average}
        & ARAH  & - & - & - & 15.88 \degree & - & - & - \\
        & R4D* & 21.81 & 0.8057 & 0.2080 & 29.72 \degree & 22.57 & 0.9123 & 0.1283 \\
        & Ours & \textbf{25.04} & \textbf{0.8113} & \textbf{0.1567} & \textbf{14.18} \degree & \textbf{23.24} & \textbf{0.9413} & \textbf{0.1005} \\
        \bottomrule
    \end{tabular}
\caption{\textbf{Metrics on the SyntheticHuman-Relit dataset.}}
\label{tab:metric_all_synhuman}
\end{table*}

\section{Additional Qualitative Results}
\label{appx:additional_qualitative_results}
\begin{figure*}[h]
    \centering
    \textbf{Ours}
    \par\medskip
    \begin{subfigure}[b]{0.12\linewidth}
        \centering
        \includegraphics[width=\linewidth, trim=13cm 1.5cm 12.5cm 3cm, clip]{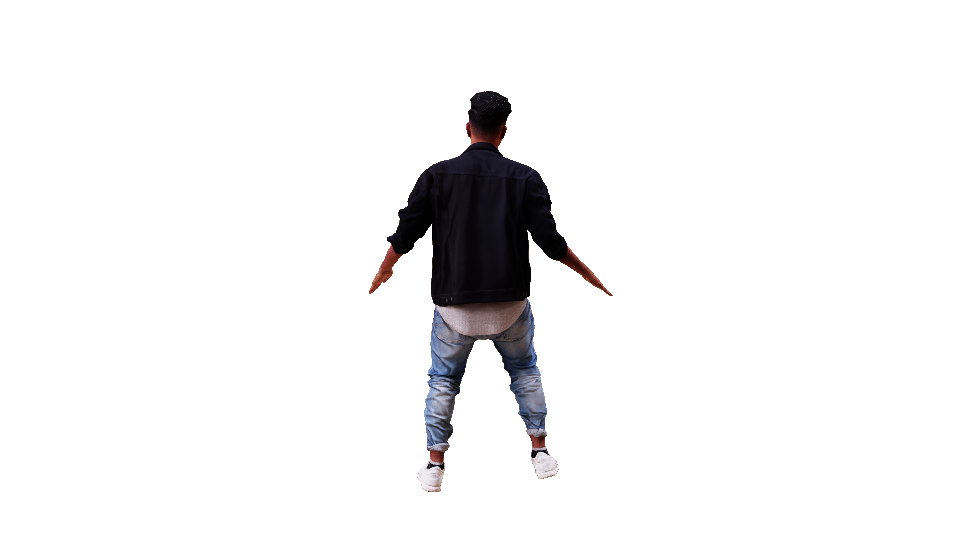}
        \caption{Input RGB}
    \end{subfigure}
    \begin{subfigure}[b]{0.12\linewidth}
        \centering
        \includegraphics[width=\linewidth, trim=13cm 1.5cm 12.5cm 3cm, clip]{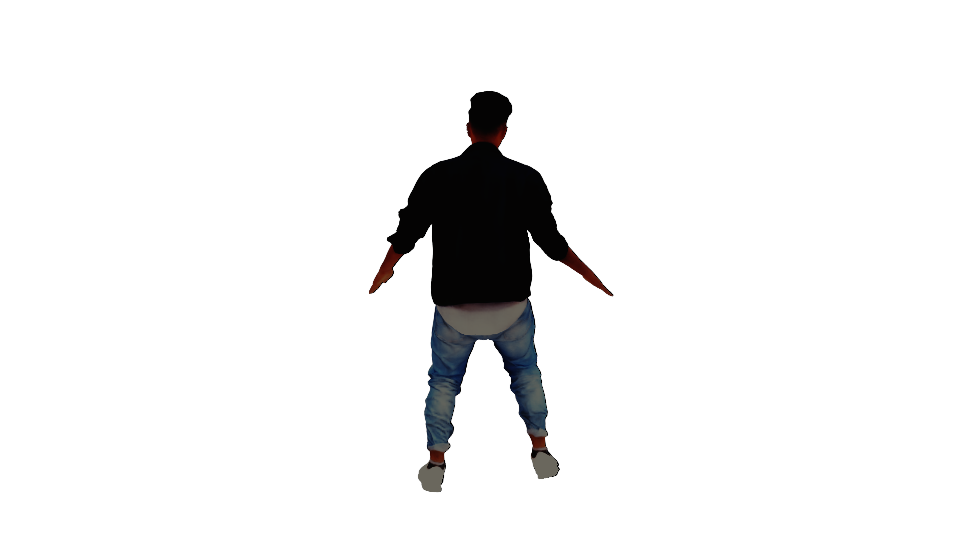}
        \caption{Albedo}
    \end{subfigure}
    \begin{subfigure}[b]{0.12\linewidth}
        \centering
        \includegraphics[width=\linewidth, trim=13cm 1.5cm 12.5cm 3cm, clip]{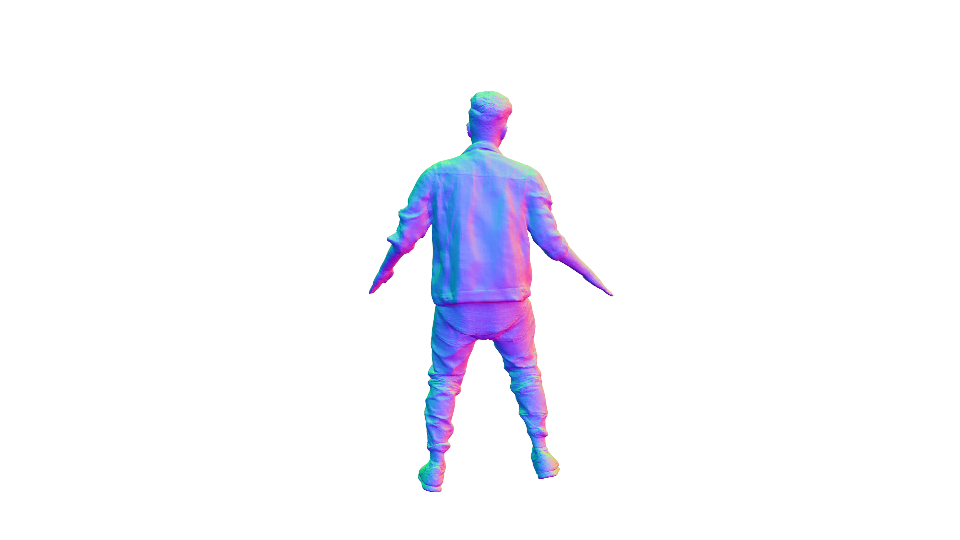}
        \caption{Geometry}
    \end{subfigure}
    \begin{subfigure}[b]{0.12\linewidth}
        \centering
        \includegraphics[width=\linewidth, trim=17cm 2cm 17cm 4cm, clip]{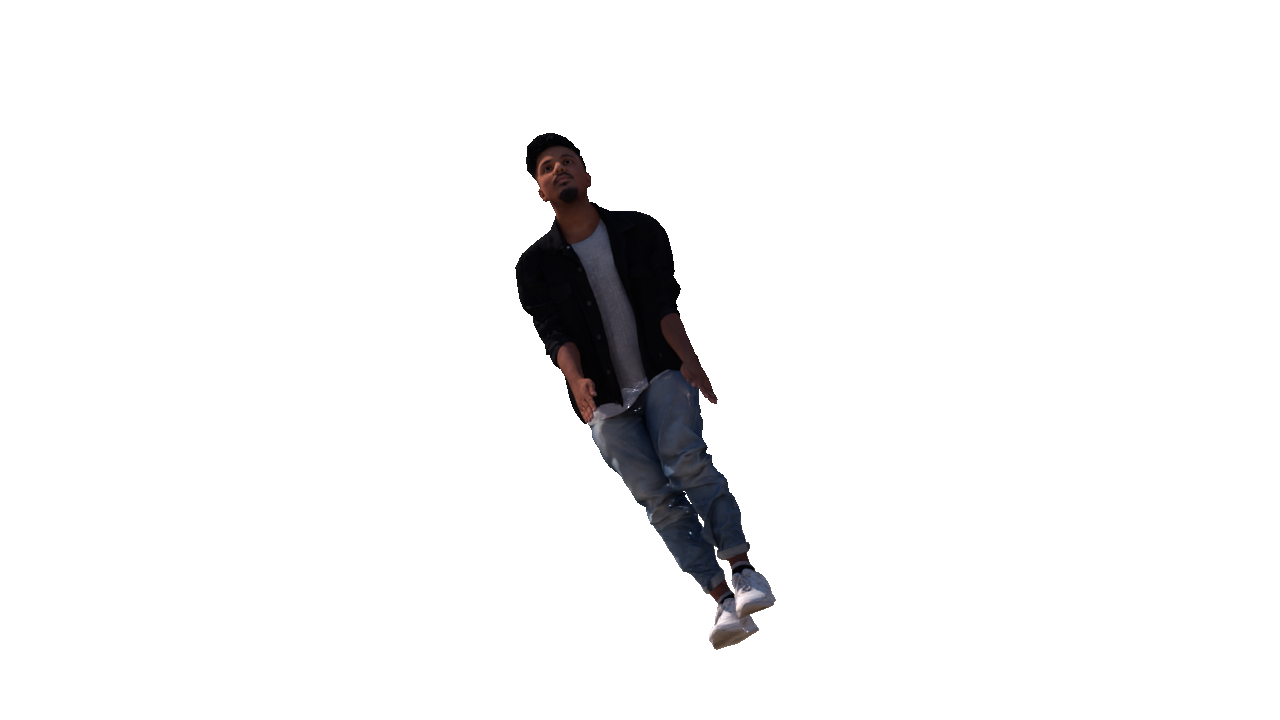}
        \caption{Repose and Relit}
    \end{subfigure}
    \begin{subfigure}[b]{0.12\linewidth}
        \centering
        \includegraphics[width=\linewidth, trim=12.5cm 1cm 12.5cm 3cm, clip]{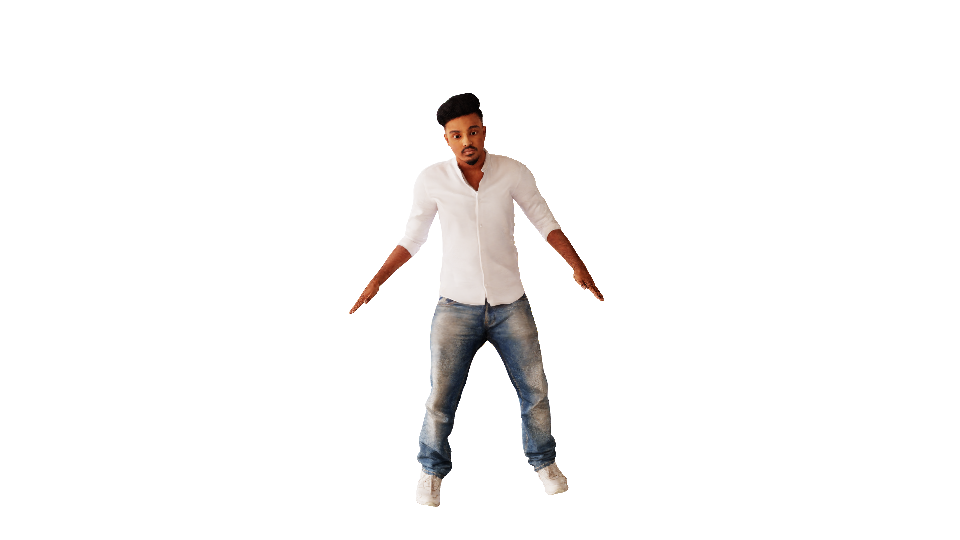}
        \caption{Input RGB}
    \end{subfigure}
    \begin{subfigure}[b]{0.12\linewidth}
        \centering
        \includegraphics[width=\linewidth, trim=12.5cm 1cm 12.5cm 3cm, clip]{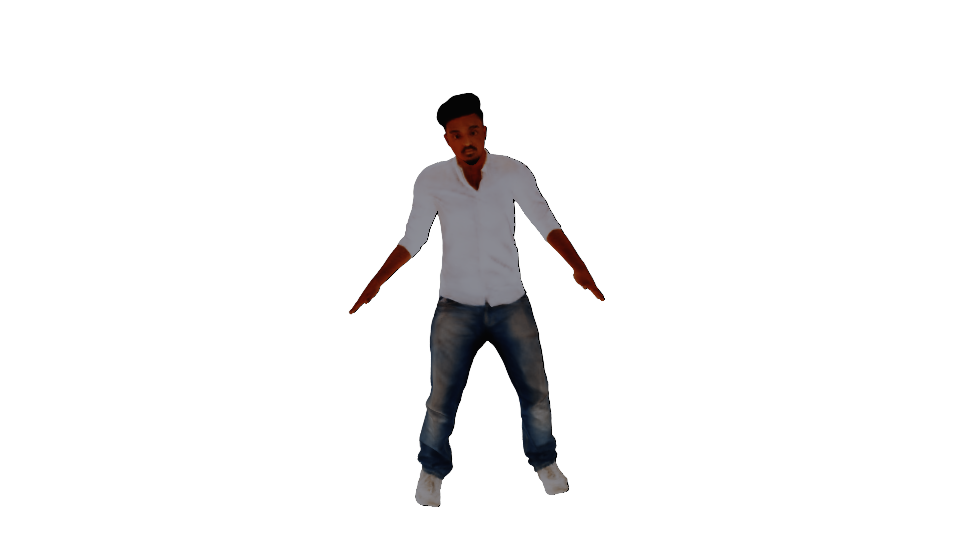}
        \caption{Albedo}
    \end{subfigure}
    \begin{subfigure}[b]{0.12\linewidth}
        \centering
        \includegraphics[width=\linewidth, trim=12.5cm 1cm 12.5cm 3cm, clip]{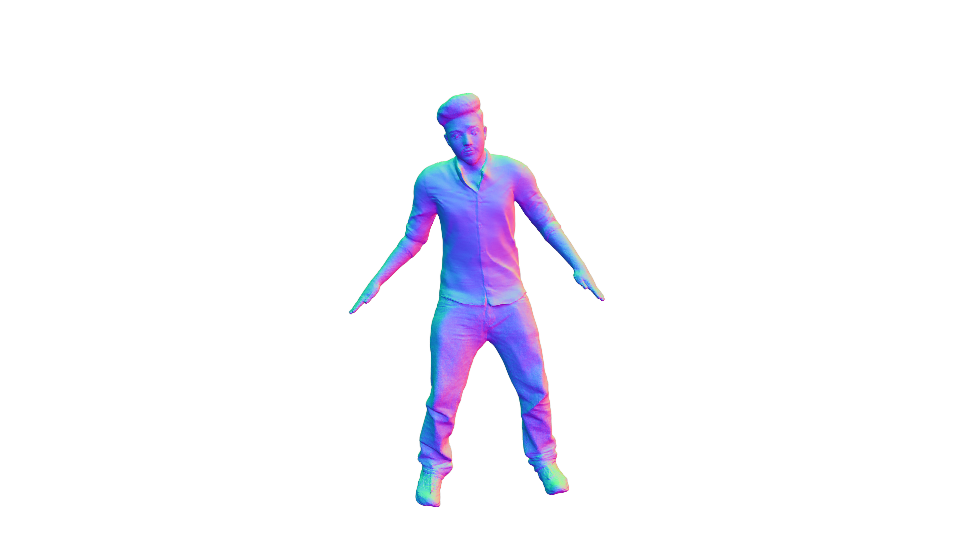}
        \caption{Geometry}
    \end{subfigure}
    \begin{subfigure}[b]{0.12\linewidth}
        \centering
        \includegraphics[width=\linewidth, trim=18.5cm 1cm 14.5cm 3.5cm, clip]{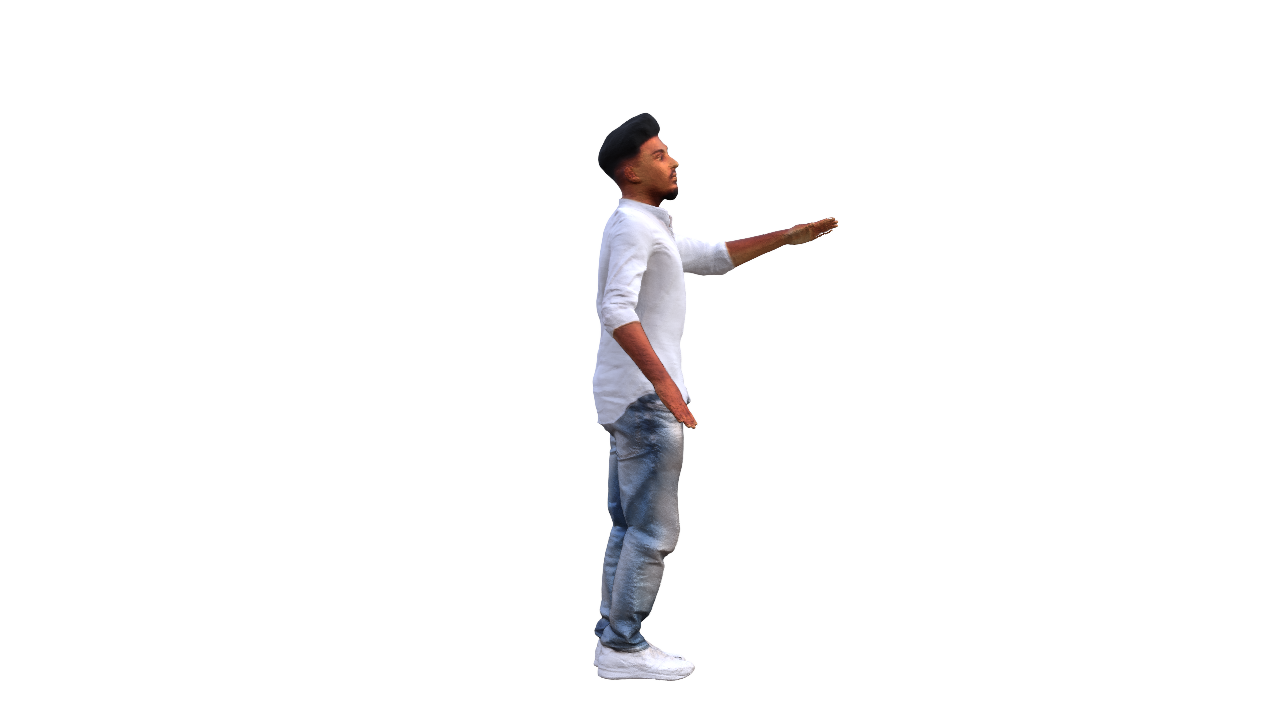}
        \caption{Repose and Relit}
    \end{subfigure}
    \vskip\baselineskip
    \textbf{R4D*}
    \par\medskip
    \begin{subfigure}[b]{0.12\linewidth}
        \centering
        \includegraphics[width=\linewidth, trim=13cm 1.5cm 12.5cm 3cm, clip]{fig/images/01/rgb_train_0049_gt.png}
        \caption{Input RGB}
    \end{subfigure}
    \begin{subfigure}[b]{0.12\linewidth}
        \centering
        \includegraphics[width=\linewidth, trim=13cm 1.5cm 12.5cm 3cm, clip]{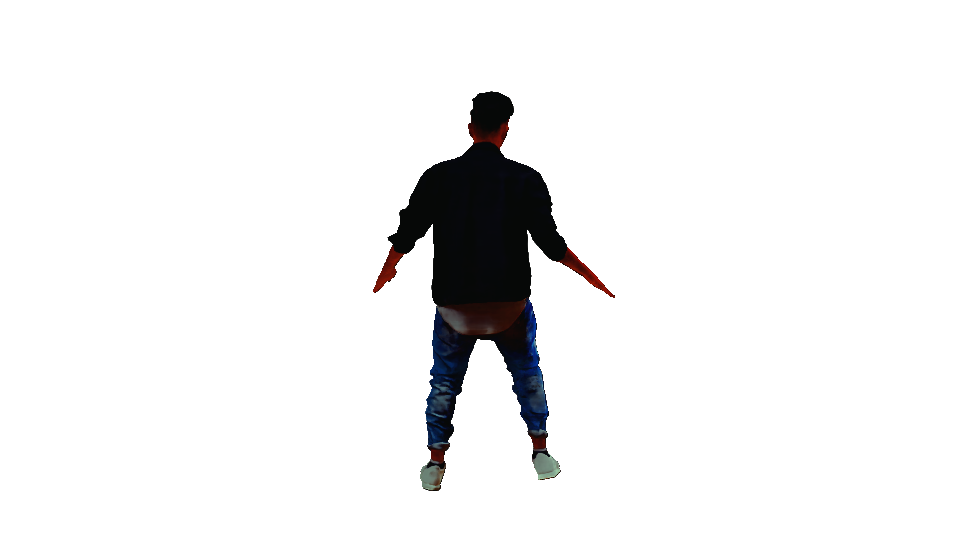}
        \caption{Albedo}
    \end{subfigure}
    \begin{subfigure}[b]{0.12\linewidth}
        \centering
        \includegraphics[width=\linewidth, trim=13cm 1.5cm 12.5cm 3cm, clip]{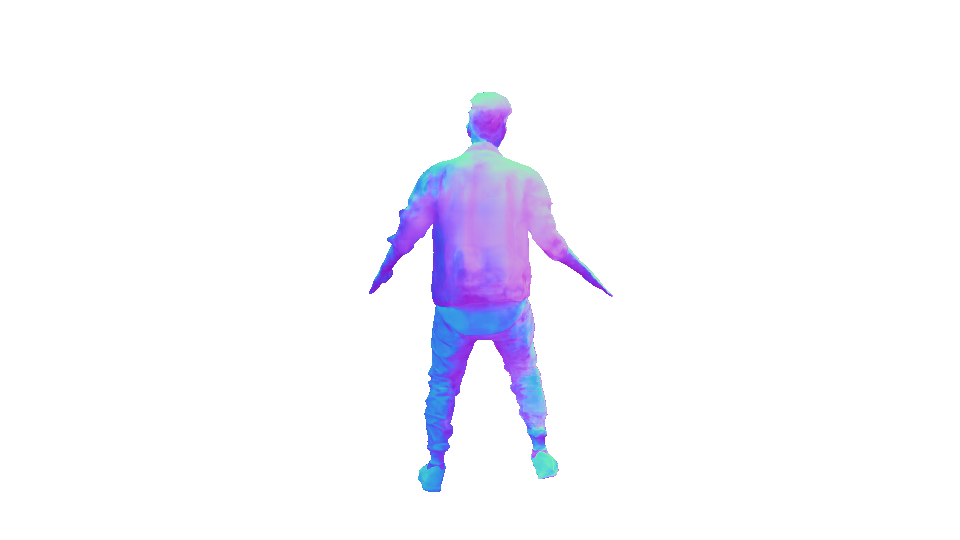}
        \caption{Geometry}
    \end{subfigure}
    \begin{subfigure}[b]{0.12\linewidth}
        \centering
        \includegraphics[width=\linewidth, trim=17cm 2cm 17cm 4cm, clip]{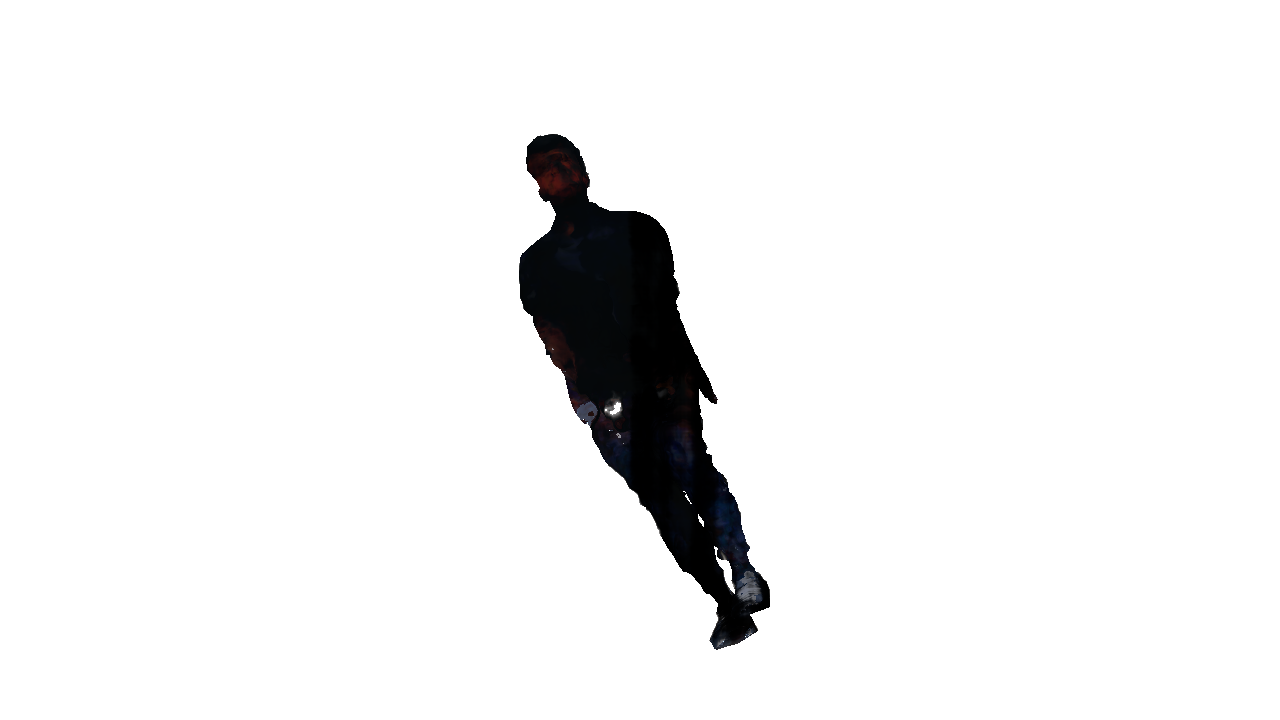}
        \caption{Repose and Relit}
    \end{subfigure}
    \begin{subfigure}[b]{0.12\linewidth}
        \centering
        \includegraphics[width=\linewidth, trim=12.5cm 1cm 12.5cm 3cm, clip]{fig/images/02/rgb_train_0000_gt.png}
        \caption{Input RGB}
    \end{subfigure}
    \begin{subfigure}[b]{0.12\linewidth}
        \centering
        \includegraphics[width=\linewidth, trim=12.5cm 1cm 12.5cm 3cm, clip]{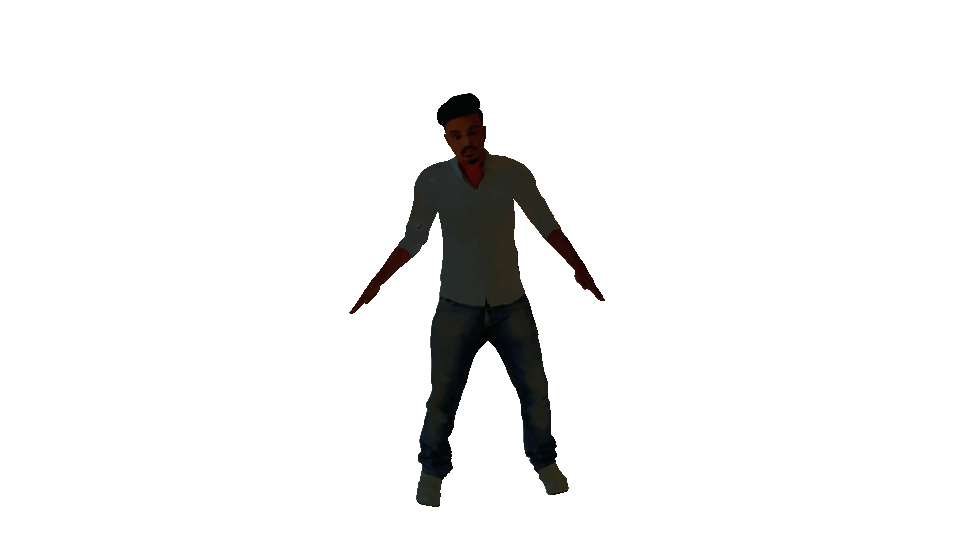}
        \caption{Albedo}
    \end{subfigure}
    \begin{subfigure}[b]{0.12\linewidth}
        \centering
        \includegraphics[width=\linewidth, trim=12.5cm 1cm 12.5cm 3cm, clip]{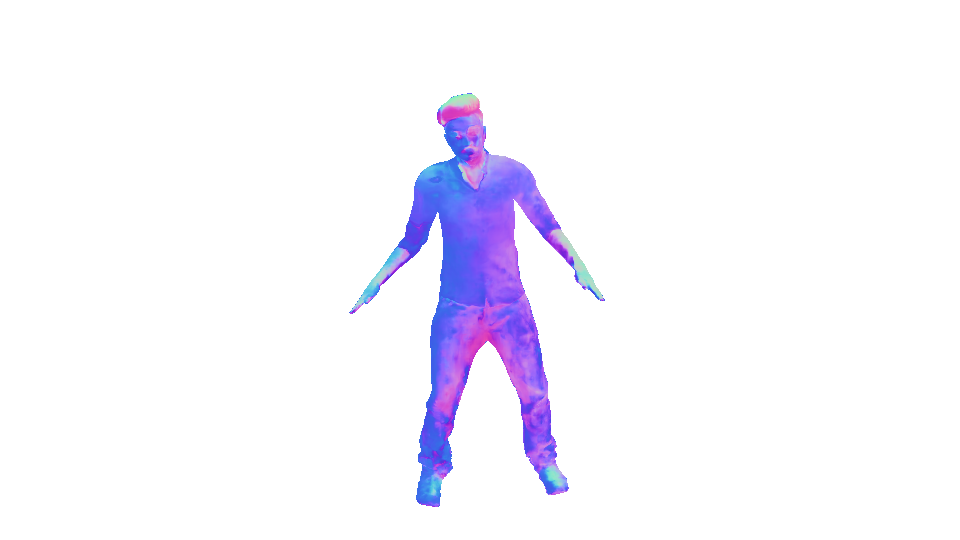}
        \caption{Geometry}
    \end{subfigure}
    \begin{subfigure}[b]{0.12\linewidth}
        \centering
        \includegraphics[width=\linewidth, trim=18.5cm 1cm 14.5cm 3.5cm, clip]{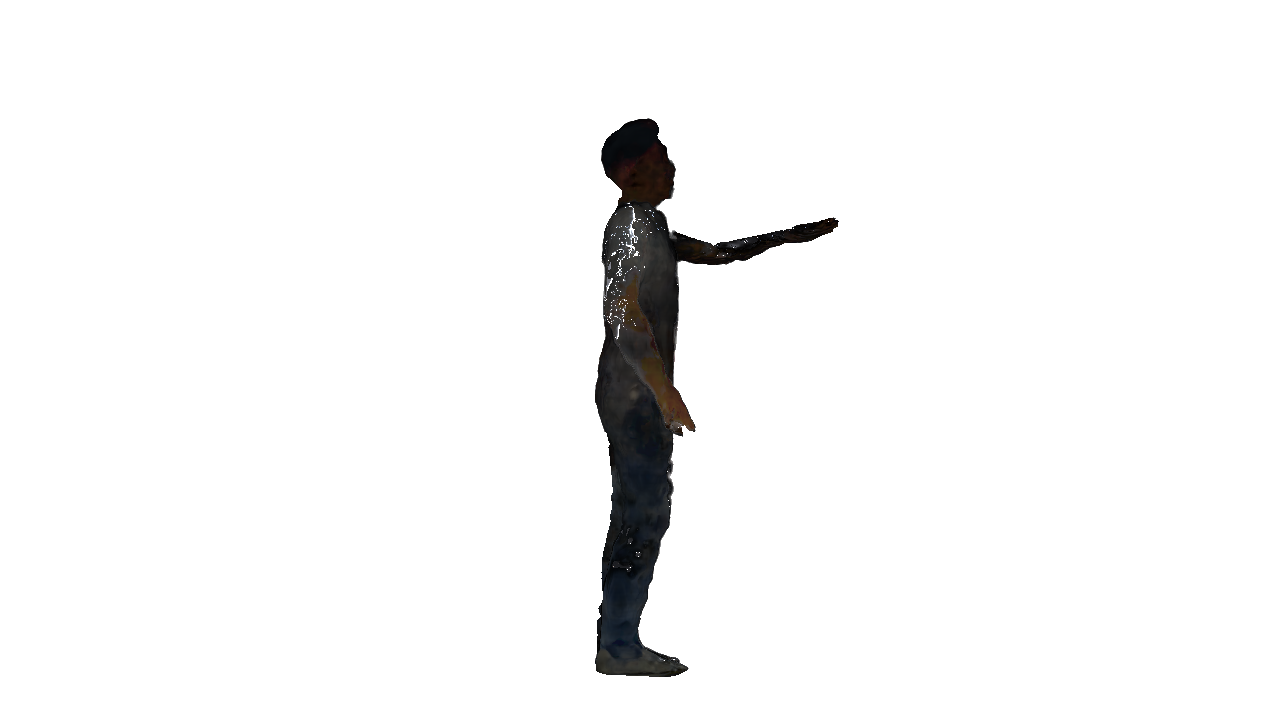}
        \caption{Repose and Relit}
    \end{subfigure}
    \vskip\baselineskip
    \textbf{Ground Truth}
    \par\medskip
    \begin{subfigure}[b]{0.12\linewidth}
        \centering
        \includegraphics[width=\linewidth, trim=13cm 1.5cm 12.5cm 3cm, clip]{fig/images/01/rgb_train_0049_gt.png}
        \caption{Input RGB}
    \end{subfigure}
    \begin{subfigure}[b]{0.12\linewidth}
        \centering
        \includegraphics[width=\linewidth, trim=13cm 1.5cm 12.5cm 3cm, clip]{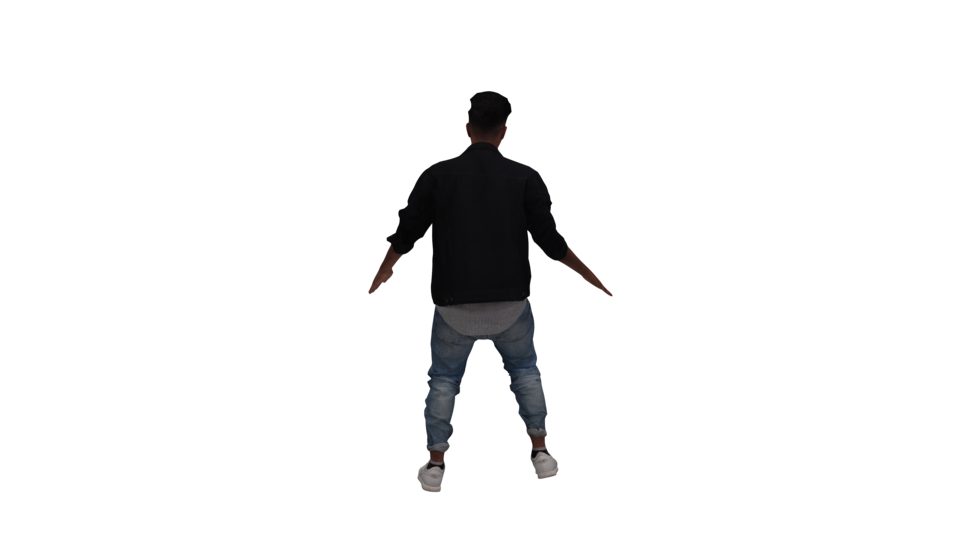}
        \caption{Albedo}
    \end{subfigure}
    \begin{subfigure}[b]{0.12\linewidth}
        \centering
        \includegraphics[width=\linewidth, trim=13cm 1.5cm 12.5cm 3cm, clip]{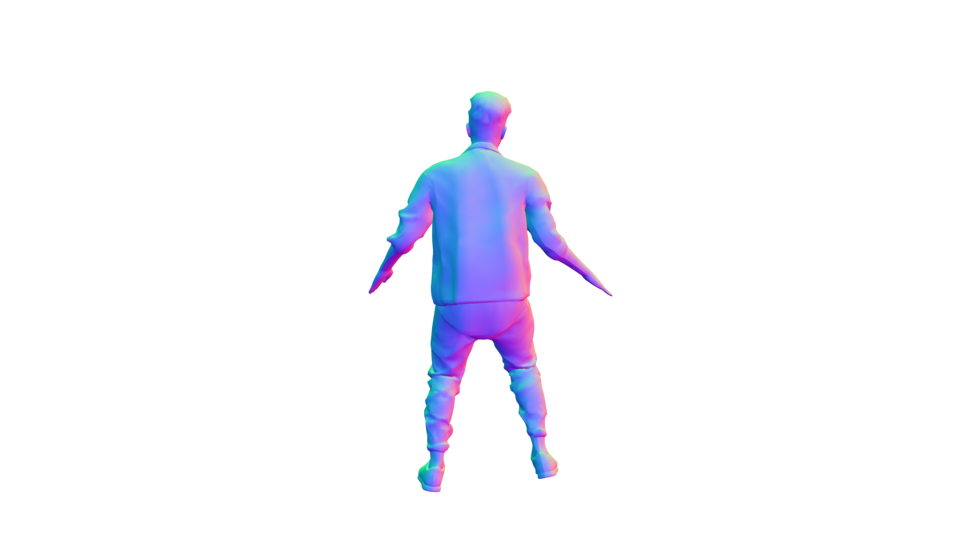}
        \caption{Geometry}
    \end{subfigure}
    \begin{subfigure}[b]{0.12\linewidth}
        \includegraphics[width=\linewidth, trim=17cm 2cm 17cm 4cm, clip]{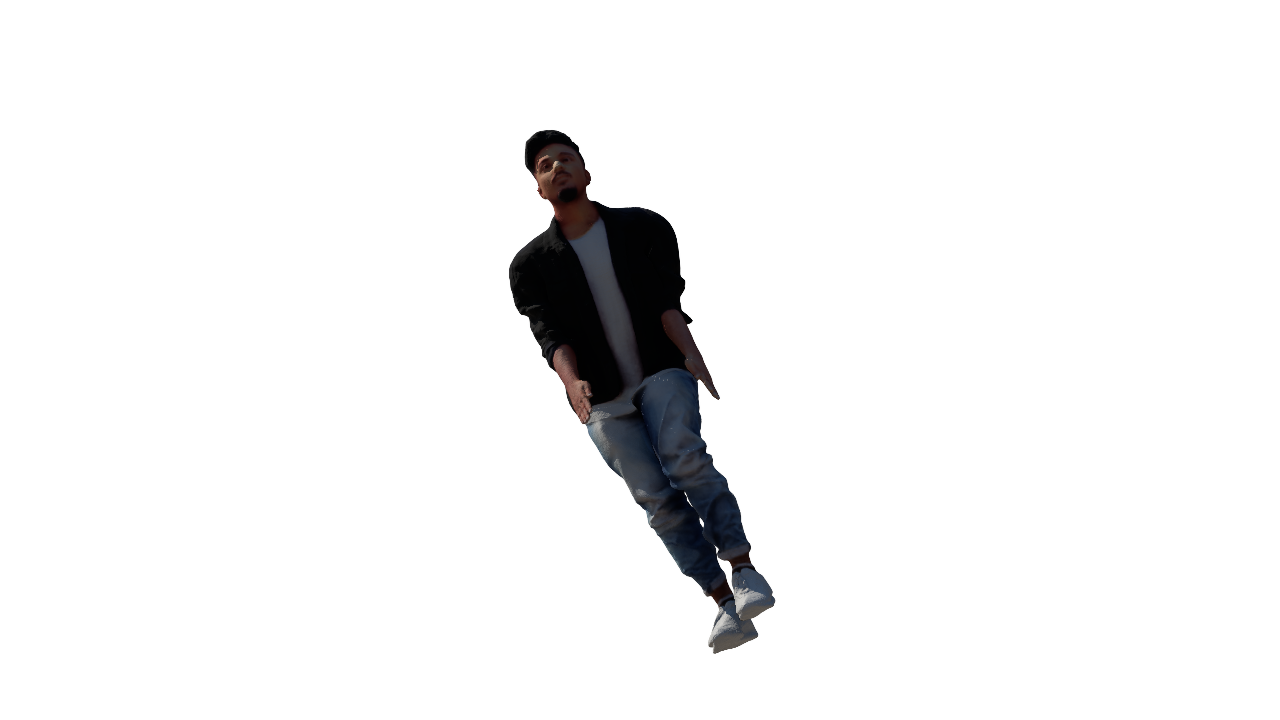}
        \caption{Repose and Relit}
    \end{subfigure}
    \begin{subfigure}[b]{0.12\linewidth}
        \includegraphics[width=\linewidth, trim=12.5cm 1cm 12.5cm 3cm, clip]{fig/images/02/rgb_train_0000_gt.png}
        \caption{Input RGB}
    \end{subfigure}
    \begin{subfigure}[b]{0.12\linewidth}
        \includegraphics[width=\linewidth, trim=12.5cm 1cm 12.5cm 3cm, clip]{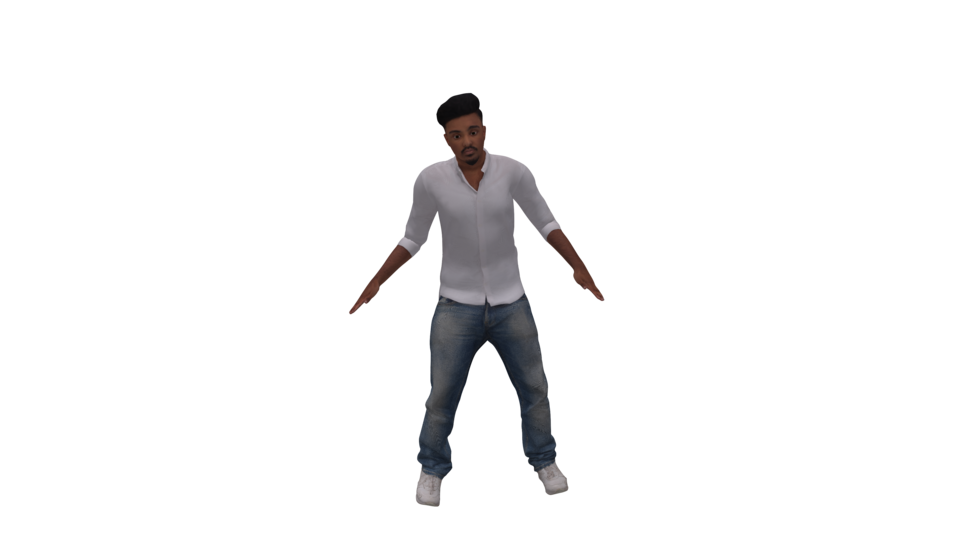}
        \caption{Albedo}
    \end{subfigure}
    \begin{subfigure}[b]{0.12\linewidth}
        \includegraphics[width=\linewidth, trim=12.5cm 1cm 12.5cm 3cm, clip]{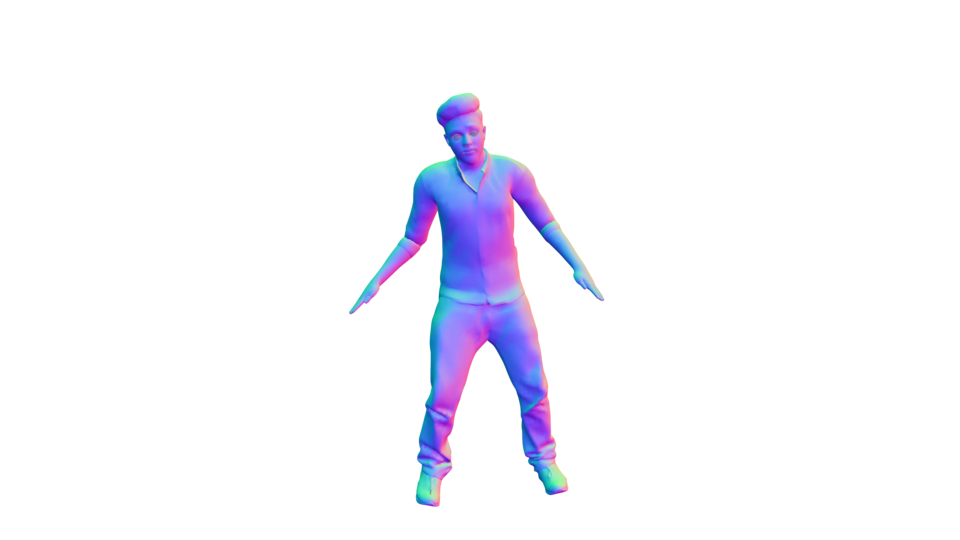}
        \caption{Geometry}
    \end{subfigure}
    \begin{subfigure}[b]{0.12\linewidth}
        \includegraphics[width=\linewidth, trim=18.5cm 1cm 14.5cm 3.5cm, clip]{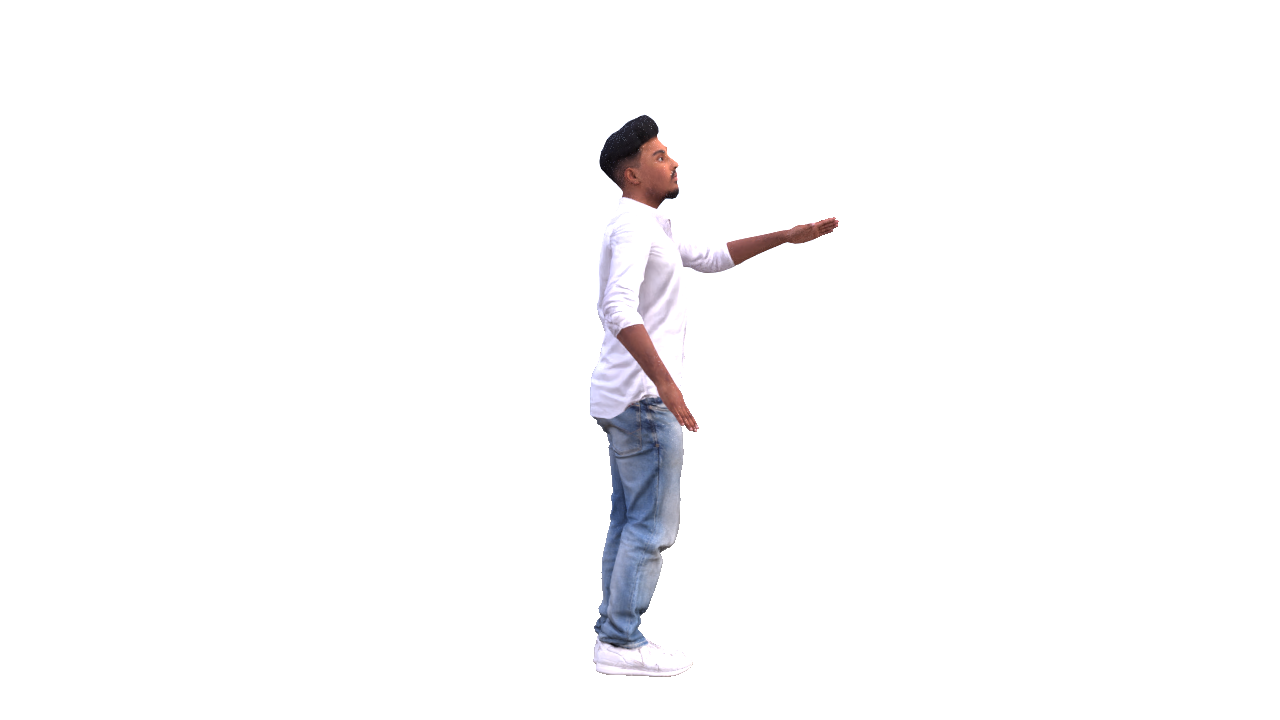}
        \caption{Repose and Relit}
    \end{subfigure}
    \caption{\textbf{Additional qualitative results on the RANA dataset.} We
    note that our method removes the shadow from the estimated albedo, whereas R4D*
    bakes shadow into albedo (column 2). On another subject, we produce albedo
    close to ground truth while R4D* produces overly dark albedo (column 6).}
    \label{app:fig:qualitative1}
\end{figure*}

\begin{figure*}[h]
    \centering
    \textbf{Ours}
    \par\medskip
    \begin{subfigure}[b]{0.12\linewidth}
        \includegraphics[width=\linewidth, trim=13cm 0.75cm 12cm 3cm, clip]{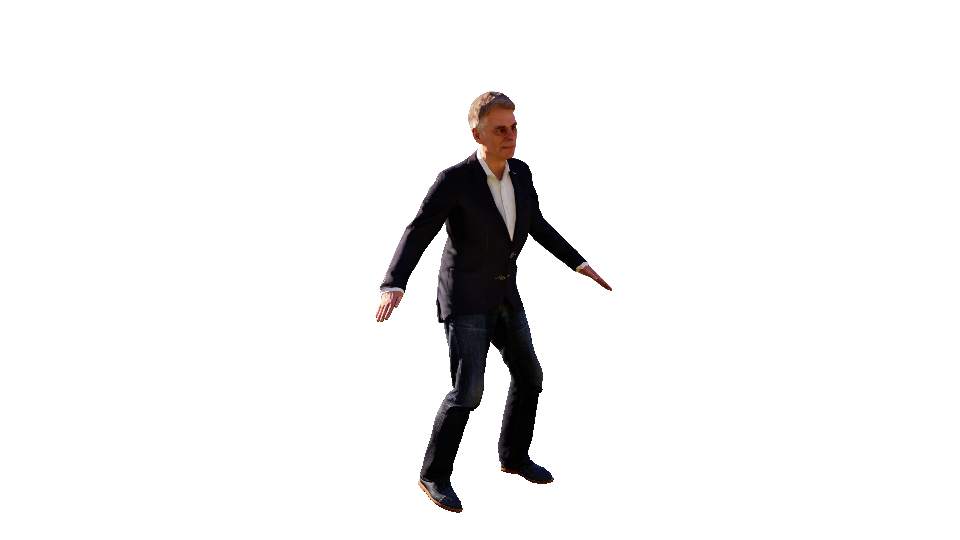}
        \caption{Input RGB}
    \end{subfigure}
    \begin{subfigure}[b]{0.12\linewidth}
        \includegraphics[width=\linewidth, trim=13cm 0.75cm 12cm 3cm, clip]{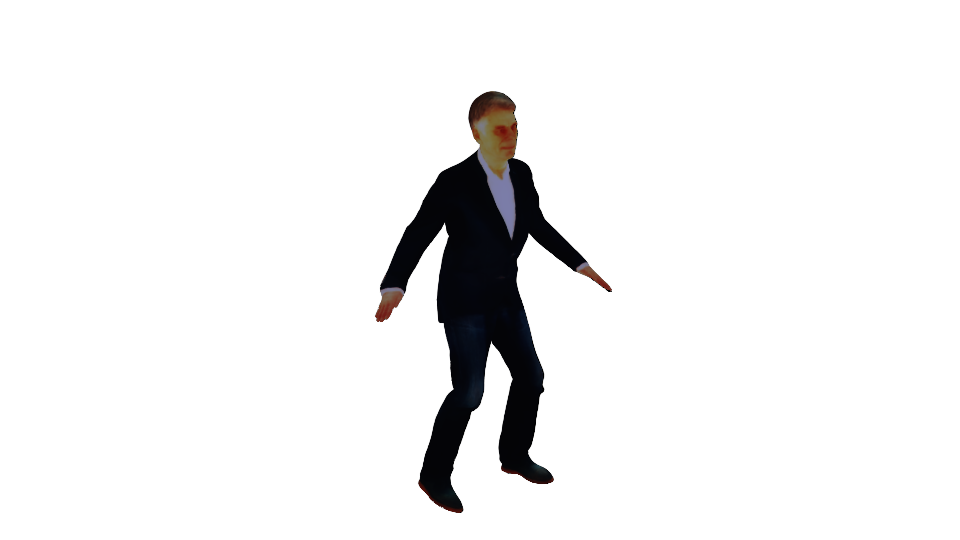}
        \caption{Albedo}
    \end{subfigure}
    \begin{subfigure}[b]{0.12\linewidth}
        \includegraphics[width=\linewidth, trim=13cm 0.75cm 12cm 3cm, clip]{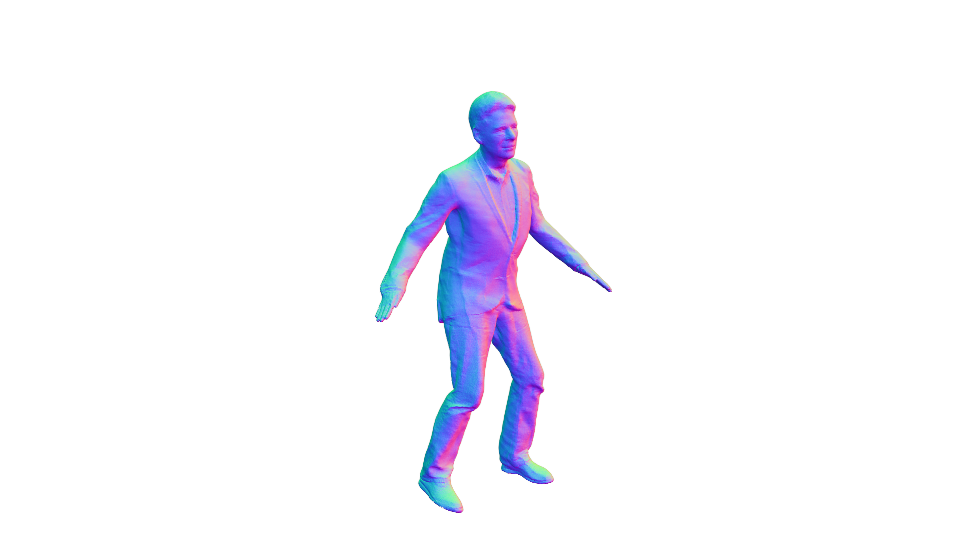}
        \caption{Geometry}
    \end{subfigure}
    \begin{subfigure}[b]{0.12\linewidth}
        \includegraphics[width=\linewidth, trim=15cm 1cm 18cm 4cm, clip]{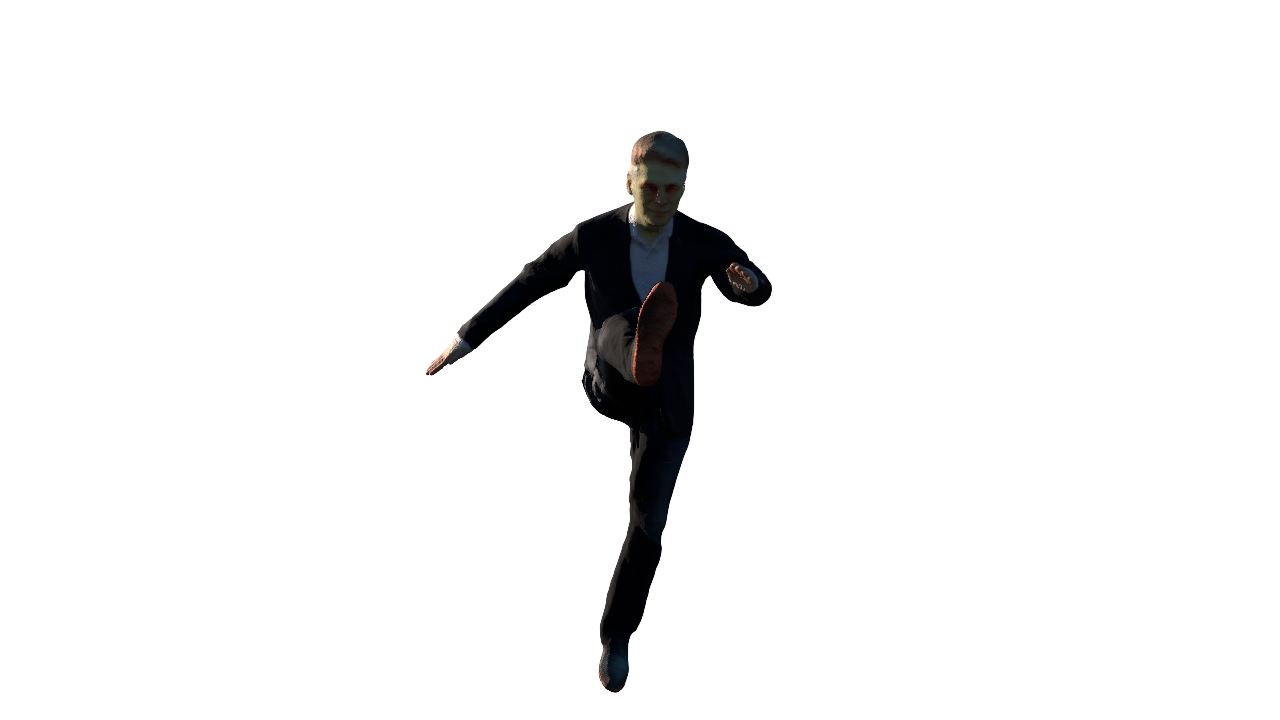}
        \caption{Repose and Relit}
    \end{subfigure}
    \begin{subfigure}[b]{0.12\linewidth}
        \includegraphics[width=\linewidth, trim=12.5cm 0.5cm 12.5cm 3cm, clip]{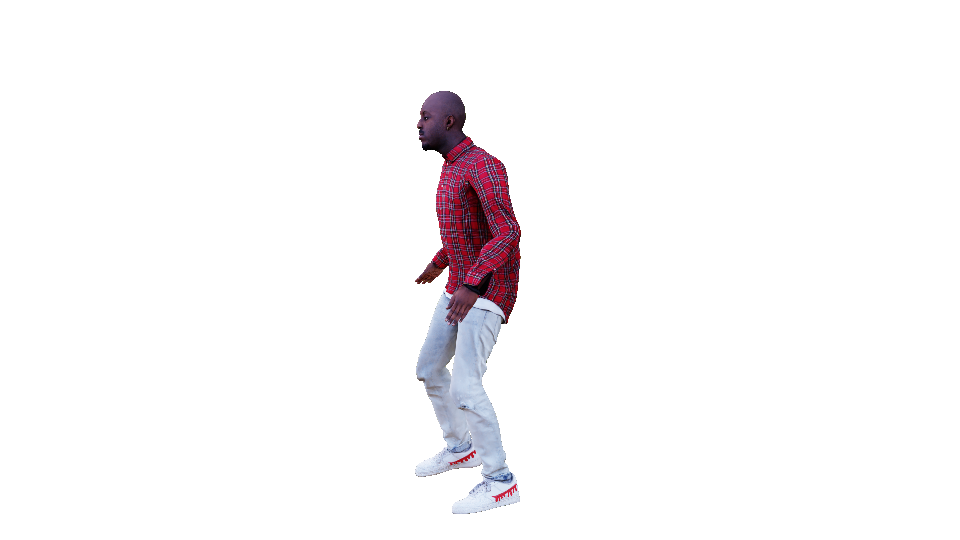}
        \caption{Input RGB}
    \end{subfigure}
    \begin{subfigure}[b]{0.12\linewidth}
        \includegraphics[width=\linewidth, trim=12.5cm 0.5cm 12.5cm 3cm, clip]{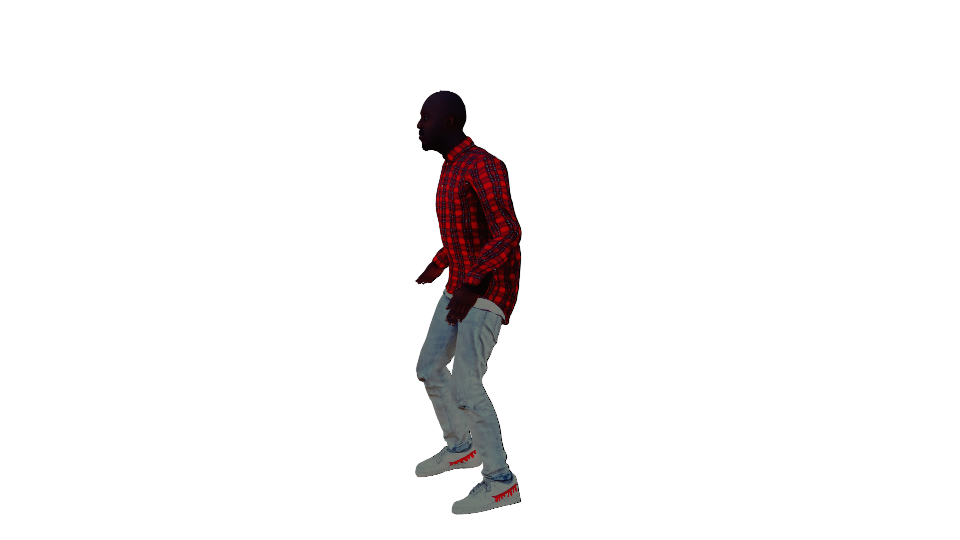}
        \caption{Albedo}
    \end{subfigure}
    \begin{subfigure}[b]{0.12\linewidth}
        \includegraphics[width=\linewidth, trim=12.5cm 0.5cm 12.5cm 3cm, clip]{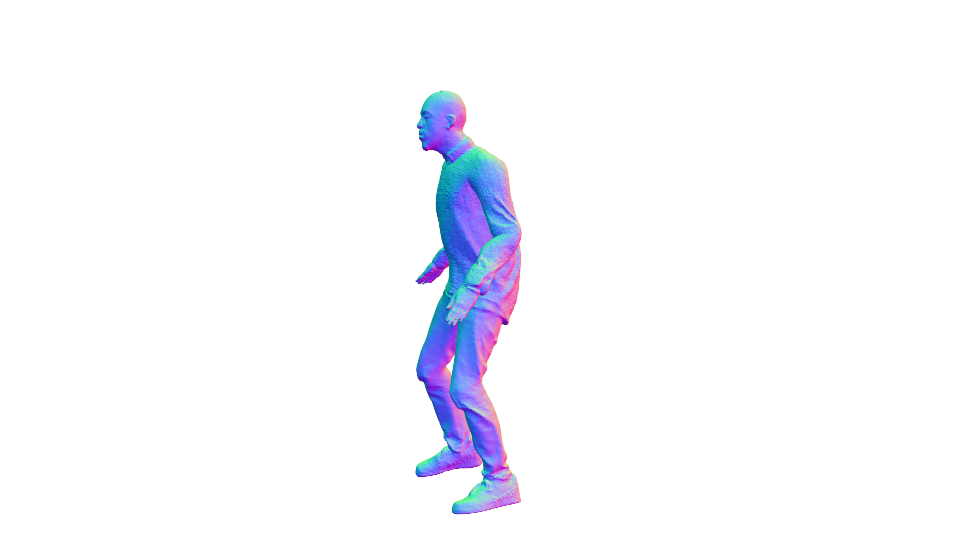}
        \caption{Geometry}
    \end{subfigure}
    \begin{subfigure}[b]{0.12\linewidth}
        \includegraphics[width=\linewidth, trim=16cm 2cm 17.5cm 4.5cm]{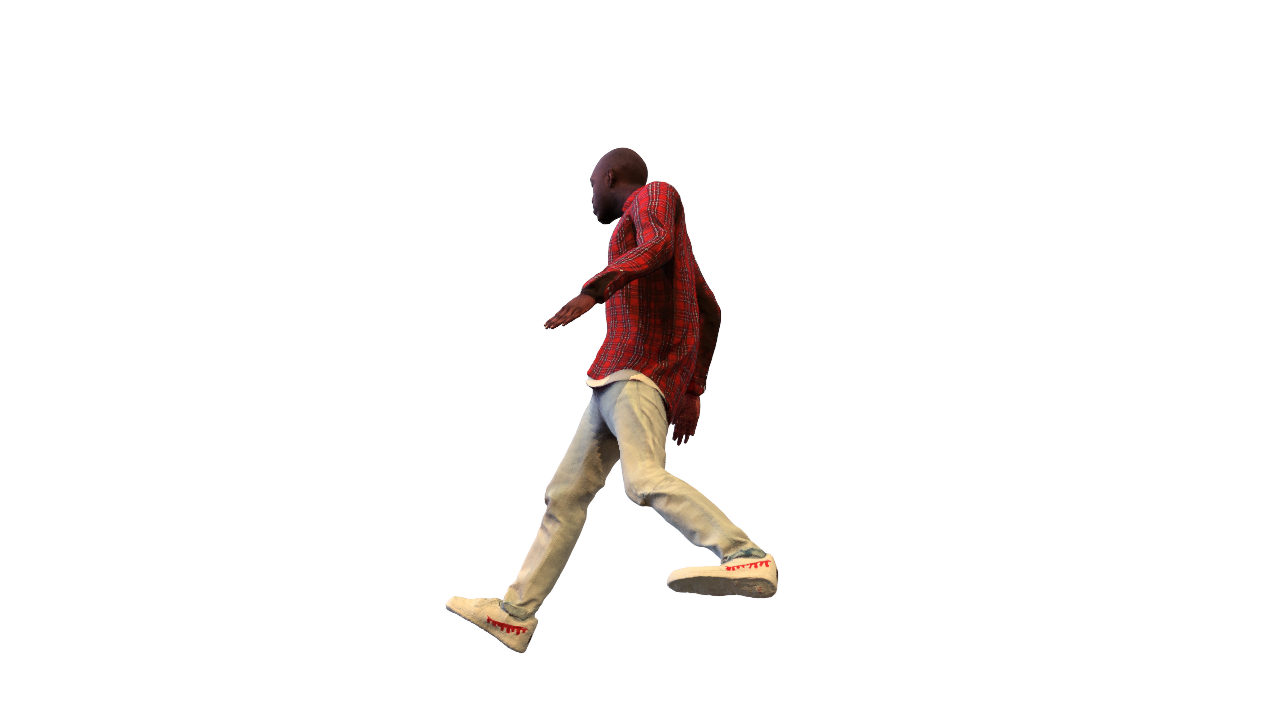}
        \caption{Repose and Relit}
    \end{subfigure}
    \vskip\baselineskip
    \textbf{R4D*}
    \par\medskip
    \begin{subfigure}[b]{0.12\linewidth}
        \includegraphics[width=\linewidth, trim=13cm 0.75cm 12cm 3cm, clip]{fig/images/06/rgb_train_0010_gt.png}
        \caption{Input RGB}
    \end{subfigure}
    \begin{subfigure}[b]{0.12\linewidth}
        \includegraphics[width=\linewidth, trim=13cm 0.75cm 12cm 3cm, clip]{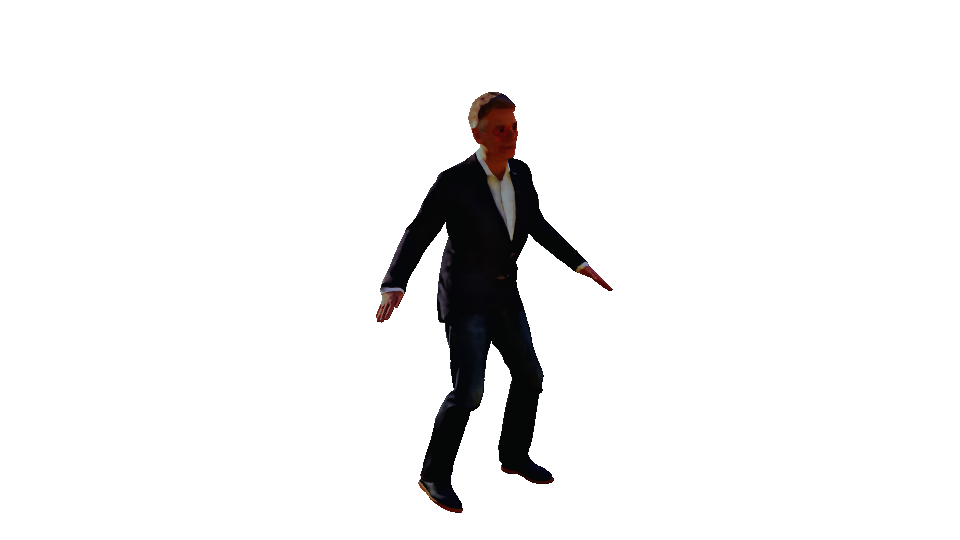}
        \caption{Albedo}
    \end{subfigure}
    \begin{subfigure}[b]{0.12\linewidth}
        \includegraphics[width=\linewidth, trim=13cm 0.75cm 12cm 3cm, clip]{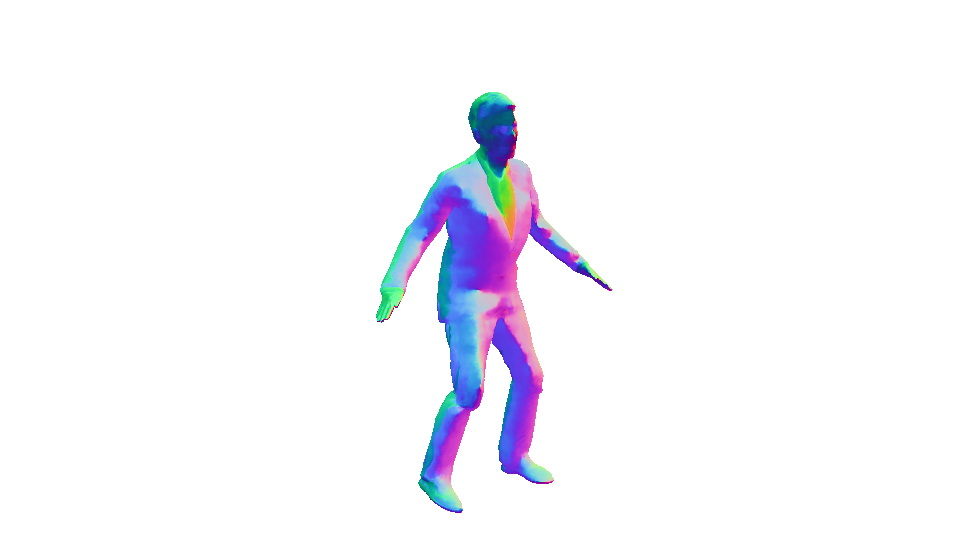}
        \caption{Geometry}
    \end{subfigure}
    \begin{subfigure}[b]{0.12\linewidth}
        \includegraphics[width=\linewidth, trim=15cm 1cm 18cm 4cm, clip]{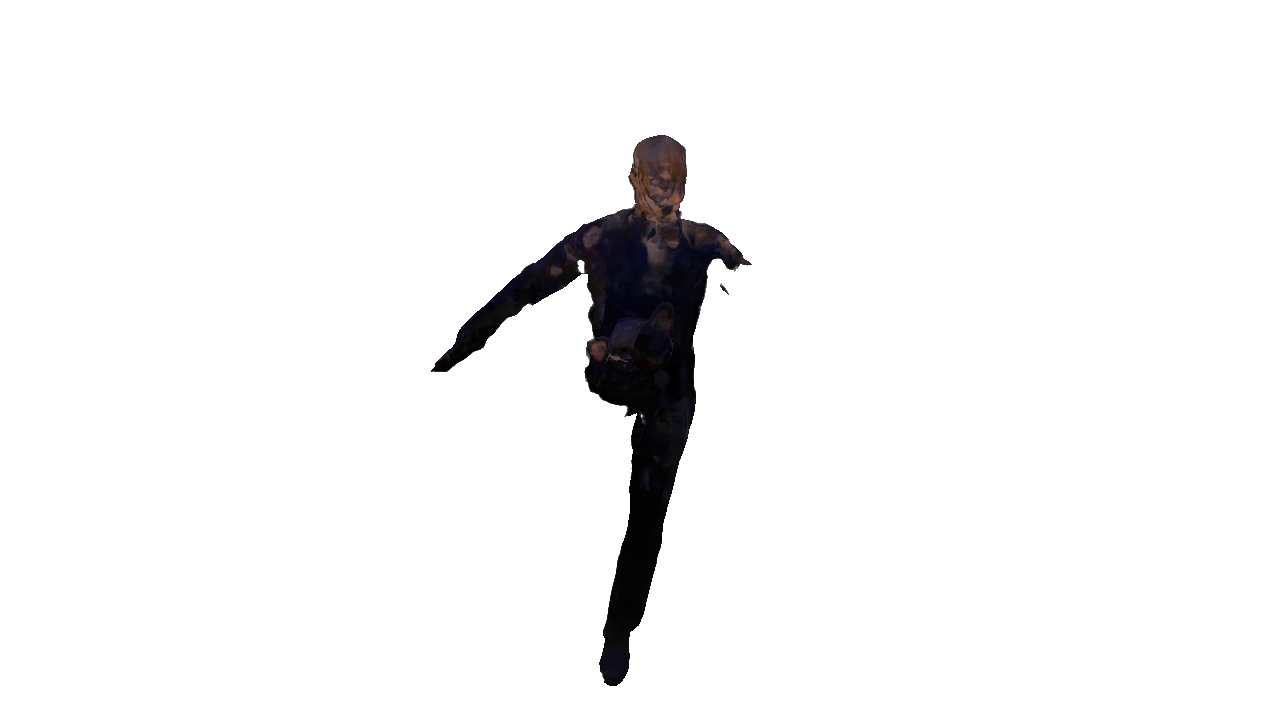}
        \caption{Repose and Relit}
    \end{subfigure}
    \begin{subfigure}[b]{0.12\linewidth}
        \includegraphics[width=\linewidth, trim=12.5cm 0.5cm 12.5cm 3cm, clip]{fig/images/33/rgb_train_0079_gt.png}
        \caption{Input RGB}
    \end{subfigure}
    \begin{subfigure}[b]{0.12\linewidth}
        \includegraphics[width=\linewidth, trim=12.5cm 0.5cm 12.5cm 3cm, clip]{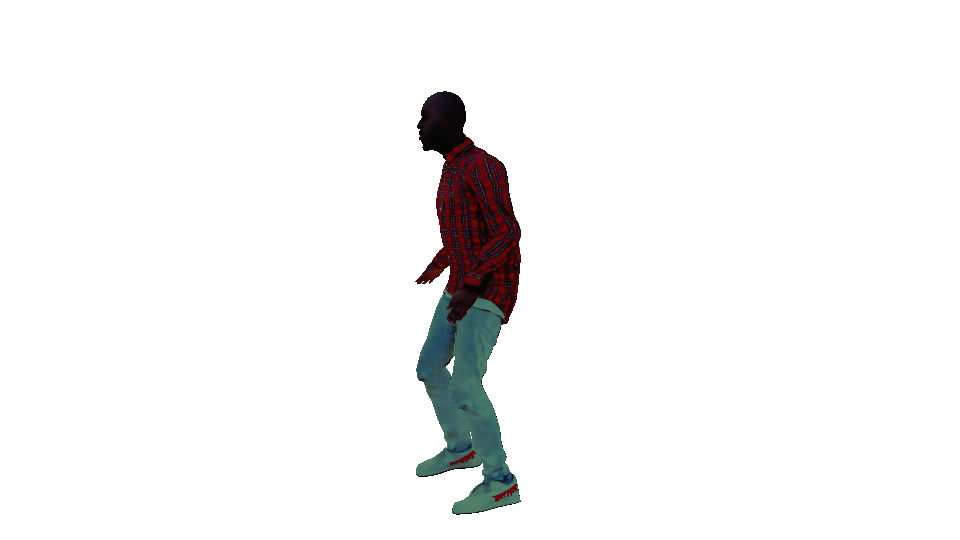}
        \caption{Albedo}
    \end{subfigure}
    \begin{subfigure}[b]{0.12\linewidth}
        \includegraphics[width=\linewidth, trim=12.5cm 0.5cm 12.5cm 3cm, clip]{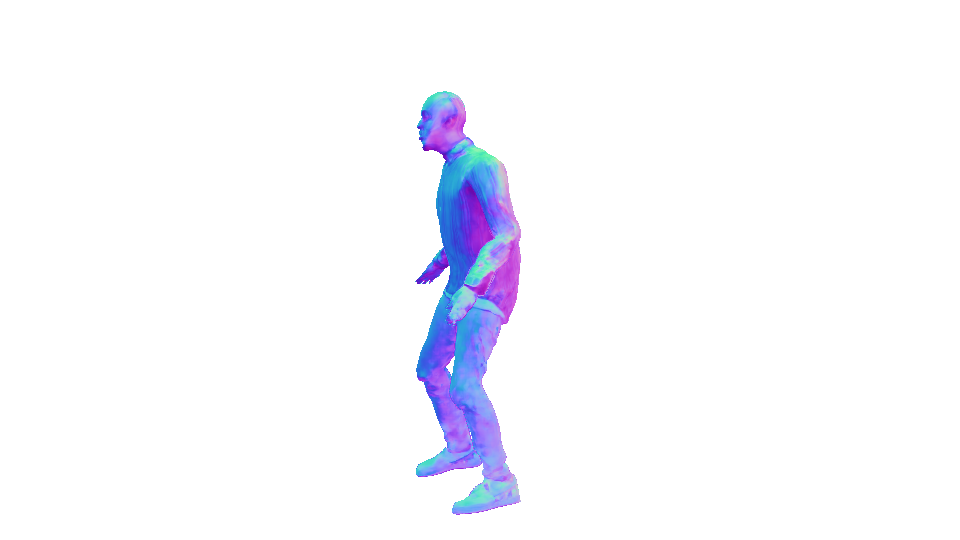}
        \caption{Geometry}
    \end{subfigure}
    \begin{subfigure}[b]{0.12\linewidth}
        \includegraphics[width=\linewidth, trim=16cm 2cm 17.5cm 4.5cm]{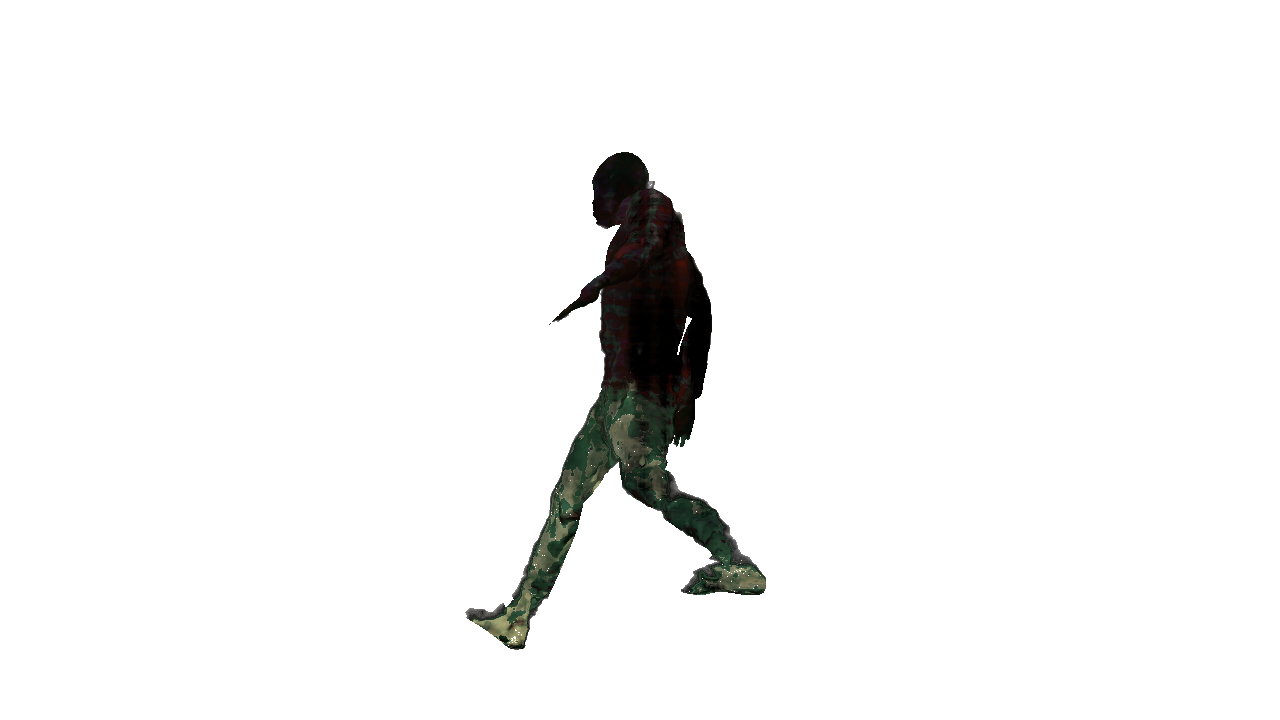}
        \caption{Repose and Relit}
    \end{subfigure}
    \vskip\baselineskip
    \textbf{Ground Truth}
    \par\medskip
    \begin{subfigure}[b]{0.12\linewidth}
        \includegraphics[width=\linewidth, trim=13cm 0.75cm 12cm 3cm, clip]{fig/images/06/rgb_train_0010_gt.png}
        \caption{Input RGB}
    \end{subfigure}
    \begin{subfigure}[b]{0.12\linewidth}
        \includegraphics[width=\linewidth, trim=13cm 0.75cm 12cm 3cm, clip]{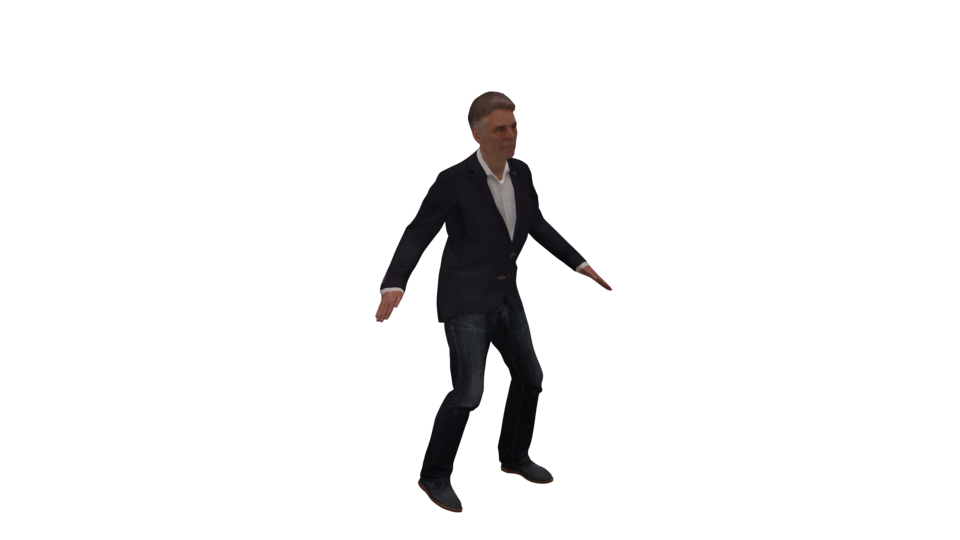}
        \caption{Albedo}
    \end{subfigure}
    \begin{subfigure}[b]{0.12\linewidth}
        \includegraphics[width=\linewidth, trim=13cm 0.75cm 12cm 3cm, clip]{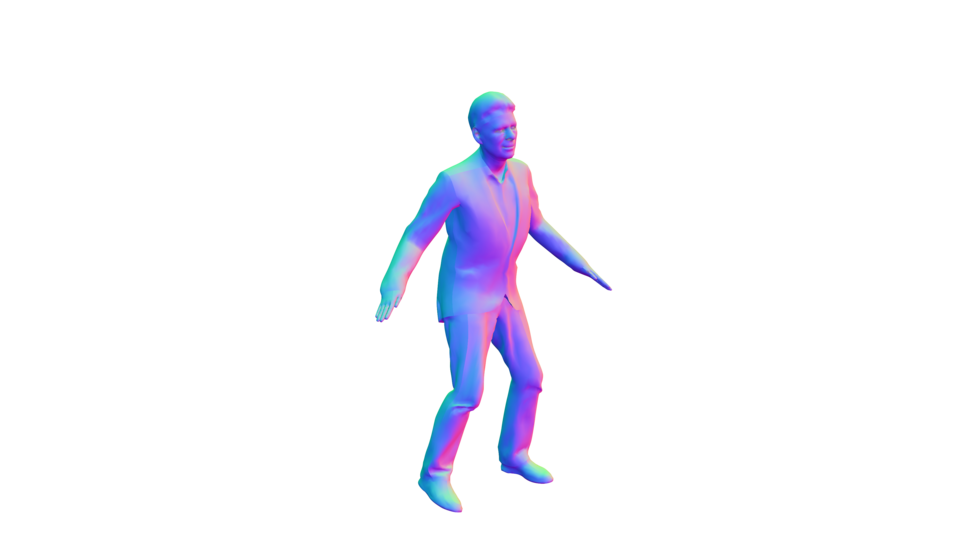}
        \caption{Geometry}
    \end{subfigure}
    \begin{subfigure}[b]{0.12\linewidth}
        \includegraphics[width=\linewidth, trim=15cm 1cm 18cm 4cm, clip]{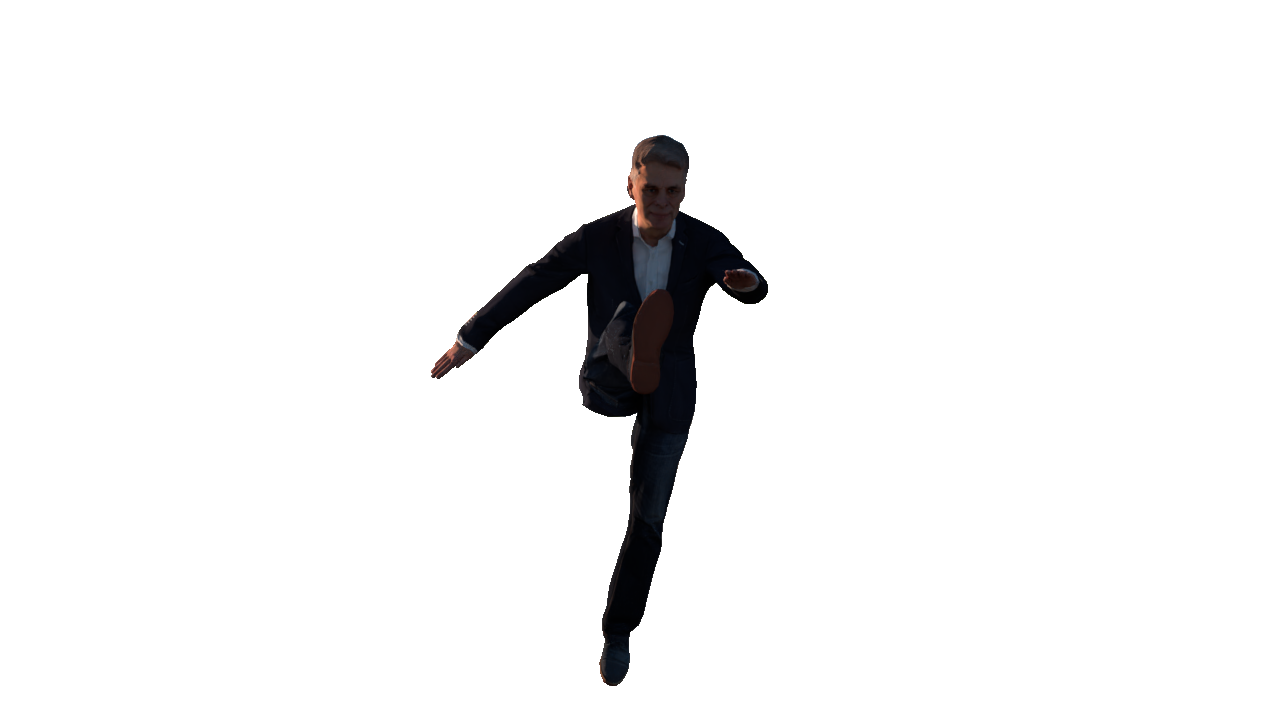}
        \caption{Repose and Relit}
    \end{subfigure}
    \begin{subfigure}[b]{0.12\linewidth}
        \includegraphics[width=\linewidth, trim=12.5cm 0.5cm 12.5cm 3cm, clip]{fig/images/33/rgb_train_0079_gt.png}
        \caption{Input RGB}
    \end{subfigure}
    \begin{subfigure}[b]{0.12\linewidth}
        \includegraphics[width=\linewidth, trim=12.5cm 0.5cm 12.5cm 3cm, clip]{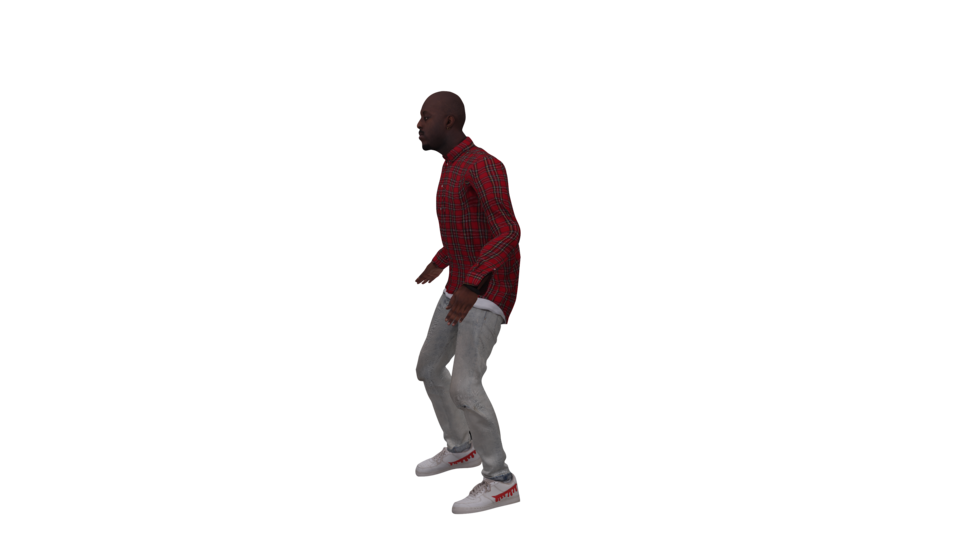}
        \caption{Albedo}
    \end{subfigure}
    \begin{subfigure}[b]{0.12\linewidth}
        \includegraphics[width=\linewidth, trim=12.5cm 0.5cm 12.5cm 3cm, clip]{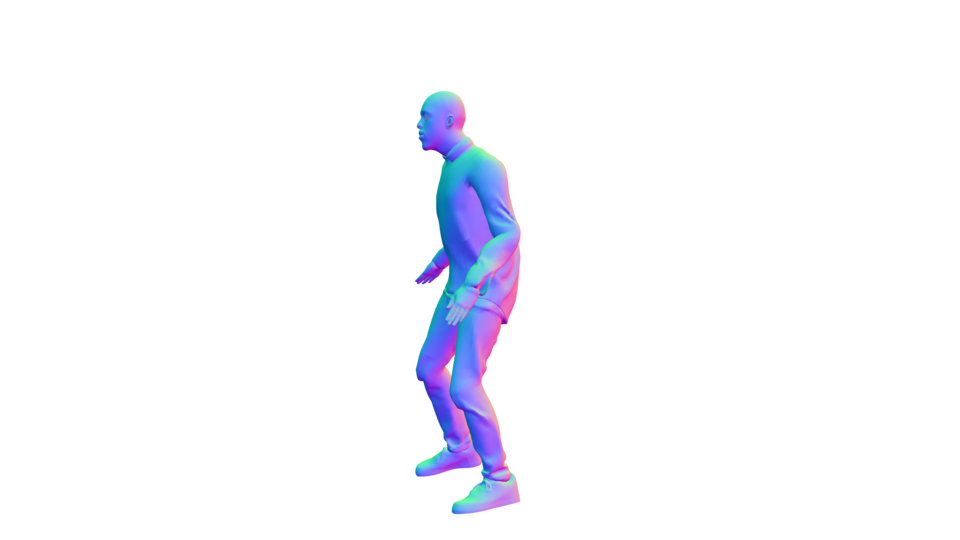}
        \caption{Geometry}
    \end{subfigure}
    \begin{subfigure}[b]{0.12\linewidth}
        \includegraphics[width=\linewidth, trim=16cm 2cm 17.5cm 4.5cm]{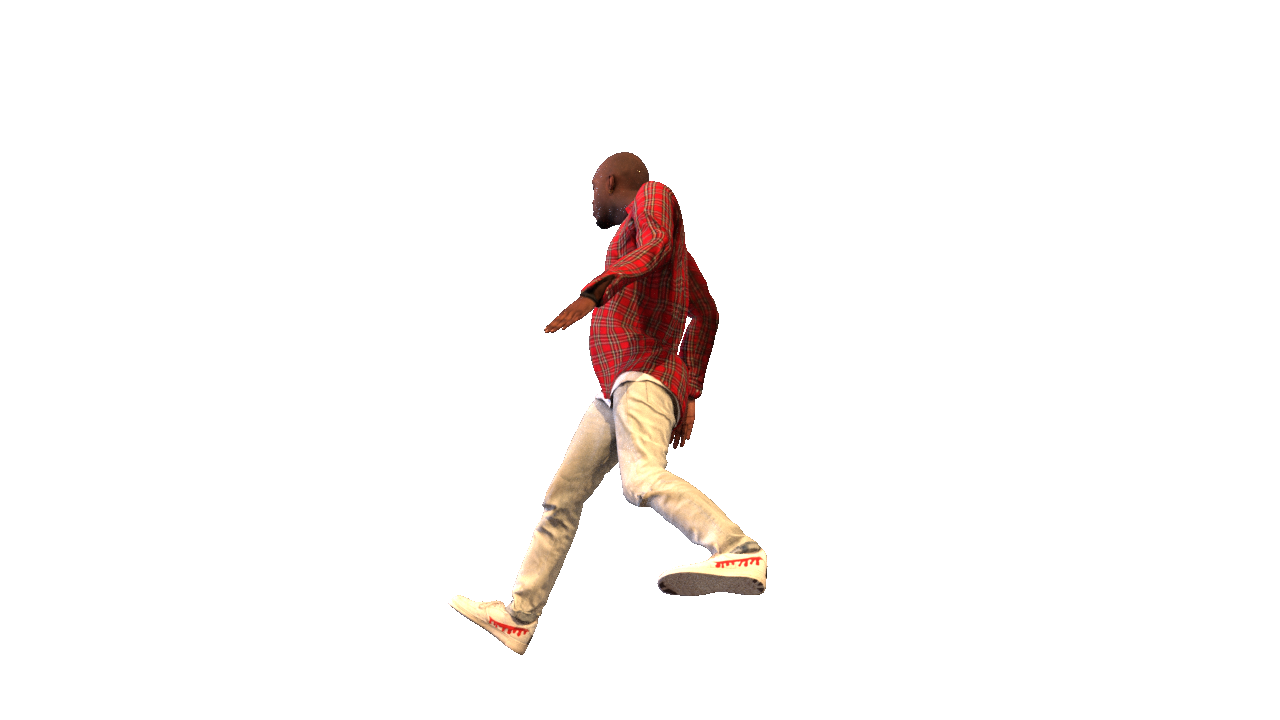}
        \caption{Repose and Relit}
    \end{subfigure}
    \caption{\textbf{Additional qualitative results on the RANA dataset.} R4D*
    demonstrates various failures in albedo estimation such as baked lighting.
    The normal estimation is also inaccurate.  In contrast, our method produces
    accurate normals and in general does not bake lighting into albedo.}
    \label{app:fig:qualitative2}
\end{figure*}

\begin{figure*}[h]
    \centering
    \textbf{Ours}
    \par\medskip
    \begin{subfigure}[b]{0.12\linewidth}
        \includegraphics[width=\linewidth, trim=8cm 1.5cm 4cm 3cm, clip]{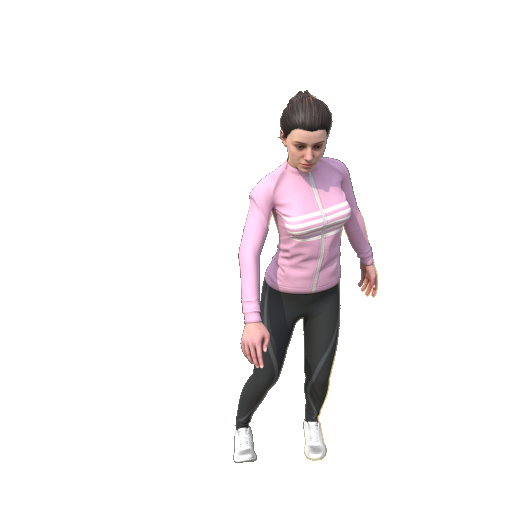}
        \caption{Input RGB}
    \end{subfigure}
    \begin{subfigure}[b]{0.12\linewidth}
        \includegraphics[width=\linewidth, trim=8cm 1.5cm 4cm 3cm, clip]{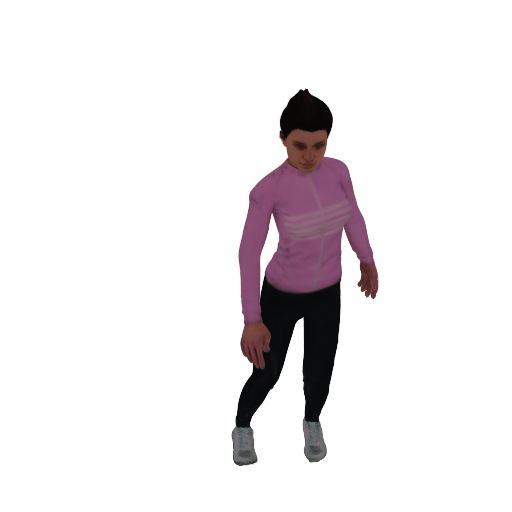}
        \caption{Albedo}
    \end{subfigure}
    \begin{subfigure}[b]{0.12\linewidth}
        \includegraphics[width=\linewidth, trim=8cm 1.5cm 4cm 3cm, clip]{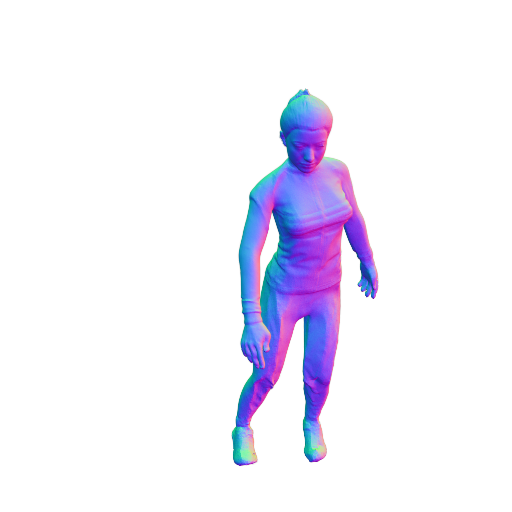}
        \caption{Geometry}
    \end{subfigure}
    \begin{subfigure}[b]{0.12\linewidth}
        \includegraphics[width=\linewidth, trim=8cm 1.5cm 4cm 3cm, clip]{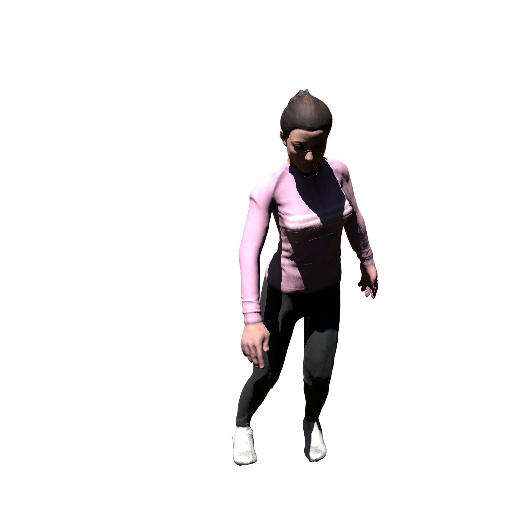}
        \caption{Relit}
    \end{subfigure}
    \begin{subfigure}[b]{0.12\linewidth}
        \includegraphics[width=\linewidth, trim=5.5cm 1cm 6cm 2cm, clip]{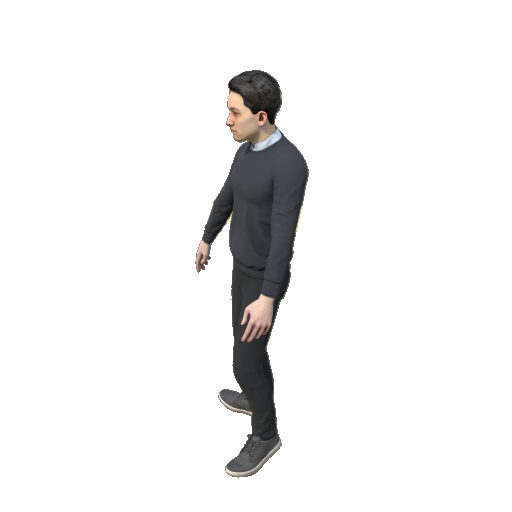}
        \caption{Input RGB}
    \end{subfigure}
    \begin{subfigure}[b]{0.12\linewidth}
        \includegraphics[width=\linewidth, trim=5.5cm 1cm 6cm 2cm, clip]{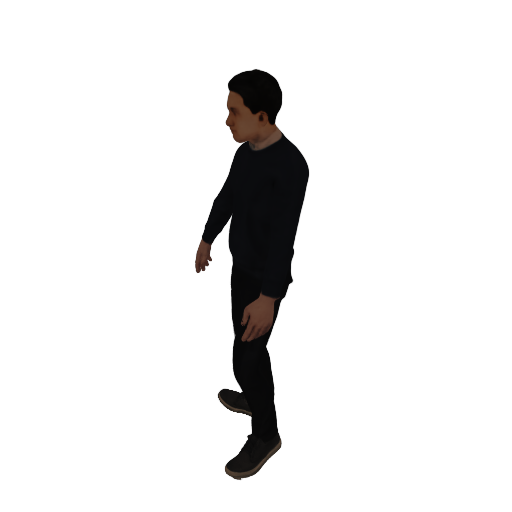}
        \caption{Albedo}
    \end{subfigure}
    \begin{subfigure}[b]{0.12\linewidth}
        \includegraphics[width=\linewidth, trim=5.5cm 1cm 6cm 2cm, clip]{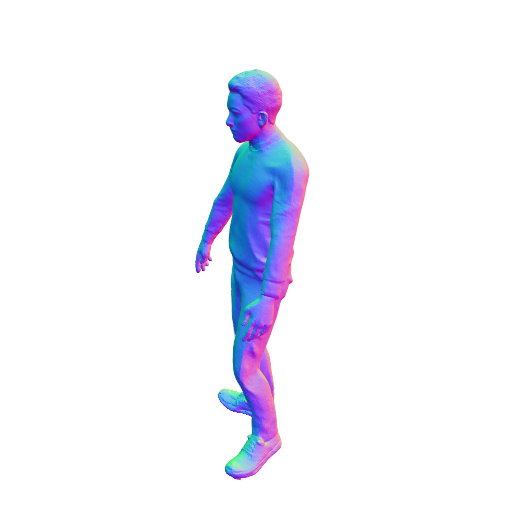}
        \caption{Geometry}
    \end{subfigure}
    \begin{subfigure}[b]{0.12\linewidth}
        \includegraphics[width=\linewidth, trim=5.5cm 1cm 6cm 2cm, clip]{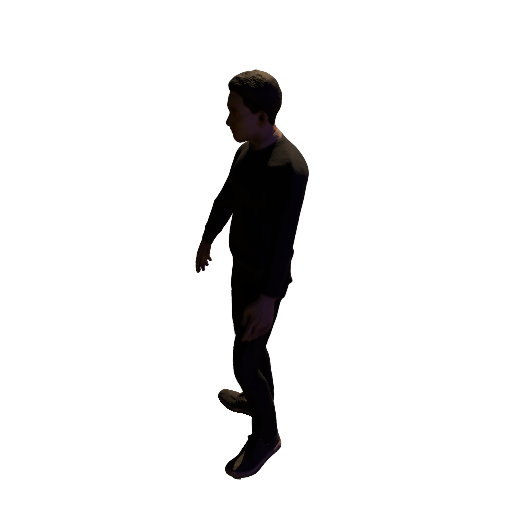}
        \caption{Relit}
    \end{subfigure}
    \vskip\baselineskip
    \textbf{R4D*}
    \par\medskip
    \begin{subfigure}[b]{0.12\linewidth}
        \includegraphics[width=\linewidth, trim=8cm 1.5cm 4cm 3cm, clip]{fig/images/jody/rgb_train_0045_gt.png}
        \caption{Input RGB}
    \end{subfigure}
    \begin{subfigure}[b]{0.12\linewidth}
        \includegraphics[width=\linewidth, trim=8cm 1.5cm 4cm 3cm, clip]{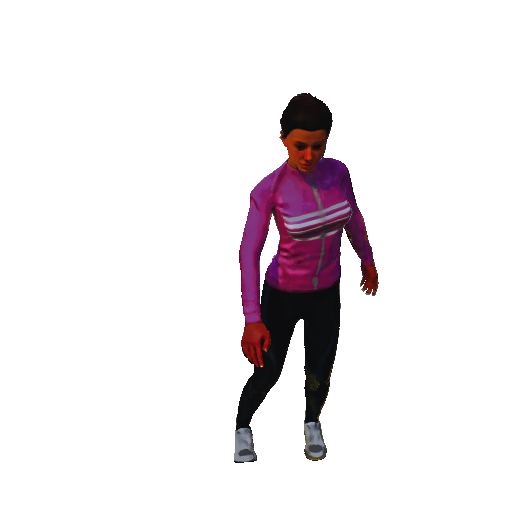}
        \caption{Albedo}
    \end{subfigure}
    \begin{subfigure}[b]{0.12\linewidth}
        \includegraphics[width=\linewidth, trim=8cm 1.5cm 4cm 3cm, clip]{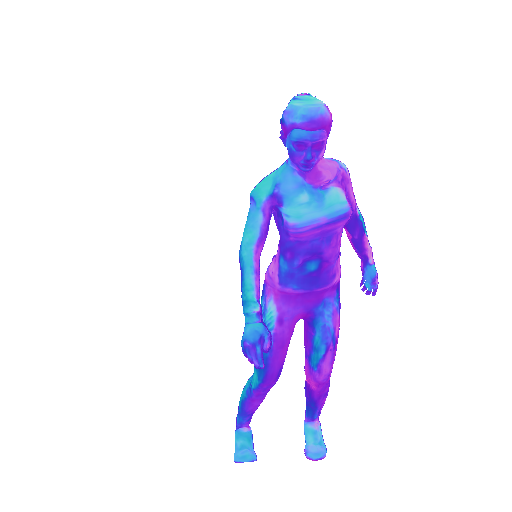}
        \caption{Geometry}
    \end{subfigure}
    \begin{subfigure}[b]{0.12\linewidth}
        \includegraphics[width=\linewidth, trim=8cm 1.5cm 4cm 3cm, clip]{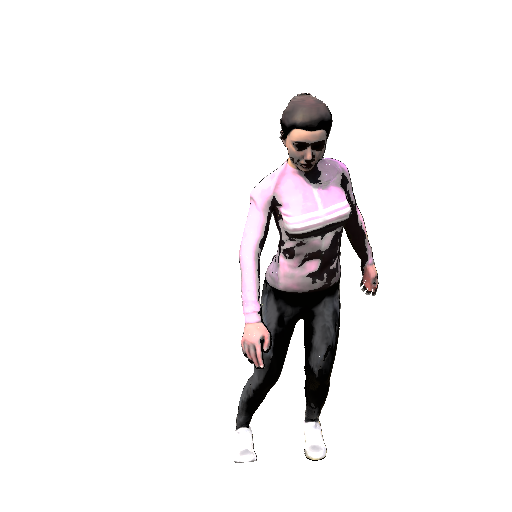}
        \caption{Relit}
    \end{subfigure}
    \begin{subfigure}[b]{0.12\linewidth}
        \includegraphics[width=\linewidth, trim=5.5cm 1cm 6cm 2cm, clip]{fig/images/leonard/rgb_train_0067_gt.png}
        \caption{Input RGB}
    \end{subfigure}
    \begin{subfigure}[b]{0.12\linewidth}
        \includegraphics[width=\linewidth, trim=5.5cm 1cm 6cm 2cm, clip]{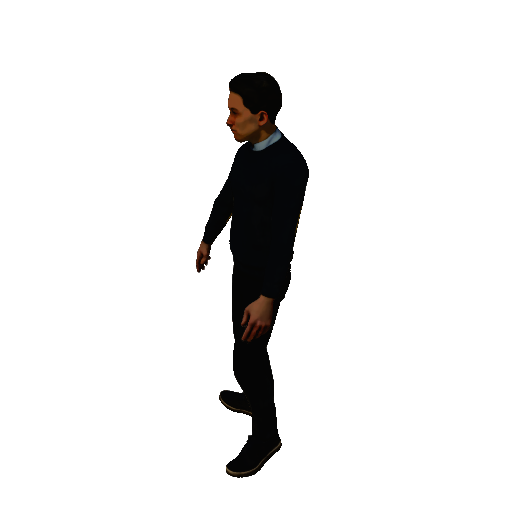}
        \caption{Albedo}
    \end{subfigure}
    \begin{subfigure}[b]{0.12\linewidth}
        \includegraphics[width=\linewidth, trim=5.5cm 1cm 6cm 2cm, clip]{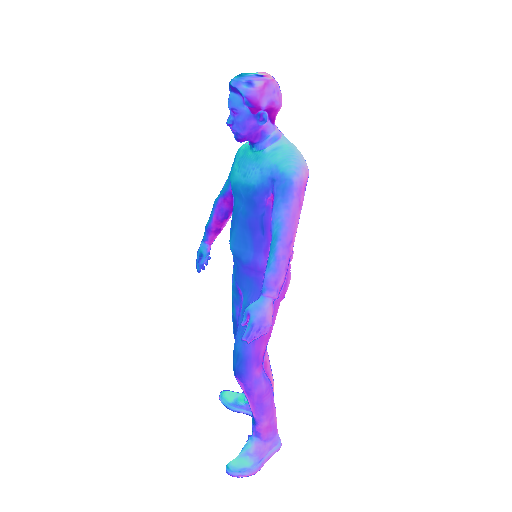}
        \caption{Geometry}
    \end{subfigure}
    \begin{subfigure}[b]{0.12\linewidth}
        \includegraphics[width=\linewidth, trim=5.5cm 1cm 6cm 2cm, clip]{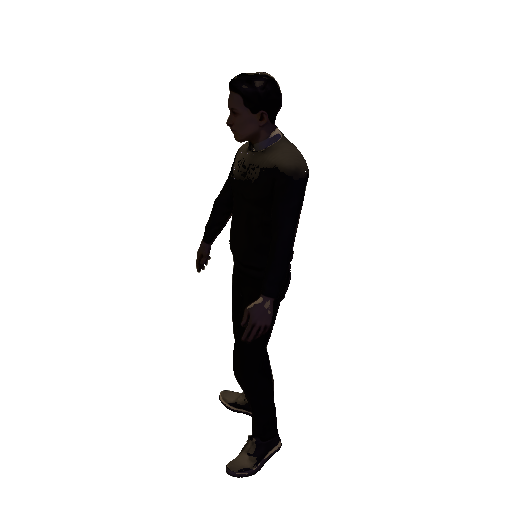}
        \caption{Relit}
    \end{subfigure}
    \vskip\baselineskip
    \textbf{Ground Truth}
    \par\medskip
    \begin{subfigure}[b]{0.12\linewidth}
        \includegraphics[width=\linewidth, trim=8cm 1.5cm 4cm 3cm, clip]{fig/images/jody/rgb_train_0045_gt.png}
        \caption{Input RGB}
    \end{subfigure}
    \begin{subfigure}[b]{0.12\linewidth}
        \includegraphics[width=\linewidth, scale=0.5, trim=16cm 3cm 8cm 6cm, clip]{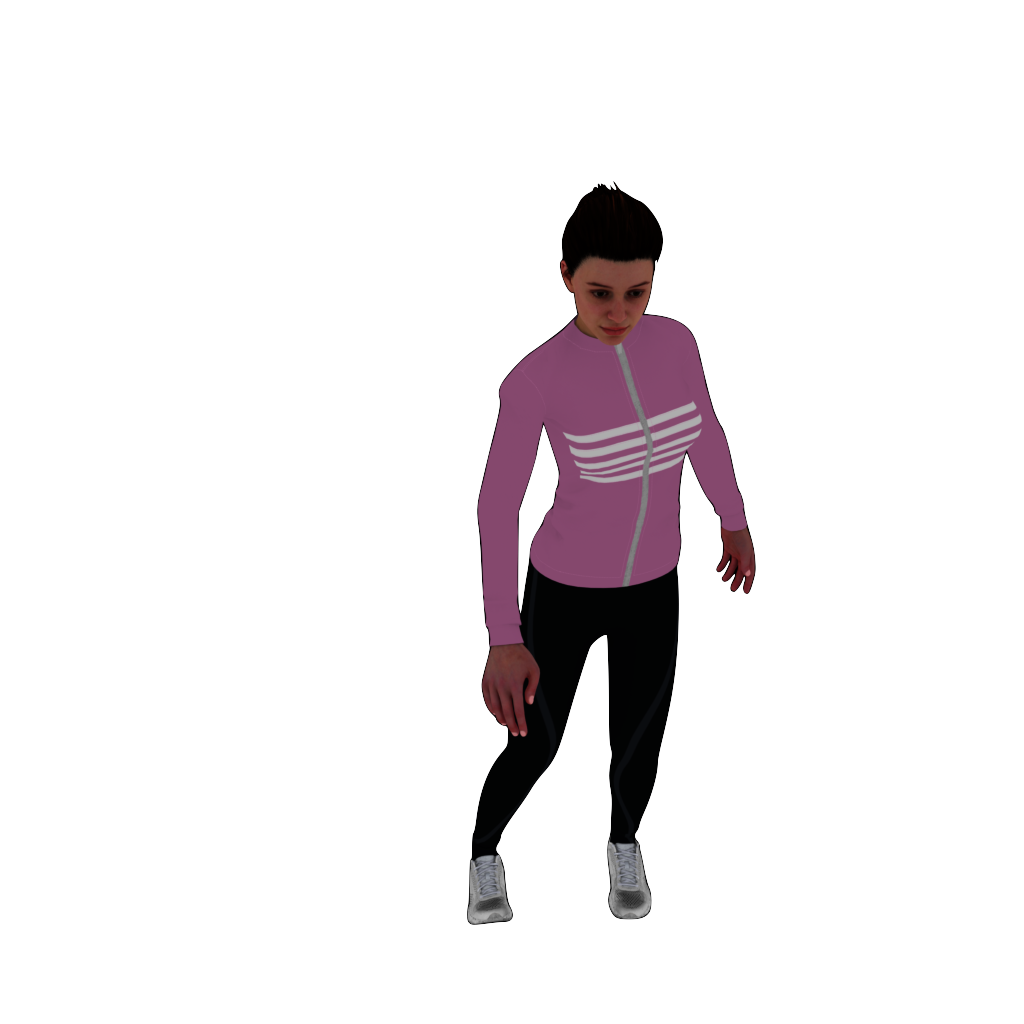}
        \caption{Albedo}
    \end{subfigure}
    \begin{subfigure}[b]{0.12\linewidth}
        \includegraphics[width=\linewidth,scale=0.5, trim=16cm 3cm 8cm 6cm, clip]{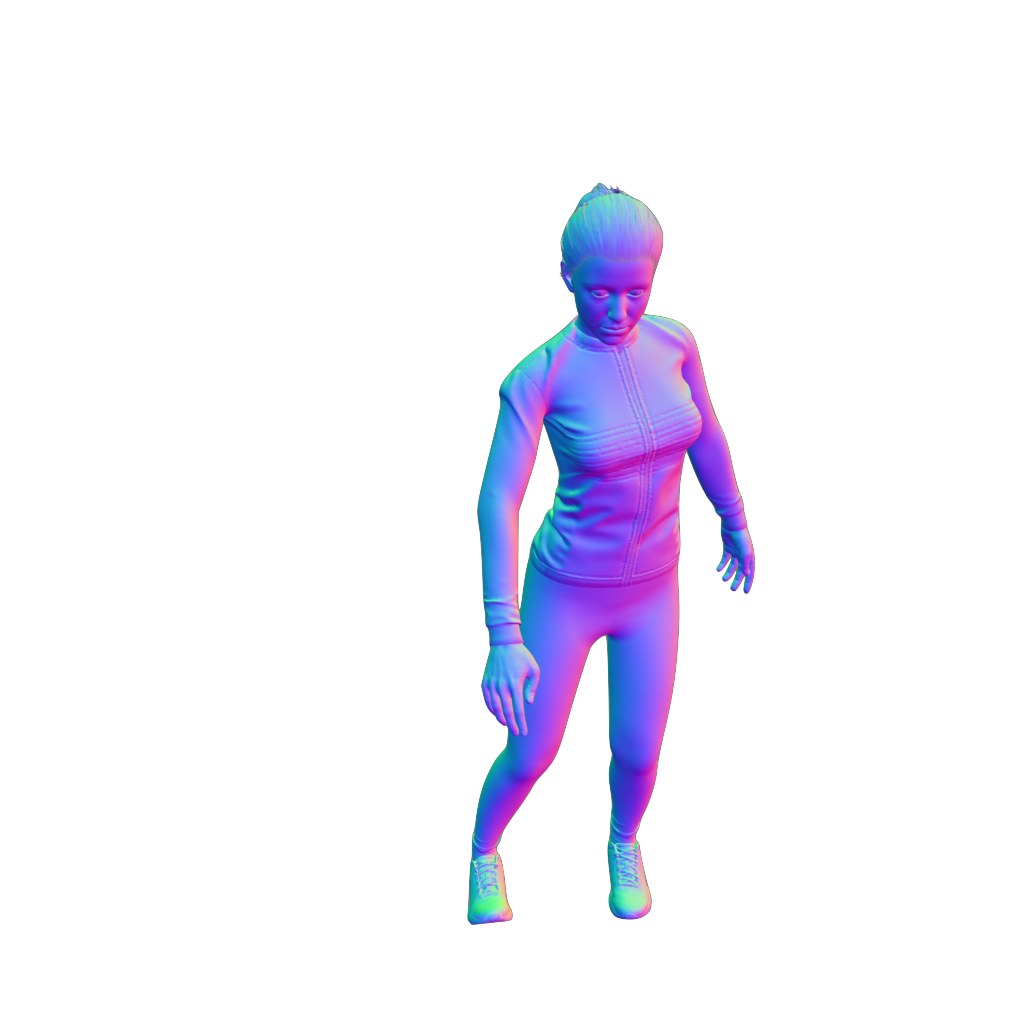}
        \caption{Geometry}
    \end{subfigure}
    \begin{subfigure}[b]{0.12\linewidth}
        \includegraphics[width=\linewidth, trim=8cm 1.5cm 4cm 3cm, clip]{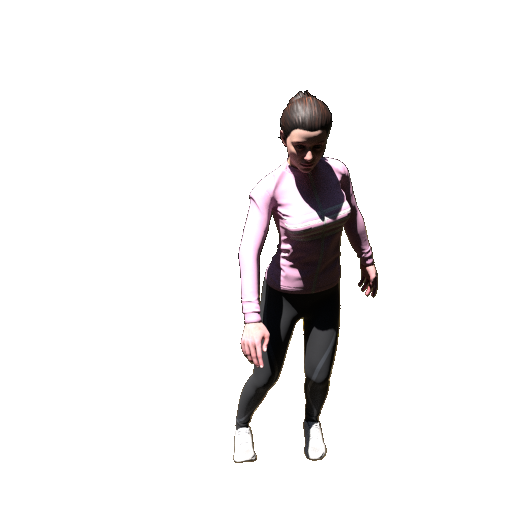}
        \caption{Relit}
    \end{subfigure}
    \begin{subfigure}[b]{0.12\linewidth}
        \includegraphics[width=\linewidth, trim=5.5cm 1cm 6cm 2cm, clip]{fig/images/leonard/rgb_train_0067_gt.png}
        \caption{Input RGB}
    \end{subfigure}
    \begin{subfigure}[b]{0.12\linewidth}
        \includegraphics[width=\linewidth,scale=0.5, trim=11cm 2cm 12cm 4cm, clip]{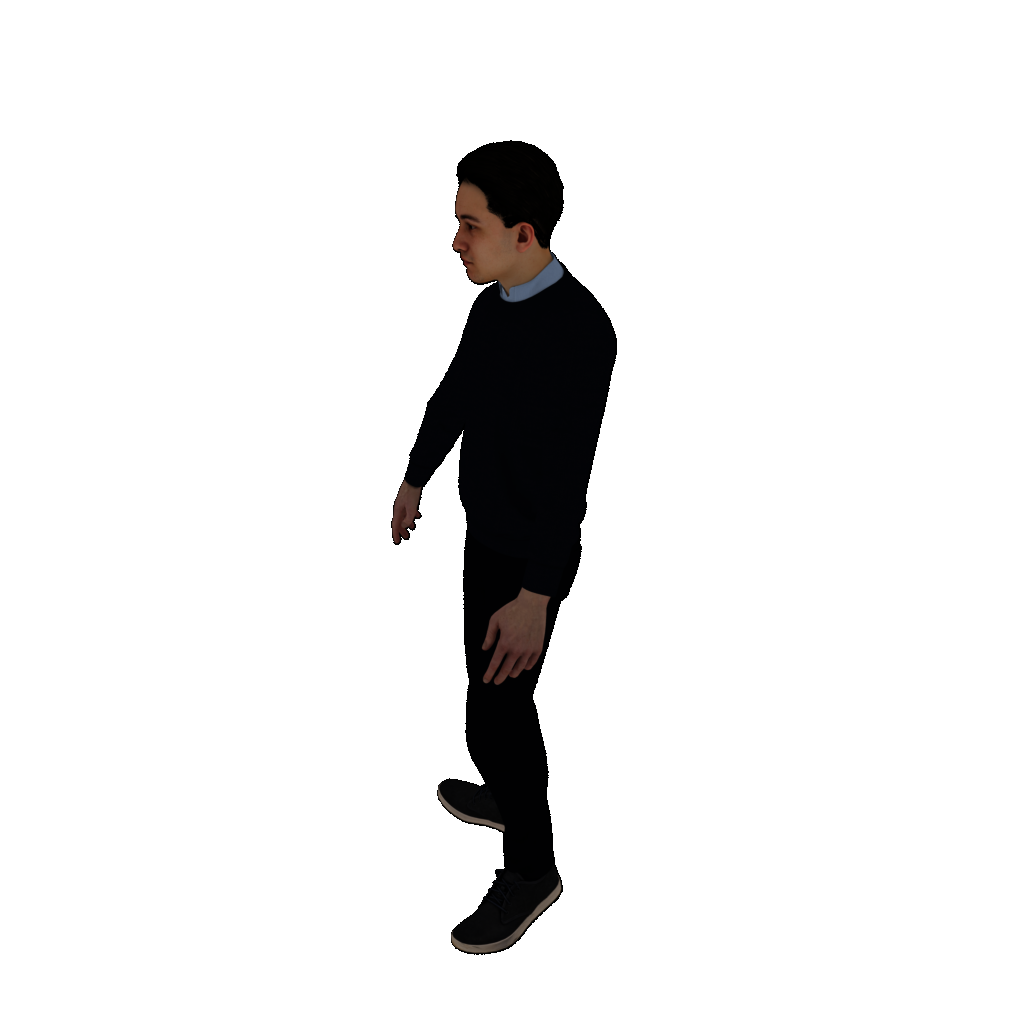}
        \caption{Albedo}
    \end{subfigure}
    \begin{subfigure}[b]{0.12\linewidth}
        \includegraphics[width=\linewidth,scale=0.5, trim=11cm 2cm 12cm 4cm, clip]{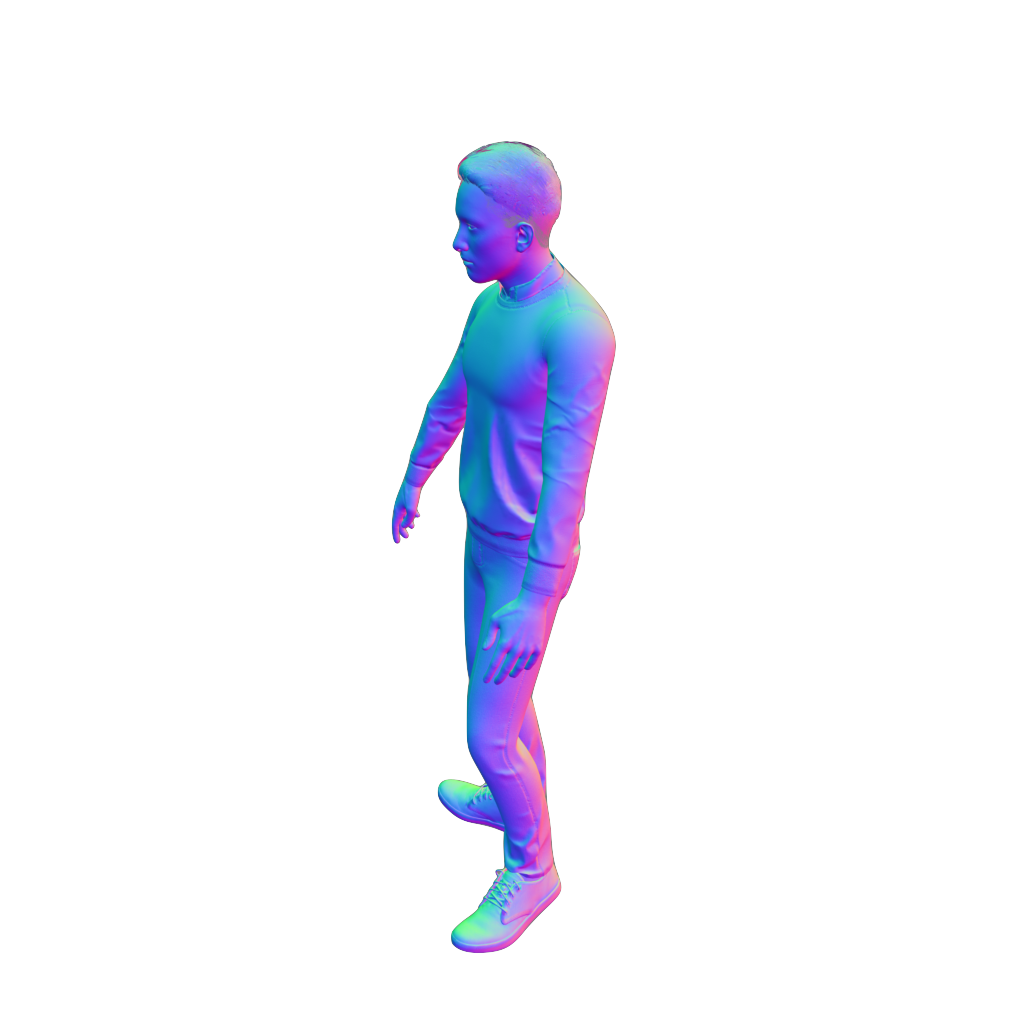}
        \caption{Geometry}
    \end{subfigure}
    \begin{subfigure}[b]{0.12\linewidth}
        \includegraphics[width=\linewidth, trim=5.5cm 1cm 6cm 2cm, clip]{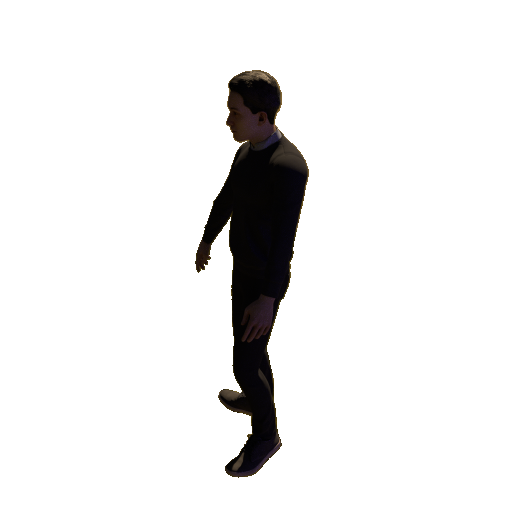}
        \caption{Relit}
    \end{subfigure}
    \caption{\textbf{Additional qualitative results on the SyntheticHuman-Relit
    dataset.} We note that R4D* tends to bake lighting and shadows not only into
    albedo but also into predicted normals. In contrast, our normals are
    obtained from the underlying SDF field and thus would not have such baked
    effects.}
    \label{app:fig:qualitative3}
\end{figure*}

\begin{figure*}[h]
    \centering
    \includegraphics[width=\linewidth]{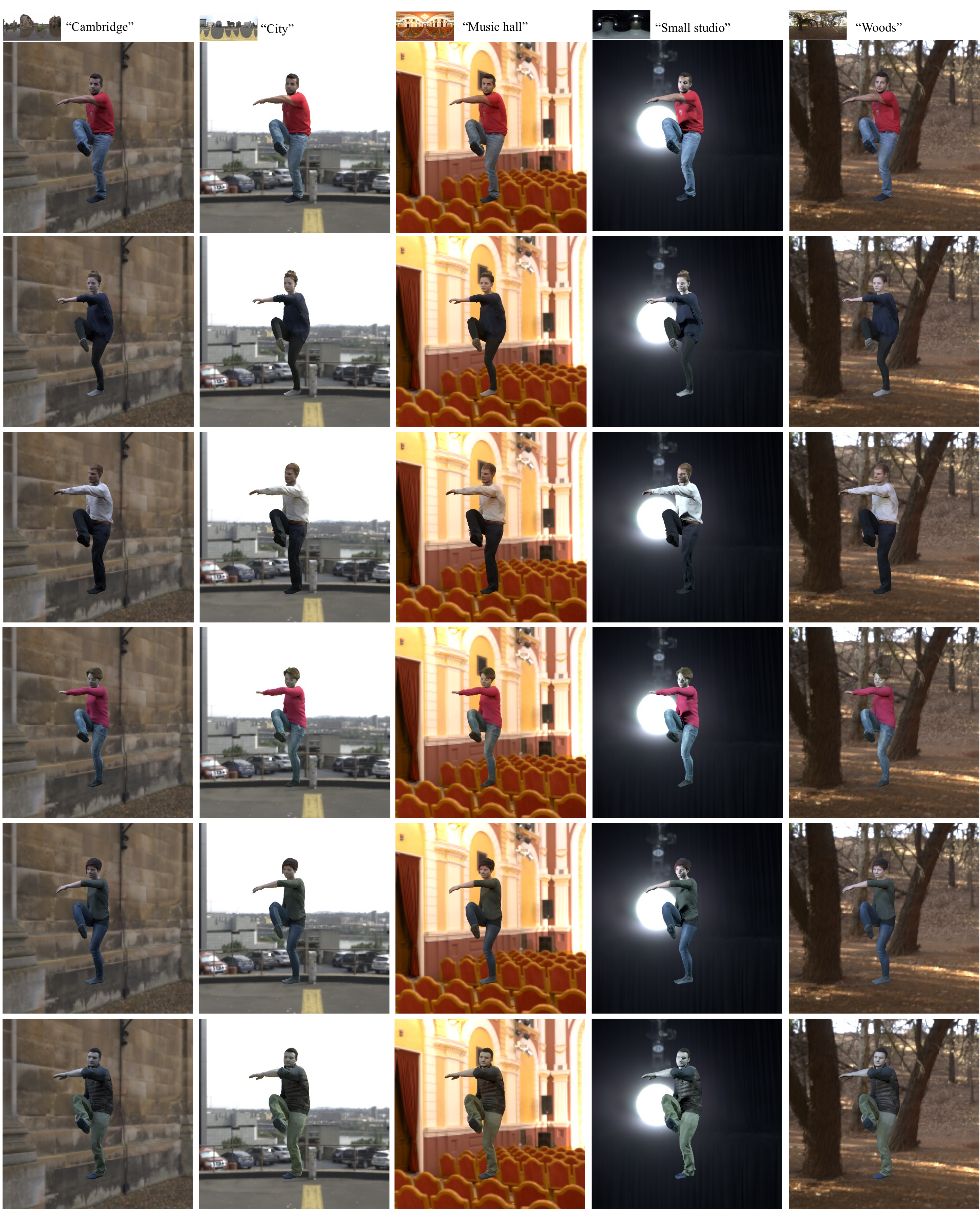}
    \caption{\textbf{Additional qualitative results on the PeopleSnapshot dataset.} We show 6 subjects from the dataset under novel pose and novel illuminations.}
    \label{app:fig:qualitative4}
\end{figure*}

\begin{figure*}[h]
    \centering
    \includegraphics[width=\linewidth]{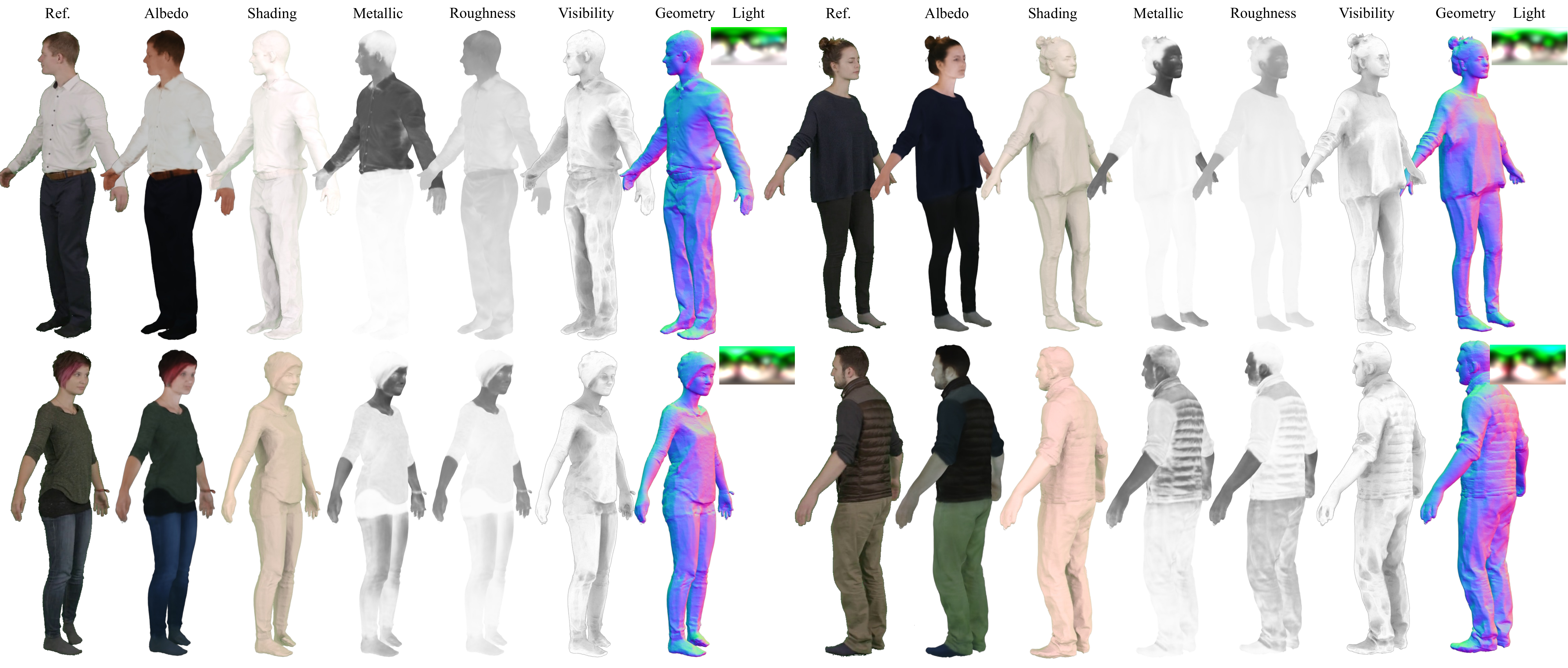}
    \caption{\textbf{Additional results on learned intrinsic properties from the PeopleSnapshot dataset.}}
    \label{app:fig:qualitative5}
\end{figure*}

\begin{figure*}[h]
    \centering
    \includegraphics[width=\linewidth]{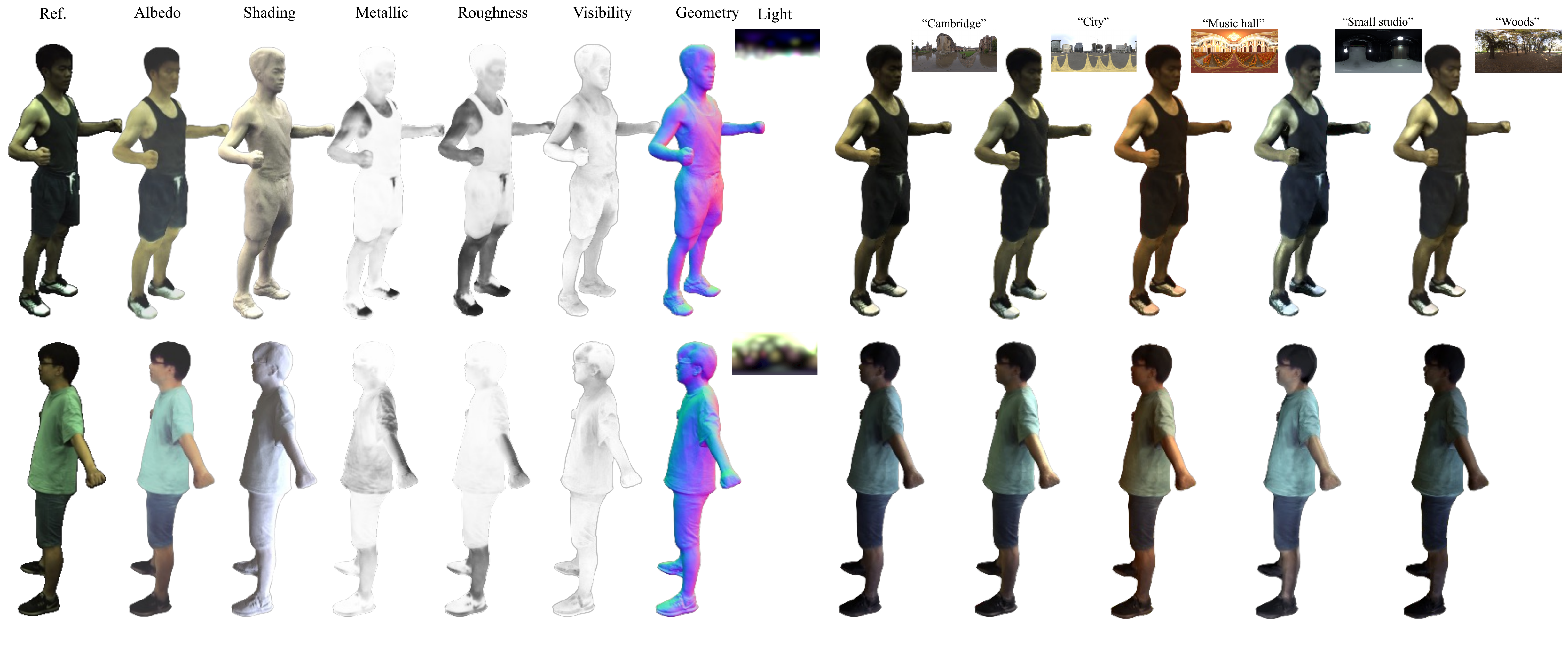}
    \caption{\textbf{Results on the ZJU-MoCap dataset.} We show results on
    monocular input (top row) and 4-view input (bottom row). Note that the
    environment lighting in the ZJU-MoCap dataset is relatively dark, resulting
    in darker albedo in the monocular setup.}
    \label{app:fig:qualitative6}
\end{figure*}

We present additional qualitative results on the RANA dataset in
Fig.~\ref{app:fig:qualitative1} and Fig.~\ref{app:fig:qualitative2}, while
Fig.~\ref{app:fig:qualitative3} shows additional qualitative results on the
SyntheticHuman-Relit dataset.  We also present more qualitative results on the
PeopleSnapshot dataset in Fig.~\ref{app:fig:qualitative4} and
Fig.~\ref{app:fig:qualitative5}.  We additionally show qualitative results on
the ZJU-MoCap~\cite{Peng2021CVPR} dataset in Fig.~\ref{app:fig:qualitative6}.

\captionsetup[subfigure]{labelformat=empty}
\setlength{\fboxsep}{0pt}

\section{Limitations and Future Work}
\label{appx:limitations}
Since we focus on video sequences for people holding still and rotating in front
of the camera, we did not consider pose-dependent non-rigid motion, similar to
the assumption of~\cite{Jiang2023CVPR,Iqbal2023ICCV}. Our approach can also fail
if estimated poses or segmentation masks are too noisy. Furthermore, our
canonical pose representation is not suitable for the animation of very loose
clothes such as skirts or capes.

Our approach is also relatively slow at inference time, requiring about 20
seconds to render a single 540x540 image on a single RTX 3090 GPU.
Regardless, our model's outputs are fully compatible with existing physically based rendering
pipelines, further acceleration can be achieved by using more optimized
implementation of volumetric scattering at inference time.

In the future, we plan to extend our approach to more challenging scenarios,
such as modeling large pose-dependent non-rigid deformations.  Applying
physically based inverse rendering for relightable scenes and avatar
reconstruction is also a promising direction~\cite{Chen2023CVPR,Liu2023ICCV}.

\end{document}